\documentclass{article}
\usepackage{graphicx} % Required for inserting images
\usepackage{subcaption}
\usepackage{url}
\usepackage{hyperref}
\usepackage{xcolor}
\usepackage{float}
\usepackage{amsfonts}
\usepackage{amsmath}
\usepackage{bm}
\usepackage[ruled,vlined]{algorithm2e}
\usepackage{tikz-cd}
\usepackage[margin=1in]{geometry} % for margins

\title{ARC-AGI-2 Technical Report}
\author{Wallyson Lemes de Oliveira \and Mekhron Bobokhonov \and Matteo Caorsi \and Aldo Podestà \and Gabriele Beltramo \and Luca Crosato \and Matteo Bonotto \and Federica Cecchetto \and Hadrien Espic \and Dan Titus Salajan \and Stefan Taga \and Luca Pana \and Joe Carthy}
\date{November 2025}

\begin{document}

\maketitle

\begin{abstract}

The Abstraction and Reasoning Corpus (ARC) is designed to assess generalization beyond pattern matching, requiring models to infer symbolic rules from very few examples. In this work, we present a transformer-based system that advances ARC performance by combining neural inference with structure-aware priors and online task adaptation. Our approach is built on four key ideas. First, we reformulate ARC reasoning as a sequence modeling problem using a compact task encoding with only 125 tokens, enabling efficient long-context processing with a modified LongT5 architecture. Second, we introduce a principled augmentation framework based on group symmetries, grid traversals, and automata perturbations, enforcing invariance to representation changes. Third, we apply \textit{test-time training} (TTT) with lightweight LoRA adaptation, allowing the model to specialize to each unseen task by learning its transformation logic from demonstrations. Fourth, we design a symmetry-aware decoding and scoring pipeline that aggregates likelihoods across augmented task views, effectively performing ``multi-perspective reasoning'' over candidate solutions.

We demonstrate that these components work synergistically: augmentations expand hypothesis space, TTT sharpens local reasoning, and symmetry-based scoring improves solution consistency. Our final system achieves a significant improvement over transformer baselines and surpasses prior neural ARC solvers, closing the gap toward human-level generalization.

\end{abstract}

\tableofcontents

\newpage

\section{Introduction}

The Abstract Reasoning Corpus (ARC), introduced by François Chollet \cite{Chollet2019}, poses a unique challenge in artificial intelligence by requiring models to solve puzzles with limited data and a strong emphasis on generalization and abstract reasoning. Traditional machine learning and deep learning approaches often struggle with such tasks due to their reliance on large datasets and fixed patterns, making ARC a benchmark for assessing the ability of AI systems to reason and adapt to novel problems.

In this work, we tackle the ARC challenge through a comprehensive and carefully engineered pipeline built around a LongT5 encoder -- decoder architecture. 
Our approach begins with an extensive \emph{offline training} phase in which the model is exposed to a stratified combination of curriculum learning, multi-task learning, and grokking techniques. 
This stage enables the model to progressively acquire an understanding of the underlying reasoning patterns within ARC tasks and to develop robust internal representations.

Once a sufficiently strong offline model is obtained, we further enhance its performance at inference time through a series of \emph{online reasoning} strategies. 
These include test-time training (TTT), adaptive decoding and scoring procedures based on spatial and color symmetries, as well as filtering mechanisms that act as guardrails to discard implausible candidate solutions. 
The latter are implemented via white-box functions encoding domain-specific priors, ensuring consistency between generated hypotheses and the logical structure of the tasks.

All components of this system are designed to operate under the strict computational constraints of the Kaggle evaluation environment, which allows processing only 240 tasks within 12 hours using \(4 \times L4\) GPUs. 
Despite these limitations, our implementation leverages both high-performance optimization and algorithmic efficiency, enabling a 200M-parameter LongT5 model to achieve 27\% on the semiprivate evaluation set, a competitive result given the computational constraints. (see Table~\ref{tab:arc_datasets}).

\paragraph{Challenges and Methodology}

\begin{table}[htbp]
\centering
\begin{tabular}{lc p{7cm}}
\hline
\textbf{Dataset} & \textbf{Tasks} & \textbf{Description} \\
\hline
Training Set & 1000 tasks &
Uncalibrated, public; a spectrum of difficulty ranging from very easy to very difficult for both humans and AI. Designed to expose and teach Core Knowledge Priors. Used to train systems. \\
\\[-4pt]
Public Eval Set & 120 tasks &
Calibrated, public; all tasks solved pass@2 by at least two humans. Used to test systems. \\
\\[-4pt]
Semi-Private Eval Set & 120 tasks &
Calibrated, not public; all tasks solved pass@2 by at least two humans. Used for Kaggle live contest leaderboard and ARC Prize leaderboard. ``Semi-private'' means these tasks may have been exposed to limited third parties (e.g.\ via API). \\
\\[-4pt]
Private Eval Set & 120 tasks &
Calibrated, not public; all tasks solved pass@2 by at least two humans. Used for Kaggle final contest leaderboard. ``Private'' means these tasks have not been exposed to third parties. \\
\hline
\end{tabular}
\caption{Overview of ARC-AGI datasets and their properties.}
\label{tab:arc_datasets}
\end{table}

The ARC dataset consists of only 1120 public tasks, split evenly into training and evaluation sets (see Table~\ref{tab:arc_datasets}). 
This limited dataset size makes it unsuitable for training large models directly, necessitating extensive data augmentation. 
To overcome this, we not only incorporated other publicly available ARC-like datasets (see Table~\ref{tab:open_source_data}), but also developed a comprehensive data generation pipeline leveraging symmetry-based methods, traversal-driven transformations, and cellular automata. 
This pipeline produced more than 2.3 million synthetic ARC-like tasks, providing a solid foundation for training a compact yet capable large language model (LLM): LongT5 (approximately 200M parameters).

Beyond dataset design, we devoted substantial effort to the computational optimization of both training and inference, given the strict hardware and time constraints of the Kaggle environment (only \(4 \times L4\) GPUs and 12 hours for 240 tasks). 
Our encoder-decoder architecture was heavily customized, integrating optimized FlashAttention kernels and multiple low-level improvements to maximize GPU throughput and memory efficiency. 
In parallel, we refined grid representations to better capture pixel-wise and relational consistency, thereby improving convergence speed and final accuracy under limited compute.

Offline fine-tuning on the augmented dataset allowed the model to acquire general reasoning patterns, which we then evaluated on an internal set of 177 hand-crafted tasks. 
However, since ARC tasks often involve reasoning rules not represented in the training data, we further incorporated \emph{test-time training} (TTT) to enable task-specific adaptation. 
TTT dynamically augments and fine-tunes the model at inference time, allowing it to handle distribution shifts and previously unseen reasoning patterns.

Finally, simple greedy decoding proved insufficient to exploit the multiple perspectives inherent to each task. 
We therefore generated and evaluated around 180 candidate outputs per task, ranking them using a symmetry-based scoring function \cite{Franzen2024}. 
This filtering mechanism served as a white-box guardrail, removing implausible solutions and enforcing logical and structural priors.

\subsection{Contributions}
\label{sec:contributions}

This work advances the state of the art on the ARC-AGI challenge by integrating large language model architectures with structured prior knowledge and online reasoning capabilities. 
Through a combination of high-performance computing optimizations and algorithmic improvements, our system constitutes -- to the best of our knowledge -- one of the most versatile LongT5-centered pipelines to date.

Our main contributions are as follows:

\begin{enumerate}
    \item \textbf{Offline Training Recipe.}
    We designed a stratified training approach that combines several complementary strategies: curriculum-based learning to help the model progressively build its understanding of the ARC-AGI challenge, multi-task learning to enable the model not only to solve tasks but also to denoise them, and grokking to maximize performance and foster a robust internal representation of the underlying rules.
    \item \textbf{Test-Time Training for Per-Task Adaptation.}
    We introduce a novel application of \emph{Test-Time Training} (TTT) to ARC-AGI, enabling the model to adapt dynamically to each unseen task using only the input--output examples provided in the prompt. Unlike traditional finetuning, our TTT approach uses lightweight LoRA \cite{hu2022lora} updates with a task-specific external memory component, achieving localized specialization without catastrophic forgetting.

    \item \textbf{Structure-Aware Data Augmentation.}
    We propose three principled domain-aware augmentation techniques that inject strong structural priors into training:
    \begin{itemize}
        \item \emph{Symmetry-based augmentations}, which slightly modify the tasks rule and alter the pixel representation, teaching the model how to better generalize.
        \item \emph{Automata-based perturbations}, which preserve task semantics while altering surface representations, teaching the model rule invariance.
        \item \emph{Grid traversals} (e.g.\ snake, row-by-row), which present the same grid under alternative sequential views, exposing relational patterns and reinforcing representation robustness.
    \end{itemize}
    These augmentations improve generalization by forcing the model to internalize transformation rules rather than overfitting to spatial tokenization biases.

    \item \textbf{Symmetry-Aware Scoring.}
    Having been largely inspired by the 2024 winning solution (\cite{Franzen2024}), we leverage a scoring mechanism that evaluates candidate solutions across symmetry transformations, including rotations and reflections. Rather than committing to a single representation, our method explores multiple \emph{perspectives} of each task and uses majority agreement to validate or reject hypotheses. This leads to a more reliable evaluation pipeline and enables a better ranking method to select the best solution compared to occurrences-based ones.

\end{enumerate}

Together, these contributions demonstrate that combining neural architectures with knowledge priors, self-adaptation, and perspective-invariant reasoning yields significant improvements in ARC-AGI performance, moving closer to robust and generalizable learning.

\paragraph{\label{sec:kaggle}Kaggle results in a nutshell.} We started our kaggle journey with \textbf{llama3.1 1B} and scored 3.75\% in the first submission. We better tuned the hyper-parameters and got 5.00\%. We switched to the \textbf{LongT5} architecture (Section~\ref{sec:longt5}) and increased the score to 7.08\%. We serendipitously realized that we were running our pipeline on half of the Kaggle hidden dataset: when running over the full dataset we raised to 12.08\%. Of course, this required HPC-type optimizations to fit the time budget (Section~\ref{sec:flash_attn}). We then introduced grokking (Section~\ref{sec:grokking}) during offline training and reached 19.86\%. We eventually added 2-traversals (Section~\ref{sec:traversal}) and went up to 25.00\%. The final boost was given by multitask learning (Section~\ref{sec:training_to_understand}): it got us to 27.08\%.

\subsection{Background on the Abstract Reasoning Corpus (ARC)}

The Abstract Reasoning Corpus (ARC), proposed by François Chollet \cite{Chollet2019}, is a benchmark designed to evaluate general artificial intelligence by testing an agent's ability to perform abstract reasoning and generalize from minimal examples. Unlike traditional machine learning tasks that often rely on large datasets and pattern recognition, ARC focuses on measuring a form of fluid intelligence akin to human problem-solving abilities.

\subsubsection{Overview of ARC}

ARC consists of a collection of tasks presented as grid-based puzzles. Each task includes a small set of demonstration examples, where each example pairs an input grid with a corresponding output grid. The grids contain at most 10 symbols, visualized with colors, and can vary in size from $1 \times 1$ up to $30 \times 30$ cells. The objective is to infer the underlying rule that maps inputs to outputs and then apply this rule to new, unseen input grids to produce the correct output grids.

\begin{figure}[htbp]
    \centering
    \includegraphics[width=0.8\textwidth]{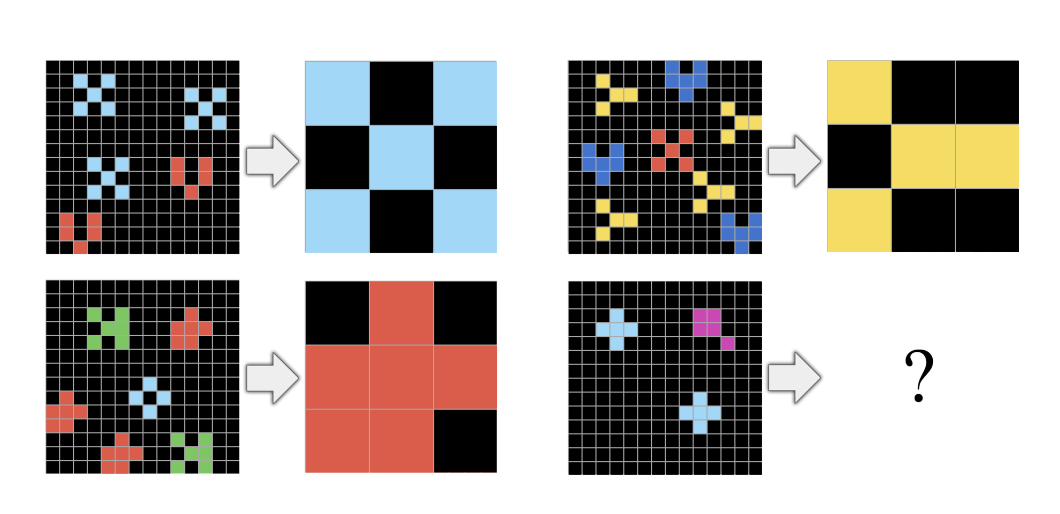}
    \caption{A task where the implicit goal is to count unique objects and select the object that
appears the most times (the actual task has more demonstration pairs than these three).}
    \label{fig:arc-example}
\end{figure}

An example of an ARC task is shown in Figure~\ref{fig:arc-example}. In this task, the input grids and their corresponding output grids are provided, and the goal is to deduce the transformation that relates them and apply it to a new input grid to produce the correct output.

A key characteristic of ARC tasks is that they are intentionally diverse and abstract, covering a wide range of reasoning types without relying on language, specific domain knowledge, or extensive training data. This design aims to mimic the kind of innate prior knowledge humans have -- such as object recognition, counting, and basic geometry --which allows people to solve novel problems with minimal examples.

\subsubsection{Challenges Presented by ARC}

ARC is particularly challenging for artificial intelligence systems for several reasons:

\begin{itemize}
    \item \textbf{Minimal Training Examples}: With only a few examples per task, models cannot rely on statistical learning from large datasets and must instead perform higher-level reasoning.
    \item \textbf{Abstract and Diverse Tasks}: The tasks cover a broad spectrum of reasoning types, including pattern recognition, spatial transformations, and logical operations, without a consistent structure across tasks.
    \item \textbf{Emphasis on Generalization}: Models must generalize from the small set of examples to correctly solve new instances, reflecting an understanding of the underlying principles rather than memorization.
    \item \textbf{Lack of Domain-Specific Knowledge}: The tasks avoid requiring external knowledge, focusing instead on core cognitive abilities such as object permanence, symmetry, and basic arithmetic.
\end{itemize}

These challenges make ARC an effective benchmark for assessing an AI system's capacity for general reasoning and adaptability, setting a high bar for models that traditionally excel in pattern recognition but struggle with abstract problem-solving.

\subsubsection{ARC-AGI-1 and ARC-AGI-2}

The ARC-AGI benchmark was developed to evaluate progress toward artificial general intelligence (AGI) through tasks that test abstraction, reasoning, and generalization. ARC-AGI-1 comprises the initial set of tasks designed for this purpose, whereas ARC-AGI-2, introduced in 2025, represents a refined and more challenging version. It preserves the original grid-based format but introduces additional elements to stimulate further research toward general intelligence. In particular, ARC-AGI-2 tasks are on average larger in size (up to $30 \times 30$), involve a greater number of colors (up to ten per task), and require the sequential application of multiple compositional rules. These tasks are also designed to be more resistant to brute-force or domain-specific language (DSL) approaches. While ARC-AGI-2 is more difficult for humans as well, each task has been validated to ensure that at least two human participants can solve it, maintaining the benchmark’s relevance as a measure of human-comparable reasoning.

\subsubsection{Significance of ARC in AI Research}

ARC serves as a benchmark for exploring the limitations of current AI approaches and pushing the development of models that can reason more like humans. By requiring systems to solve tasks with minimal data and without domain-specific training, ARC highlights the gap between human cognitive abilities and artificial intelligence.

For researchers, ARC offers an opportunity to develop and evaluate methods that incorporate aspects of human-like reasoning, such as:

\begin{itemize}
    \item \textbf{Core Knowledge Priors}: Leveraging innate principles that humans use to understand the world, like object continuity and basic geometry.
    \item \textbf{Meta-Learning and Adaptation}: Designing models that can quickly adapt to new tasks using limited information.
    \item \textbf{Symbolic and Relational Reasoning}: Integrating symbolic manipulation and understanding of relationships between objects, beyond mere pattern recognition.
\end{itemize}

Our work engages with these challenges by exploring how large language models can be trained and adapted to perform abstract reasoning on ARC tasks, contributing to the broader goal of developing AI systems with more generalized problem-solving capabilities.

\subsection{Related Work}

A variety of approaches to solving the Abstract Reasoning Corpus (ARC) have been tried. We provide an overview, first of some augmented datasets, and then of notable ARC solvers.

Given the limited size of the original ARC dataset, it is natural to explore data augmentation. In \cite{Hodel2024}, Michael Hodel presents a DSL-based approach capable of generating a large number of examples (up to 10,000) for each ARC training task, with the stated goal of facilitating experiments by enabling more meaningful model comparisons. In \cite{Moskvichev2023}, the Concept-ARC dataset is introduced, consisting of 160 new hand-crafted tasks grouped according to 16 ``basic spatial and semantic concepts'' selected by the authors. This is useful for measuring solver performance in terms of concept coverage. However, to train large deep learning models on ARC-like data, large-scale data augmentation seems necessary. Very recently, the authors of \cite{Li2024} produced datasets of up to 400,000 synthetic problems by starting from 100 seeds, which are natural-language descriptions and code for a 100-task subset of the original ARC dataset. Their method for task generation roughly consists of prompting a Large Language Model (LLM) for a novel description based on the seed descriptions, retrieving seeds with similar descriptions, and prompting the LLM again with the code corresponding to these seeds to produce the code for the novel task. This process provides both a function for generating appropriate input grids and the task transformation.

Some of the most notable early ARC solvers were based on building a domain-specific language (DSL) and searching it to find candidate test solutions. The winning solution \cite{Wind2020} of the 2020 Kaggle ARC challenge is one such example, using a DSL with 142 unary functions derived from 42 functions written by hand. These functions are composed and applied to the input grids to produce a directed acyclic graph of grids, which are then greedily stacked to obtain prediction candidates. Other DSL-based solvers include \cite{Ainooson2023}, \cite{Fischer2020}, and \cite{Xu2022}.

A different approach is neuro-symbolic AI. In \cite{Ellis2020}, Kevin Ellis et al. introduce DreamCoder, an algorithm for solving program-writing tasks that alternates between a waking phase and two sleeping phases, combining DSL and neural networks. During the waking phase, the DSL is used to write programs, and in the ``abstraction sleep'' phase, the DSL is updated with new primitives. Additionally, during ``dreaming sleep,'' a recognition model is trained to guide the program search in the waking phase. DreamCoder was adapted for ARC in \cite{Bober2024}, building on the work of \cite{Alford2021}, where the authors introduce their own PeARL DSL for ARC with 77 primitives.

Another example of a program synthesis ARC solver with both a program-writing phase and a learning phase is CodeIt (\cite{Butt2024}), short for Code Iteration. The authors claim that CodeIt is less computationally intensive than DreamCoder.

Most of the early deep learning-focused approaches either showed very low performance or failed to generalize well beyond a chosen evaluation set. However, they still provide some useful and unique insights. For example, the idea in \cite{Thoms2023} is to use an autoencoder to create vector-space embeddings of grids. The test output is then predicted by adding a training pair difference vector to the vector representing the test input. This solver scored 2\% on the 400 public evaluation tasks of ARC-AGI-1 when using three attempts (no score available on ARC-AGI-2).

The combination of a differentiable neural computer with a transformer in \cite{Kolev2020} achieves high accuracy (78.8\%) on small grids (less than 10×10), but it remains unclear how well this method generalizes to larger grids.

In \cite{Park2023}, the authors train a Decision Transformer to mimic human problem-solving, using the tools from \cite{Kim2022} to gather human solutions for ARC-Mini. They employ an interesting method for object detection via pixel clustering, but their evaluation is limited to four selected ARC tasks.

Several sources have documented attempts to prompt large language models to solve ARC tasks directly, without any fine-tuning. For instance, the authors of \cite{Mirchandani2023} and \cite{Gendron2023} compare the performance of various LLMs on ARC, with the best models reaching about 10\% on the 400 public evaluation tasks of ARC-AGI-1. One notable observation from \cite{Mirchandani2023} is that the models could still solve tasks when the numbers corresponding to colors were replaced by tokens randomly sampled from the vocabulary. According to them, this even extends to token embedding vectors not seen by the model during training (see Appendix A.3 in the source). With respect to fine-tuning, Appendix B.5 of \cite{Gendron2023} mentions an attempt with LLaMA-7B and LLaMA2-7B, without any data augmentation, that resulted in an almost doubling of their score on ARC, but still below the 12\% achieved by GPT-4 on the 400 public evaluation tasks. Another similar paper is \cite{Xu2023}, which highlights the importance of in-context prompting and object-based representations. They found that using their previous work on abstract reasoning with graph abstractions \cite{Xu2022} to describe the objects on the grids noticedably improved performance. Interestingly, however, including relations between objects (for example, when two objects lie on the same horizontal line) did not lead to further improvement.

The best-performing solution involving direct prompting reached 50\% on the public evaluation set and is detailed in \cite{Greenblatt2024}. Their main idea was to prompt GPT-4o to generate around 5000 candidate solution Python programs for a given task, then to revise the most promising ones. They use a few-shot prompt, providing step-by-step demonstrations for both task solving and code revision. Another important idea in their solution is to include additional text representations of the grids, utilizing connected components and the differences between train outputs and inputs when possible. However, this approach is not suitable for Kaggle competition, because of runtime compute constraints and the use of a closed-source model.

After the conclusion of the Arc Prize 2024 Kaggle competition, the second- and third-place teams publicly shared their solutions. A key insight is that with the right setup and implementation, particularly by taking advantage of data augmentation and test-time fine-tuning, LLMs can achieve significantly higher performance on ARC (in fact, setting the SOTA). The second-place team \cite{Franzen2024} achieved a score of 72.5 on 100 public evaluation tasks and 56.5\% on the hidden Kaggle dataset with a single solver, fine-tuning Mistral-NeMo-Minitron-8B-Base. The fine-tuning process was conducted in two stages: first on public ARC-like datasets and then on all 100 hidden tasks. Their approach also incorporated a DFS-based sampling method in inference and a strategy for selecting predictions among candidates. This strategy involved aggregating the probabilities assigned to candidates by the model across augmentations under the assumption that the correct solution would exhibit the highest stability across augmentations.

The third place solution \cite{Barbadillo2024} achieved a score of 40\% on the hidden Kaggle data set by fine-tuning an LLM both offline and online, with some notable differences compared to the second place solution. The model used, Qwen2.5-0.5B-Instruct, is significantly smaller than the one utilized in \cite{Franzen2024}. During offline fine-tuning on public data, the model was trained not only to predict test outputs, but also to generate new inputs for a given task. Furthermore, on-line/test-time fine-tuning was performed separately for each hidden task. For each test, a large number of candidates (96) were generated and the predictions were aggregated using a majority vote. This solution was ensembled with Icecuber's solver from \cite{Wind2020}, while the fine-tuned LLM alone achieved a score of 33\%.

\subsection{Structure of the Paper}

The remainder of this paper follows the logical progression of our approach, from conceptual foundations to empirical validation. 
Section~\ref{sec:our_method} outlines the complete pipeline, including model design, data strategy, and inference workflow. 
Section~\ref{sec:data} details the datasets employed, while Section~\ref{sec:augmentations} introduces our augmentation techniques based on symmetries, cellular automata, and traversals. 
Section~\ref{sec:longt5} presents the LongT5-based architecture, and Sections~\ref{sec:offline_training} and~\ref{sec:inference} describe the training and inference procedures, including test-time training and candidate ranking. 
Section~\ref{sec:monitoring} covers experiment tracking and monitoring, and Section~\ref{sec:results} reports empirical results. 
Finally, Section~\ref{sec:ablations} provides ablation studies, and Section~\ref{sec:conclusion} concludes with a discussion of implications for systematic generalization.

\section{Our Method -- Overview}
\label{sec:our_method}
\subsection{High-Level Description of the Pipeline}

\begin{figure}[htbp]
    \centering
    \includegraphics[width=\linewidth]{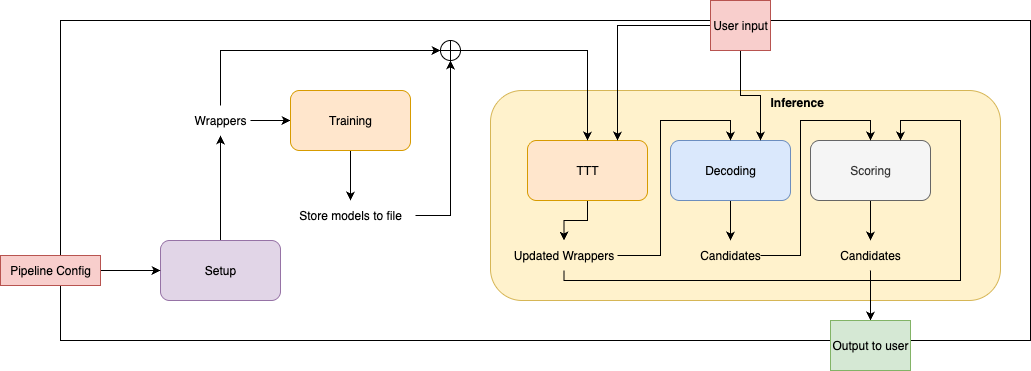}
    \caption{Overview of the ARC-AGI pipeline. The system consists of setup, training, and inference stages. Inference integrates Test-Time Training (TTT), decoding and scoring components. The user input, in the case of ARC-AGI Kaggle environment, corresponds to loading the test set from files.}
    \label{fig:pipeline}
\end{figure}

Our ARC-AGI solution is structured as a modular and extensible pipeline designed to support large-scale experimentation while maintaining adaptability to individual ARC tasks. 

As illustrated in Figure~\ref{fig:pipeline} operates in three main stages: 
\begin{itemize}
    \item The \textbf{Setup} stage initializes the environment, downloads the dataset and the model and initializes the \textbf{Wrappers}, which encapsulate model interfaces and task-specific utilities.
    \item The \textbf{Training} stage fine-tunes a base model using the datasets ingested in the setup stages. Trained weights are stored to disk for reuse during inference.  
    \item The \textbf{Inference} specializes the trained model to a specific task through lightweight adaptation, decoding, and scoring.
\end{itemize}.

The \textbf{Inference} process then operates over new task through four major components:

\begin{itemize}
    \item \textbf{Test-Time Training (TTT)} -- a lightweight adaptation step applied to the base model on a single ARC task. It performs a LoRA fine-tuning with low rank on augmented data generated from the given task, enabling the model to specialize to the new task logic without overfitting it.
    \item \textbf{Decoding} -- the adapted model generates multiple candidate solutions using a standard beam search. Each candidate represents the output grid corresponding to the test input.
    \item \textbf{Filtering} -- a set of white-box heuristics prunes inconsistent or implausible solutions based on shape consistency, color constraints, and object-level coherence. This step is run after the decoding step.
    \item \textbf{Scoring} -- the remaining candidates are evaluated via a ranking function that performs one forward pass of the model on the original and transformed variants of the task. This exploits ARC-AGI's inherent spatial symmetries (e.g. rotations, reflections, and transpositions) to compute a per-pixel confidence score for each candidate.
\end{itemize}

The one (or more) candidate with the highest aggregated score is selected as the final output, which is then returned to the user. In the ARC-AGI specific case, the score is computed on two candidate solutions.

The pipeline is defined by a human-readable \texttt{.yaml} file with a \textbf{Pipeline Configuration} that defines all parameters controlling data and model ingestion, training, inference, and monitoring.

\subsection{Design Principles}

Our method follows four guiding design principles:

\begin{enumerate}
    \item \textbf{Scalability} -- The pipeline supports distributed training and inference across multiple compute nodes, enabling large-scale experimentation with various model architectures and adaptation strategies.
    \item \textbf{Modularity} -- Each component (Setup, Training, TTT, Decoding, Scoring) is designed as an independent module with well-defined interfaces. This allows for easy substitution and extension of individual parts (e.g., alternative decoders or scoring mechanisms).
    \item \textbf{Adaptability} -- The pipeline per se is generic, and with relatively minor modifications, can be generalized to support any type of transformer model.
    \item \textbf{Reproducibility} -- There is always tension between the advantages of reproducible results and the stochastic nature of LLMs. While stochasticity is used all over the pipeline to sample transformations used for augmentation, to shuffle the dataset, and in candidate generation, we took great care of fixing the seeds to have reproducible results. Ultimately, within a tolerance of 1\%, our results are deterministic.
\end{enumerate}

\section{Data}
\label{sec:data}
\subsection{ARC Public Training Sets}

We conducted an extensive search for publicly available ARC-AGI datasets beyond those 1240 official examples provided by the organizers of the competition.

In Table~\ref{tab:open_source_data}, you will find the list of all public datasets we have found. Most of these datasets contain less than 1000 examples, with the exception of the BARC dataset. Despite its large size, we excluded BARC due to its high level of noise and the presence of tasks that lack clear underlying human-interpretable logic. While noisy data can sometimes improve robustness — as explored in the context of domain randomization \cite{tobin2017domain} — we found that the ARC-AGI benchmark demands a higher degree of precision, making such trade-offs less suitable for our objectives.

\begin{table}[htbp]
\centering
\renewcommand{\arraystretch}{1.3}
\begin{tabular}{l|p{8cm}|l}
\textbf{Dataset} & \textbf{Description} & \textbf{URL} \\\hline
ARC-synthetic-extend & Synthetic extension of the original ARC dataset with procedurally generated tasks that preserve ARC-style reasoning patterns. & \href{https://github.com/frankaging/ARC_synthetic_extend}{link} \\
ConceptARC & A curated collection of concept-driven ARC tasks designed to isolate and test specific reasoning concepts such as counting, reflection, and symmetry. & \href{https://github.com/victorvikram/ConceptARC/tree/main}{link} \\
Mini-ARC & A reduced version of the ARC dataset with simplified tasks aimed at educational and quick experimentation purposes. & \href{https://github.com/KSB21ST/MINI-ARC}{link}\\
PQA & Procedurally Generated Question-Answer (PQA) dataset inspired by ARC, focusing on abstract pattern recognition and visual analogy. & \href{https://github.com/qugank/pqa.github.io}{link} \\
Sequence-ARC & Sequence-based ARC-style tasks emphasizing temporal or ordered reasoning across grid transformations. & \href{https://github.com/seedling123/Sequence_ARC}{link} \\
Sort-of-ARC & Simplified ARC-like dataset where tasks are grouped by reasoning type, providing structure for curriculum learning. & \href{https://github.com/neoneye/arc-dataset-collection/tree/main/dataset/Sort-of-ARC}{link} \\
arc-community & Community-maintained collection of ARC-style puzzles contributed by researchers and hobbyists. & \href{https://github.com/arc-community/arc}{link} \\
arc-dataset-diva & Large-scale synthetic ARC dataset automatically generated with diverse visual and structural transformations. & \href{https://github.com/neoneye/arc-dataset-diva}{link} \\
arc-dataset-tama & Expanded ARC-like dataset containing additional procedurally created puzzles emphasizing color and object relations. & \href{https://github.com/neoneye/arc-dataset-tama}{link} \\
nosound & Small ARC-compatible dataset hosted on Kaggle, featuring unique manually created puzzles. & \href{https://www.kaggle.com/datasets/zaharch/arc-nosound-tasks}{link} \\
BARC & A massive benchmark dataset (400k tasks) inspired by ARC, used for large-scale pretraining and evaluation of pattern reasoning models. & \href{https://github.com/xu3kev/BARC}{link} \\\hline
\textbf{Total} & 11 open-source ARC-style datasets &  \\
\end{tabular}
\caption{\label{tab:open_source_data}Open-source ARC-style datasets used in this work. Each dataset extends or complements the original ARC challenge with procedurally generated, concept-specific, or community-created tasks.}
\end{table}

\subsection{Test Set}
\label{sec:data:human}
\begin{figure}[htbp]
    \centering
    \includegraphics[width=0.9\textwidth]{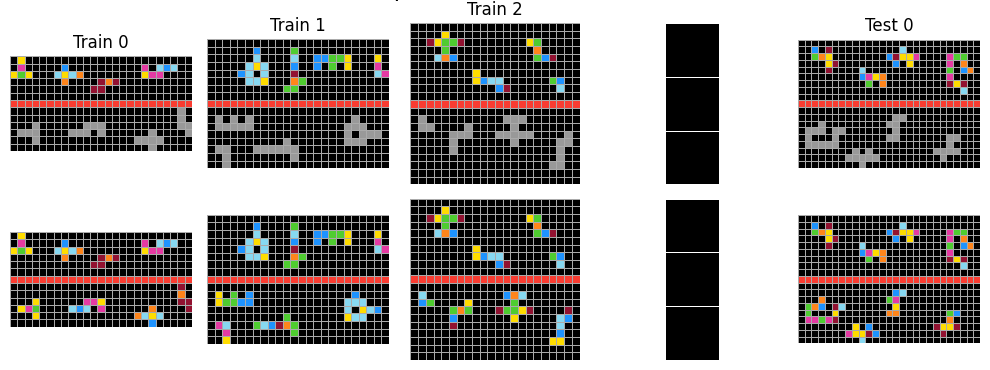} 
    \caption{This is an example of a task made by our team: the logic is similar but not identical to any task in the ARC-AGI public dataset.}
    \label{fig:human-made}
\end{figure}

We manually designed 177 new tasks, balancing diversity and feasibility within our available annotation resources. Since the Kaggle test dataset remains undisclosed to the date of writing, all the results of this paper are computed on those tasks. Figure~\ref{fig:human-made} shows an example of a human-made task.

\subsection{Encoding}
\label{sec:encoding}

A key aspect of applying language models to ARC-AGI is the design of an effective encoding strategy. Since large language models are fundamentally sequence processors, ARC grid tasks must be converted into a tokenized textual format. However, this presents two main challenges:

\paragraph{Context size limitations.}
Transformers operate within constrained context windows, and long input sequences significantly increase memory usage and inference latency while degrading long-sequence performance. A compact tokenization scheme is therefore essential.

\paragraph{Tokenization artifacts.}
Standard byte-pair encoders (BPEs) used in LLMs frequently merge digits into multi-digit tokens (e.g.\ ``112'' $\rightarrow$ ``11'', ``2''). Such merges are harmful in ARC, where each digit represents a distinct color class. To avoid this, we \textbf{fully controlled the tokenizer} by reducing the vocabulary from $32{,}128$ tokens to a compact set of \textbf{125 atomic tokens}, with one token per visual symbol.

\begin{figure}[htbp]
    \centering
    \includegraphics[width=0.8\textwidth]{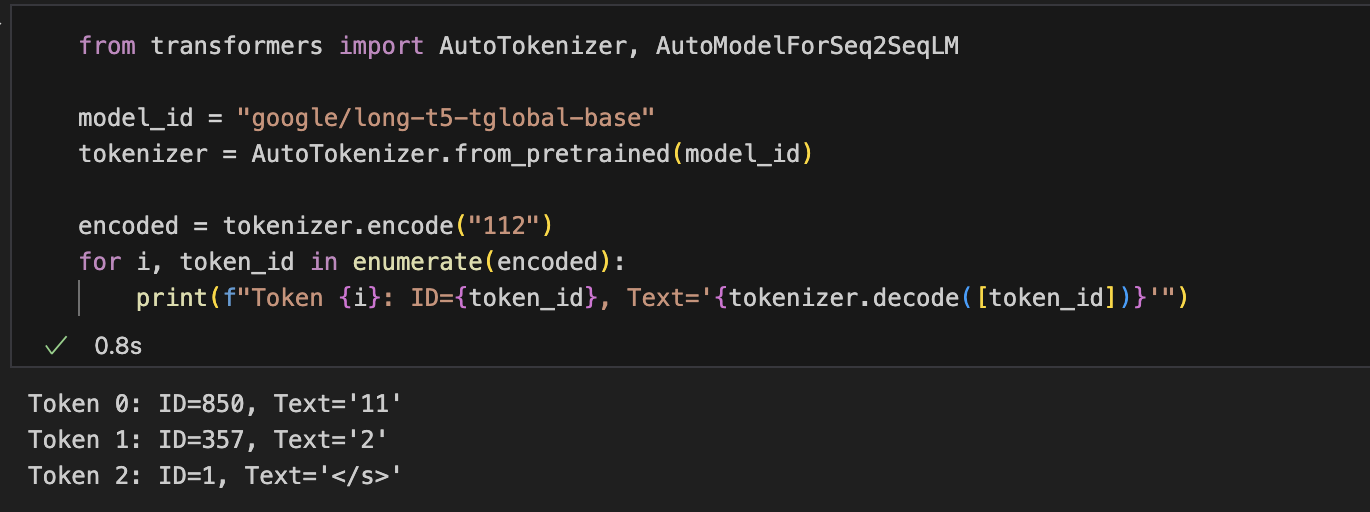} 
    \caption{An example of a tokenization artifact.}
    \label{fig:token}
\end{figure}

\paragraph{Vocabulary Design.}
Our custom vocabulary is composed of structural delimiters, color tokens and extra tokens for denoising (see \ref{sec:training_to_understand}):
\begin{itemize}
    \item \textbf{Example boundaries:} \texttt{<|start\_example|>}, \texttt{<|end\_example|>}
    \item \textbf{Input boundaries:} \texttt{<|start\_input|>}, \texttt{<|end\_input|>}
    \item \textbf{Output boundaries:} \texttt{<|start\_output|>}, \texttt{<|end\_output|>}
    \item \textbf{Row boundaries:} \texttt{<|start\_row|>}, \texttt{<|end\_row|>}
    \item \textbf{Color tokens:} \texttt{<|color\_0|>} \dots \texttt{<|color\_9|>}
    \item \textbf{End of sequence token:} \texttt{</s>}
    \item \textbf{Padding token:} \texttt{<pad>}
    \item \textbf{Traversals tokens:} \texttt{<|row\_by\_row|>}, \texttt{<|snake|>}
    \item \textbf{UL2 tokens:} \texttt{<task\_id\_S>}, \texttt{<task\_id\_X>}, \texttt{<task\_id\_R>}, and \texttt{<extra\_id\_0>}, \texttt{<extra\_id\_1>}, \dots \texttt{<extra\_id\_99>}

\end{itemize}

\paragraph{Example Encoding.}
Given the example:
\[
(\text{input}, \text{output}) =
\big( [[1,2],[3,4]],\; [[5,6]] \big),
\]
its serialized textual representation becomes:
\begin{verbatim}
<|start_example|><|start_input|>
  <|start_row|><|color_1|><|color_2|><|end_row|>
  <|start_row|><|color_3|><|color_4|><|end_row|>
<|end_input|>
<|start_output|>
  <|start_row|><|color_5|><|color_6|><|end_row|>
<|end_output|><|end_example|>
\end{verbatim}

This compact, semantically aligned encoding significantly reduced vocabulary size and model parameters\footnote{We do not use weight-tying, hence the number of parameters is reduced by $2 \times (32128 - 22) \times 768 = 49{,}314{,}816$.} (from $250$M to $201$M, a $20\%$ reduction), with improved trainability.

\section{Injecting Prior Knowledge}
\label{sec:augmentations}
A key ingredient of our approach to solving ARC-AGI tasks is the incorporation of \emph{prior knowledge}. Priors encode inductive biases about the structure of tasks and constrain the hypothesis space explored by the model. In the context of ARC, this is particularly important because the training distribution is extremely sparse and heterogeneous, while the test tasks are intentionally adversarial to pure statistical learning. 
We employ three complementary strategies for injecting priors into the system:
\begin{enumerate}
    \item \textbf{Architectural priors via equivariance}, encoded directly in the neural architecture. While we explored this direction, it was not part of our final system.
    \item \textbf{Filtering priors} -  symbolic validation rules that prune invalid or inconsistent candidate solutions.
    \item \textbf{Data augmentation priors}, obtained by applying symmetry transformations consistent with the ARC domain, promoting generalization by discouraging brittle pixel-level memorization and enforcing rule-based reasoning.
\end{enumerate}
Each of these strategies is described in the following subsections.
Before detailing each type of prior, we introduce the following notation:
\begin{itemize}
    \item $O$ denotes the test output grid
    \item $O_i$ denotes an output grid in the training set of a task
    \item $I$ denotes the test input grid 
    \item $I_i$ denotes an input grid in the training set of a task
    \item $X$ denotes a generic grid and $x_i$ denotes its $i$-th token
\end{itemize}

\paragraph{Structural positional equivariance.}
Standard Transformers discard spatial grid structure. Although we tried using 2D-aware positional encoding, the results did not improve. Furthermore, we decided not to use padding, as it would increase the input length too much, making the computations highly inefficient. On the other hand, we believe that the augmentations we perform do instill 2D knowledge to the LLMs (however, we did not test this feeling beyond kaggle).

\paragraph{Conclusion.} We did not use equivariant architectures as we ultimately decided to adopt the longT5 encoder-decoder model with the Transient Global attention mechanism (ref \ref{sec:longt5}). However, initial experiments with smaller equivariant architectures showed encouraging results.
% ======================
\subsection{Filtering via Symbolic Priors}
\label{sec:filtering}
Priors can also be enforced \emph{symbolically} at inference time, rather than being embedded directly in the network architecture. In the ARC domain, many structural regularities can be expressed as logical constraints that candidate solutions must satisfy. This motivates a \emph{filtering layer} that evaluates symbolic consistency rules and discards invalid candidate outputs.

Given a candidate output $\hat{O}$ for a test input $I$, we define a filtering function
\[
    \mathcal{F}(\hat{O}, I, \mathcal{D}_{\mathrm{train}}) \in \{0,1\},
\]
which returns $1$ if $\hat{O}$ is consistent with the prior knowledge extracted from the training pairs $\mathcal{D}_{\mathrm{train}}$, and $0$ otherwise. We include three main filtering functions:

\paragraph{Color consistency filter.}
In the majority of ARC tasks, the output uses only colors that appear in the task. Let $C(X)$ denote the set of colors in grid $X$. Valid candidates must satisfy:
\[
    C(\hat{O}) \subseteq \left(\bigcup_{(I_i, O_i) \in \mathcal{D}_{\mathrm{train}}} C(I_i) \cup C(O_i) \right) \cup I.
\]
\textbf{Exception.} If exactly nine colors $\{0, \dots, 9\} \setminus \{c\}$ appear in all examples and test input, it may be implied that the remaining $c$ will appear in the test output. We never saw such task: if it exists, we will currently filter it out.

\paragraph{Grid size consistency filter.}
We use two types of size filters:
\begin{itemize}
\item If all output grids in the demonstrations share the same height and width, i.e.
\[
    |O_i| = (h, w) \quad \forall i,
\]
then the test output must have the same dimensions:
\[
    |\hat{O}| = (h, w).
\]
\item If all output grids have the same ratio to their respective input grids, i.e.
    \[
    |O_i| = k \cdot |I_i|  \quad \forall i,
\]
for a fixed $ k \in \mathbb{N}$,
then the test output must preserve the same ratio with respect to its test input:
\[
    |\hat{O}| = k \cdot |I|.
\]
\end{itemize}

These filters eliminate a few candidates, usually keeping the correct ones (there is only 1\% degradation according to our internal tests), lightening the candidate ranking step.

\paragraph{Inclusion filtering.}
Many ARC transformations preserve a containment relationship between input and output. For example, in tasks where the output is always a subset of the input (e.g.\ object extraction or masking), the same holds for the test case. Conversely, in constructive tasks, the input is contained within the output. We formalize two symmetric priors:

\[
\textbf{Output-Contained: } \forall (I_i,O_i),\; O_i \subseteq I_i \Rightarrow \hat{O} \subseteq I,
\]
\[
\textbf{Input-Contained: } \forall (I_i,O_i),\; I_i \subseteq O_i \Rightarrow I \subseteq \hat{O}.
\]

These relations are also enforced \emph{up to rigid transformations}: if $T$ is a member of the dihedral group $D_4$, we allow containment checks under $T(I)$ and $T(O)$ to capture rotated or flipped outputs:
\[
O_i \subseteq T(I_i) \;\Rightarrow\; \hat{O} \subseteq T(I).
\]

% ======================
\subsection{Data Augmentation via Symmetry Priors}
\label{sec:data_aug}

A third way to inject priors is to expand the training data using symmetries from the ARC domain. Many tasks are invariant under geometric transformations that preserve adjacency and local patterns. We augment all the examples using the symmetry group
\[
    \mathcal{G} = D_4 = \{\text{identity},\ 90^\circ,\ 180^\circ,\ 270^\circ\ \text{rotations},\ \text{horizontal/vertical/diagonal flips}\}.
\]
For each training pair $(I_i, O_i)$, we generate its full orbit under $\mathcal{G}$:
\[
    \mathcal{O}(I_i, O_i) = \{(g \cdot I_i,\ g \cdot O_i) : g \in \mathcal{G}\}.
\]
This provides an $8 \times$ effective expansion of the dataset while preserving semantic consistency. In addition, augmentation regularizes the model and prevents it from learning spurious orientation-sensitive heuristics.

These same symmetries shall be applied to the test input as well, and the inverse transform shall be applied to the generated test output, to make it comparable with the rest of the candidates.

\subsection{Data Augmentation via computer-vision-like transformation}
\label{subsec:data_aug_cv_like}

\begin{figure}[hbpt]
    \centering
    \begin{subfigure}{0.22\textwidth}
    \centering
    \includegraphics[height=5.5cm]{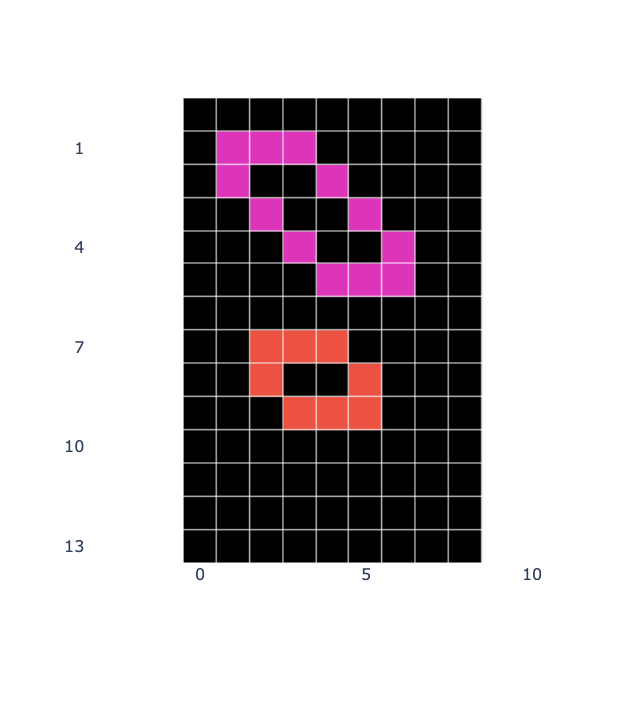}
    \caption{\label{fig:cvlike_aug_1}}
    \end{subfigure}
    \begin{subfigure}{0.22\textwidth}
    \centering
    \includegraphics[height=5.5cm]{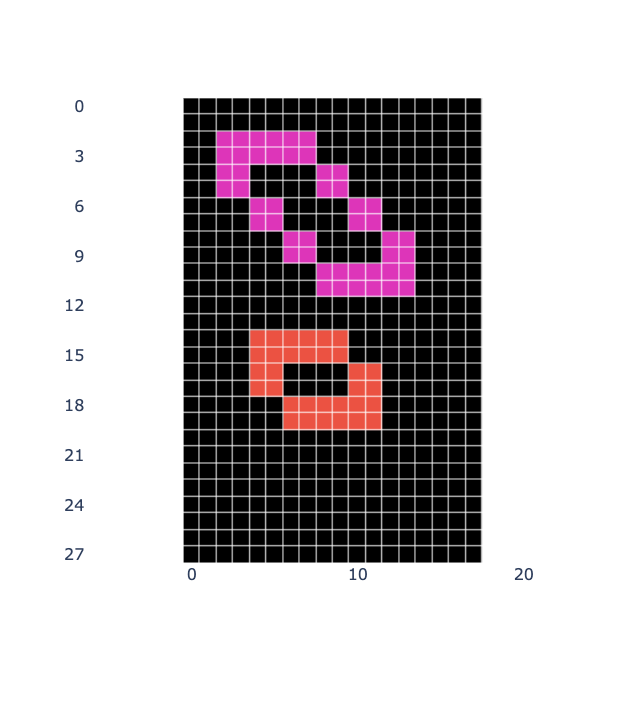}
    \caption{\label{fig:cvlike_aug_2}}
    \end{subfigure}
    \begin{subfigure}{0.22\textwidth}
    \centering
    \includegraphics[height=5.5cm]{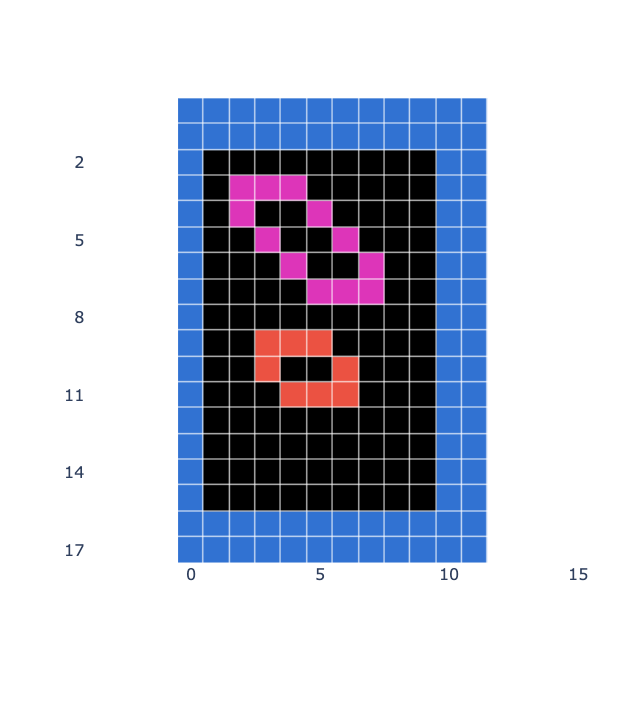}
    \caption{\label{fig:cvlike_aug_3}}
    \end{subfigure}
    \begin{subfigure}{0.22\textwidth}
    \centering
    \includegraphics[height=5.5cm]{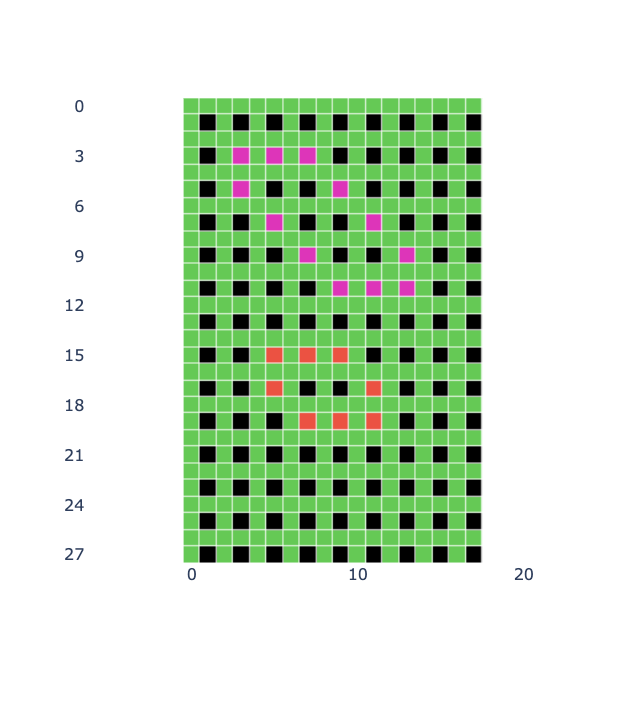}
    \caption{\label{fig:cvlike_aug_4}}
    \end{subfigure}
\caption{\label{fig:cvlike_aug} Examples computer-vision-like augmentations applied to the first input grid of task 025d127b; \ref{fig:cvlike_aug_1} original grid; \ref{fig:cvlike_aug_2} upscale 2x, directions=$both$; \ref{fig:cvlike_aug_3} adding frame; \ref{fig:cvlike_aug_4} adding metagrid, $s=1$, directions=$both$.}
\end{figure}

To improve the model spatial understanding of the grid, three additional transformations have been used to augment the data during the offline training, all inspired by classical computer vision techniques. Each transformation is randomly applied either to the input or output only, or to both of them. Here follows a detailed description, while an example can be seen in figure \ref{fig:cvlike_aug}:
\begin{itemize}
    \item \textbf{upscale}. The upscale factor is randomly sampled among $[\times2, \times3]$ and the applied direction among $[row, column, both]$. The grids are up-scaled accordingly.
    \item \textbf{framing}. Framing is formally equivalent to a padding operation. The frame color $c$ is sampled from the complementary colors (i.e., the colors not used in any of the grids of the task), and the padding values for the four edges are sampled independently. 
    \item \textbf{metagrid}. We sample the direction $d$ among $[row, column, both]$, the step $s$ among $[1,2,3]$, and the color $c$ among the available colors. E.g., if $d=row$, we add an extra row of color $c$ every $s$ rows, if $d=column$, we add an extra column every $s$ columns, if $d=both$, we add both rows and columns.
\end{itemize}

The core reason behind these transformations is to allow the model to focus less on the token level (i.e., the single pixel of the grid) and more on generalizing concepts of objects and their spatial relations. For example, adding a metagrid pushes the pixels further away from each other: new pixels are inserted between them leaving the rule unaltered, thus forcing the model to focus more on the rule and to be resilient to long distances and information sparsity. At the same time, upscaling also improves grid regions separation. In addition, the objects of the grids are still preserve most of their properties (except size), being represented with a different amount of pixels: this forces the model to learn the concept of ``objectness''\cite{Spelke2007} more robustly.

\subsection{Traversal Priors as Representation Augmentation}
\label{sec:traversal}

In addition to symmetry-based priors, we introduce a complementary family of \emph{representation priors} based on alternative \emph{grid traversals}. Unlike symmetries, these transformations do not preserve the 2D spatial structure exactly, but instead provide different 1D serializations of the grid that expose diverse algorithmic patterns to the model. The key motivation is that ARC tasks are inherently \emph{symbolic} and \emph{rule-based}. By traversing the same grid in multiple structured ways, we encourage the model to learn the transformation rule itself rather than overfit to any specific representation of the grid.

Formally, let $x \in \{0,\dots,9\}^{H \times W}$ be a grid of size $H \times W$. A traversal transformation is a bijective mapping
\[
    \tau: \{1,\dots,HW\} \longrightarrow \{(i,j) \mid i \in \{1,\dots,H\},\, j \in \{1,\dots,W\}\},
\]
which, when applied to the full grid $X=(X_1, X_2, \dots, X_{HW})$, simply defines an ordered sequence of grid positions:
\[
    \tau(X) = \big(X_{\tau(1)}, X_{\tau(2)}, \dots, X_{\tau(HW)}\big).
\]
We explore the following traversals:

\paragraph{(1) Row-by-row traversal (Left-to-Right).}
This is the standard ordering:
\[
    \tau_{\mathrm{row}}(k) = \bigg( \big\lfloor \tfrac{k-1}{W} \big\rfloor + 1,\ (k-1 \bmod W) + 1 \bigg).
\]
This preserves the natural reading order of the grid and serves as the base representation.

\paragraph{(2) Snake (zig-zag) traversal.}
To reduce directional bias, we introduce alternating directionality across rows:
\[
\tau_{\mathrm{snake}}(k) =
\begin{cases}
    \big(i,\ j\big) & \text{if } i \text{ is odd}, \\[4pt]
    \big(i,\ W-j+1\big) & \text{if } i \text{ is even},
\end{cases}
\]
where $i = \big\lfloor \tfrac{k-1}{W} \big\rfloor + 1$ and $j = (k-1 \bmod W) + 1$. This traversal avoids a bias toward strictly left-to-right rule detection and makes patterns detectable under alternating spatial adjacencies.

Other traversals, such as diagonal traversal, spiral traversal toward the grid center or Hilbert curve ordering, may also provide representational diversity by exposing locality at multiple scales. Although such traversals were not used in our final pipeline, they represent promising directions for future work in structured representation priors.

\paragraph{Impact on learning}
Given a training pair $(I_k,O_k)$, traversal augmentation generates new training examples $(\tau(I_k), \tau(O_k))$ across different $\tau \in \mathcal{T}$, where $\mathcal{T}$ is the set of allowed traversals. We will use these novel examples to create new tasks: we do not leverage them for in-task augmentation, as we noticed that the increase of the context length usually has bad impacts on the model's performance. Unlike geometric symmetries, these traversals deliberately alter the representation while leaving the underlying mapping invariant. The goal is to teach the model to discover the underlying transformation rule $f$ rather than exploit an incidental spatial encoding.

\[
    \text{If } O_k = f(I_k), \text{ then training with traversals enforces } \tau(O_k) = f(\tau(I_k)), \ \forall k.
\]

This approach aligns with the idea that \emph{multiple representations promote rule abstraction}. By encountering the same transformation across different serializations, the model becomes robust to superficial input structure and learns a deeper task representation.

Beyond traversal priors, we also leveraged \emph{automata-based augmentations}, where additional randomized but structure-preserving transformations are used to inject controlled noise around each rule. These methods are explored in the next section.

\subsubsection{Examples}
\label{sec:traversal:examples}

\paragraph{Traversal Tokens.}
To enable flexible grid serialization, we introduced two traversal mode tokens:
\begin{itemize}
    \item \texttt{<|row\_by\_row|>} (default): left-to-right, top-to-bottom.
    \item \texttt{<|snake|>}: alternating direction per row.
\end{itemize}
These tokens are prepended to each prompt, allowing the model to condition its reasoning on the traversal perspective.

\paragraph{Snake Traversal Example.}
For the same example of Section~\ref{sec:encoding}, the snake traversal representation becomes:
\begin{verbatim}
<|snake|><|start_example|><|start_input|>
  <|start_row|><|color_1|><|color_2|><|end_row|>
  <|end_row|><|color_4|><|color_3|><|start_row|>
<|end_input|>
<|start_output|>
  <|start_row|><|color_5|><|color_6|><|end_row|>
<|end_output|><|end_example|>
\end{verbatim}

\paragraph{Results.}
Training with traversal augmentation improved robustness to decoding uncertainty and spatial misalignment. Without modifying the model architecture, traversal conditioning improved performance from \textbf{19.86\% to 25.00\%} on Kaggle, a \textbf{+6.14\% absolute gain} over the baseline.

\subsection{Data Augmentation via Cellular Automata}
\label{fig:automata_data_gen}

A \href{https://en.wikipedia.org/wiki/Cellular_automaton}{Cellular} \href{https://www.kaggle.com/code/arsenynerinovsky/cellular-automata-as-a-language-for-reasoning}{Automaton} (CA) is a discrete computational model that transforms grids into new grids based on local transition rules. Each cell’s new color (or state) is determined by its current color and the colors of the cells in its neighborhood. These rules are uniform across the grid and applied simultaneously to all cells. Typically, the grid is updated iteratively, often continuing until a stable state (saturation) is reached. The iconic example of a cellular automaton is \href{https://en.wikipedia.org/wiki/Conway%27s_Game_of_Life}{Conway's Game of Life}.

\begin{figure}[htbp]
\centering
\includegraphics[width=0.9\linewidth]{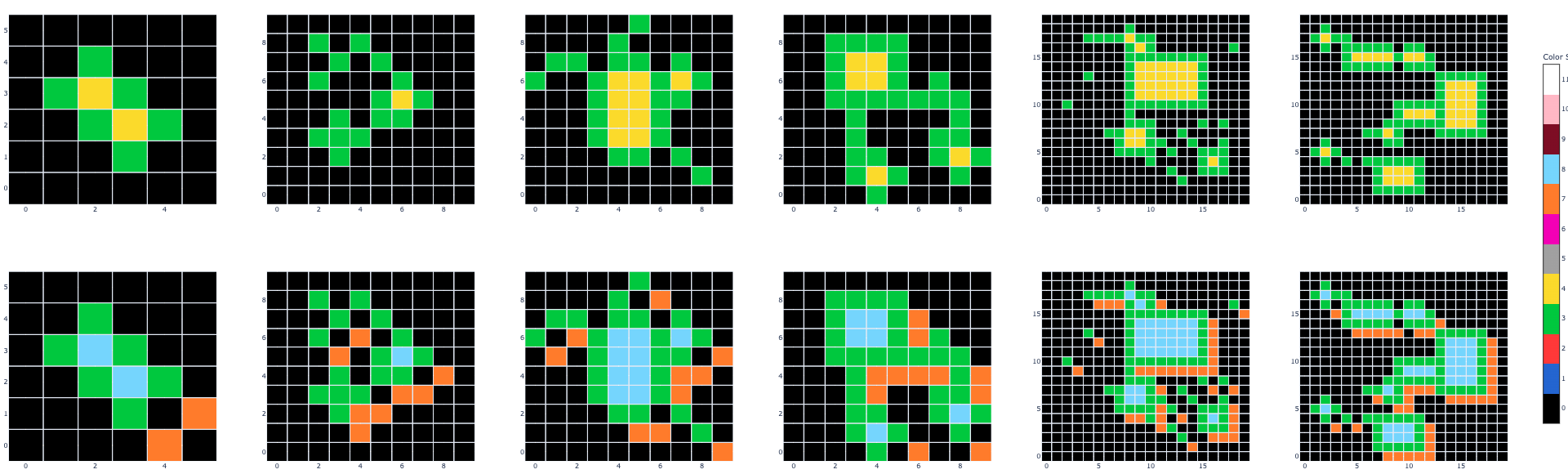}
\caption{\label{fig:automata_example.png}Example of automaton applied to the output of task $00d62c1b$. The two rules are: 
(1) yellow pixels become cyan (2) black pixels with green upper-left neighbors become orange.}
\end{figure}

Cellular Automata (CA) are an excellent fit for experimenting with ARC data, allowing us to generate over $750,000$ new tasks. An example of an automaton applied to an ARC task is shown in Figure~\ref{fig:automata_example.png}.

Given a task consisting of a list of input-output grid pairs $(I_k, O_k)$, we randomly sample a low-complexity automaton $f$ and apply one of the following generation schemes (Appendix~\ref{app:datagen_automata_algos}):

\begin{enumerate} 
\label{automata_data_gen}
\item Apply the automaton to the input grids so that the new task becomes $T' = (I_k, f(I_k))$. This is an entirely new logic. 
\item Apply the automaton to the output grids so that the new task becomes $T' = (I_k, f(O_k))$. This adds up to the existing logic.
\item If the automaton does not cause a loss of information, apply the automaton only to the input grids while keeping the original output. The resulting task is $T' = (f(I_k), O_k)$, adding up again to the existing logic. 
\item If the automaton does not cause a loss of information, apply the automata to both the input and output grids. The resulting task is $T' = (f(I_k), f(O_k))$, very closely related to the original logic. 
\end{enumerate}

The first two schemes always work, regardless of the automaton. However, the last two schemes require that the automaton $f$ does not cause a loss of information. This condition is equivalent to $f$ being locally invertible on the inputs (see Appendix~\ref{app:conjugation} for technical details).

\begin{figure}[htbp]
\centering
\includegraphics[width=0.55\linewidth]{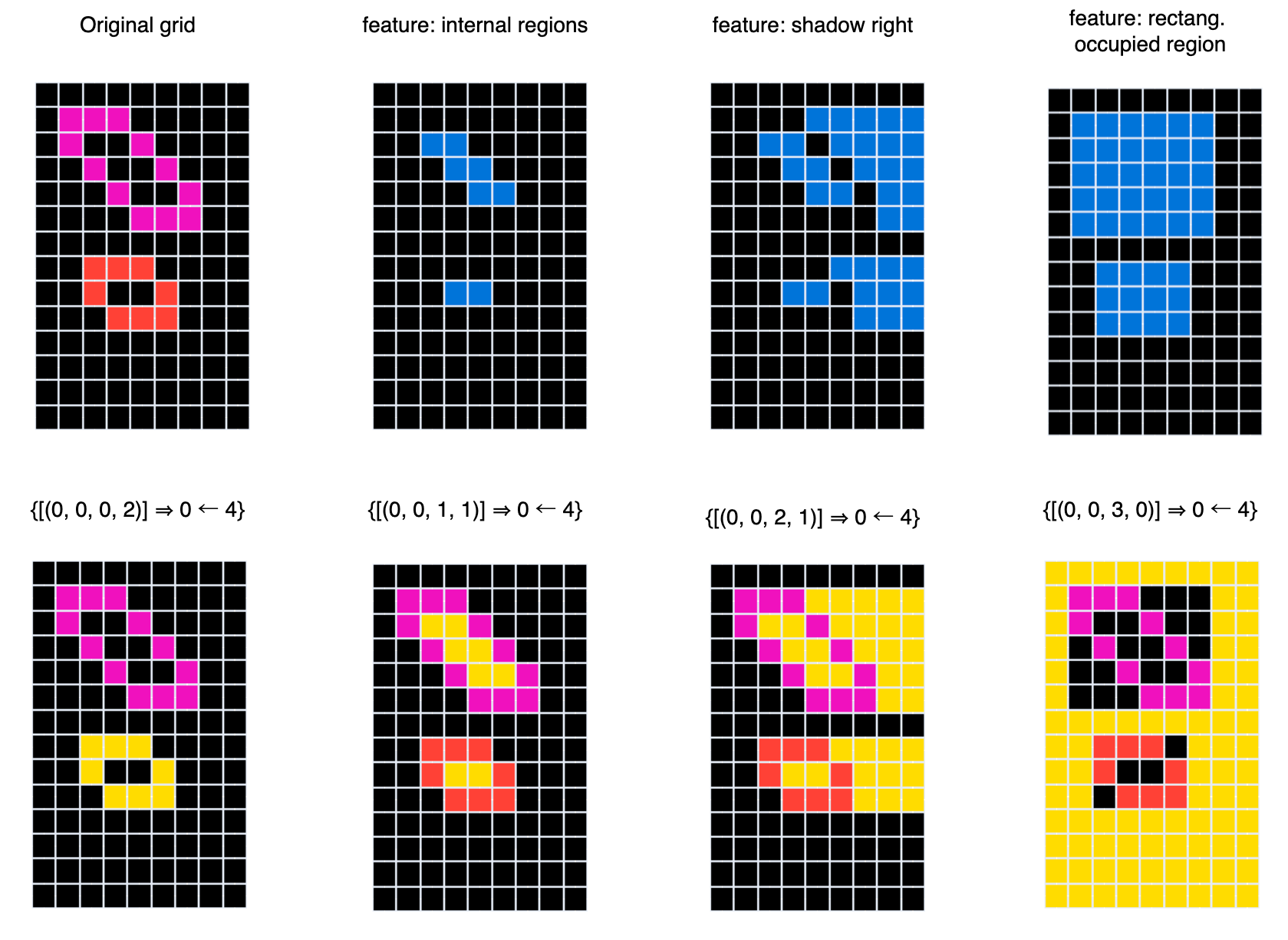}
\caption{\label{fig:example_automata_features.png}Top: example of 3 pixel features for an input grid of task 025d127b. Bottom: CA using pixel features.}
\end{figure}

\begin{figure}[htbp]
    \centering
    \begin{subfigure}{0.8\textwidth}
    \centering
    \includegraphics[height=3.5cm]{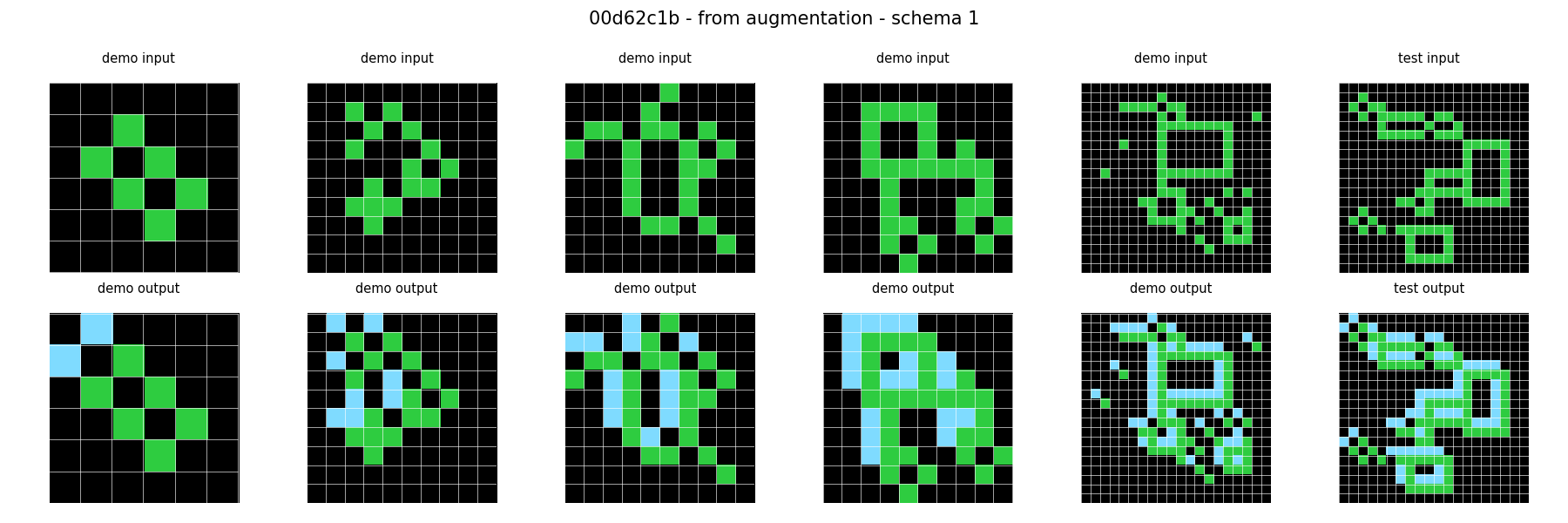}
    \caption{\label{fig:00d62c1b_aug_schema_1} CA + feature augmentation schema 1}
    \end{subfigure}
    \begin{subfigure}{0.8\textwidth}
    \centering
    \includegraphics[height=3.5cm]{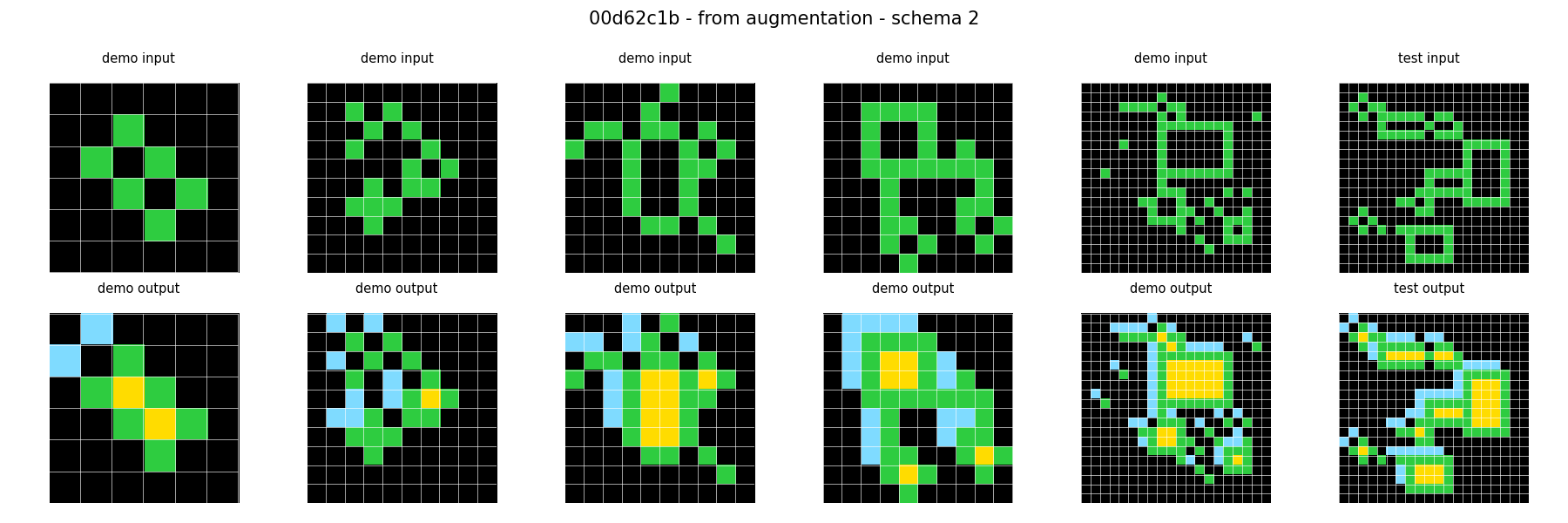}
    \caption{\label{fig:00d62c1b_aug_schema_2}CA + feature augmentation schema 2}
    \end{subfigure}
    \begin{subfigure}{0.8\textwidth}
    \centering
    \includegraphics[height=3.5cm]{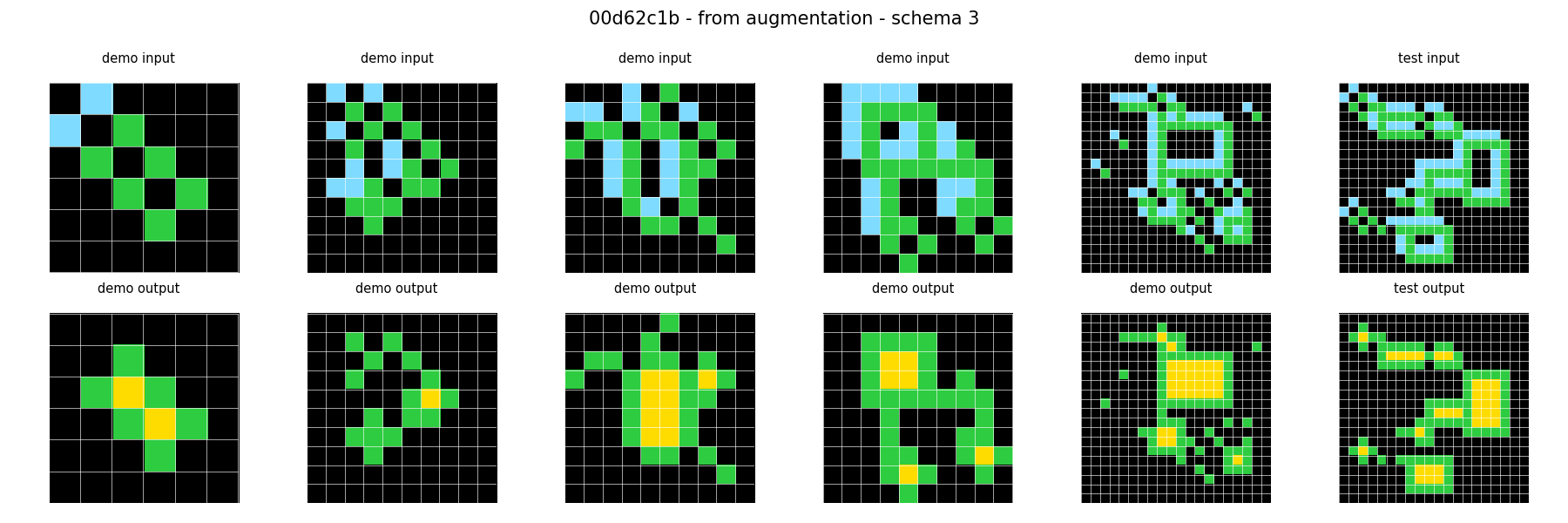}
    \caption{\label{fig:00d62c1b_aug_schema_3}CA + feature augmentation schema 3}
    \end{subfigure}
    \begin{subfigure}{0.8\textwidth}
    \centering
    \includegraphics[height=3.5cm]{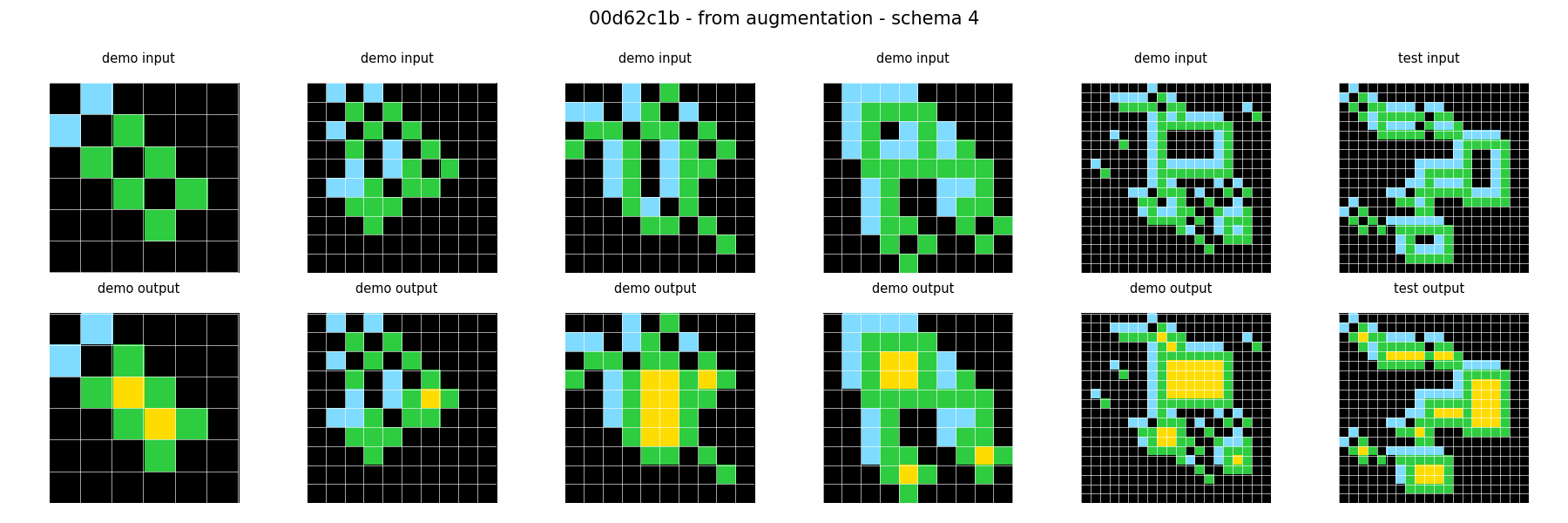}
    \caption{\label{fig:00d62c1b_aug_schema_4}CA + feature augmentation schema 4}
    \end{subfigure}
\caption{\label{fig:00d62c1b_aug_schemas} Examples of generation schemas using CA and pixel features for task 00d62c1b.}
\end{figure}

To address the lack of interpretability in automata, we also extend the approach by incorporating features such as symmetries, objects, holes, and other forms of prior knowledge. Technical details are provided in Appendix~\ref{app:automata_features}. Examples obtained by this technique can be seen in Figure~\ref{fig:example_automata_features.png} and Figures~\ref{fig:00d62c1b_aug_schema_1}–\ref{fig:00d62c1b_aug_schema_4}.

Finally, we note that our automata pipeline can be applied to any ARC data. In addition to the original tasks and our internal dataset, we also used it on other open-source datasets, as detailed in Table~\ref{tab:open_source_datasets}.

\begin{table}[htbp]
\centering
\begin{tabular}{l|cc}
dataset & size & no. of samples used \\\hline
\href{https://github.com/frankaging/ARC_synthetic_extend}{ARC-synthetic-extend} & 419 & all \\
\href{https://github.com/victorvikram/ConceptARC/tree/main}{ConceptARC} & 160 & all \\
\href{https://github.com/KSB21ST/MINI-ARC}{Mini-ARC} & 149 & all \\
\href{https://github.com/qugank/pqa.github.io}{PQA} & 70 & all  \\
\href{https://github.com/seedling123/Sequence_ARC}{Sequence-ARC} & 25 & all \\
\href{https://github.com/neoneye/arc-dataset-collection/tree/main/dataset/Sort-of-ARC}{Sort-of-ARC} & 20 & all \\
\href{https://github.com/arc-community/arc}{arc-community} & 23 & all \\
\href{https://github.com/neoneye/arc-dataset-diva}{arc-dataset-diva} & 1200 & all \\
\href{https://github.com/neoneye/arc-dataset-tama}{arc-dataset-tama} & 951 & all \\
\href{https://www.kaggle.com/datasets/zaharch/arc-nosound-tasks}{nosound} & 9 & all \\
\href{https://github.com/xu3kev/BARC}{BARC} & 400k & $\approx$150 \\\hline
Total & &  $\approx 2275$
\end{tabular}
\caption{\label{tab:open_source_datasets}Open source datasets used in this work.}
\end{table}

\subsection{Summary of Augmentation Strategy Across TTT, Decoding, and Scoring}
\label{sec:aug_summary}

As argued above, augmentations play a central role in our pipeline by enabling \emph{task-level generalization}, \emph{symmetry consistency}, and \emph{robust decoding}. Rather than treating augmentation as a simple data expansion technique, we use it as a unifying principle across \textbf{fine-tuning, inference, and candidate scoring}. Above we discussed what they are and why they are relevant, below we summarize how augmentations are applied in each stage of our inference pipeline. For details about the offline training augmentations, please refer to Section~\ref{sec:offline:aug}.

\paragraph{Stage 1: Test-time training (AE).}
During Test-Time Training (TTT), we construct a small adaptation dataset per task by iterating over demonstrations as pseudo-test pairs. For each mini-dataset, we apply:
\begin{itemize}
    \item \textbf{A -- Augment}: sampled per iteration and consistently applied to all grids
    \begin{itemize}
        \item \textbf{Rigid transformations (D4)}: rotations $\{0^\circ, 90^\circ, 180^\circ, 270^\circ\}$ and flips (horizontal/vertical/diagonal). We use always all 8 of them.
        \item \textbf{Color permutations}: random bijective mappings of colors (randomly sampled).
        \item \textbf{Demo reordering}: random permutation of demonstration order (randomly sampled).
    \end{itemize}
    \item \textbf{E -- Encode}: serialize grids using one of two traversals:
    \[
    E = \{\texttt{row\_by\_row (RbR)},\; \texttt{snake (SNK)}\}
    \]
\end{itemize}
This yields a rich augmented adaptation set without changing the semantic mapping of the task. Fine-tuning is performed using LoRA ($r=8$) with $35$ epochs per task and a learning rate of $lr=8 \times 10^{-4}$. In our final configuration, we only used the \texttt{RbR} traversal in TTT (while we kept both traversals for the offline training).

\paragraph{Stage 2: Candidate Generation (AEIDR).}
During decoding, augmentations are used not only to enhance robustness but also to \emph{enrich hypotheses} by generating candidate outputs under multiple views:
\begin{itemize}
    \item \textbf{A -- Augment}: sample $A=18$ augmented views of the task using D4 + color permutations + reordering.
    \item \textbf{E -- Encode}: serialize only with \texttt{RbR}.
    \item \textbf{I -- Infer}: run beam search ($B=10$) to produce candidates.
    \item \textbf{D -- Decode}: invert encoding to 2D grids.
    \item \textbf{R -- Reverse augment}: map predictions obtained via augmentations back to the original task space.
\end{itemize}
This produces up to $18 \times 10 = 180$ candidate grids in original coordinates. Candidates violating symbolic priors (Section~\ref{sec:filtering}) are discarded.

\paragraph{Stage 3: Candidate Scoring (AEI).}
To select the best hypotheses, we evaluate each candidate grids $\hat{O}$ from Stage 2 across multiple \emph{symmetry perspectives}:
\begin{itemize}
    \item \textbf{A -- Augment}: apply $A=8$ rigid $D_4$ transforms.
    \item \textbf{E -- Encode}: use only the $E=\{\texttt{RbR}\}$ traversal.
    \item \textbf{I -- Infer}: compute log-likelihood $\log p_\theta(y \mid x)$ under each $(A, E)$ pair without generating new tokens but simply via a forward pass.
\end{itemize}
The final candidate score aggregates all symmetry-conditioned likelihoods:
\[
S(\hat{O}) = \sum_{a \in D_4} \sum_{e \in \{RbR,SNK\}} \log p_\theta\big(e(a(\hat{O})) \mid e(a(I))\big).
\]
We rank candidates by $S(\hat{O})$ and select the top 2 for evaluation. This can be interpreted as a \textbf{change of perspective strategy}, where the best solution is the one that remains most consistent under symmetry transformations.

\paragraph{Summary.}
\begin{center}
\begin{tabular}{lccc}
\hline
\textbf{Stage} & \textbf{Goal} & \textbf{Augmentations Used} & \textbf{Encodings} \\
\hline
Stage 1: Fine-tuning & Adapt per task (TTT) & $D_4$ + colors + reordering & RbR \\
Stage 2: Generation & Diversity in decoding & $D_4$ + colors + reordering & RbR \\
Stage 3: Scoring & Consistency under symmetry & $D_4$ & RbR \\
\hline
\end{tabular}
\end{center}

\section{Model Architecture}
\label{sec:longt5}

\subsection{Overview and Motivation}

The ARC challenge requires \emph{pixel-perfect} reasoning: a single incorrect token in the output grid leads to a score of zero for the task. This strict scoring paradigm imposes a strong requirement on the attention mechanism: it must process long sequences with \emph{high token-level precision}. In our formulation, each ARC grid is serialized into a sequence of up to 10{,}000 tokens (when combining multiple input-output demonstrations). Standard Transformer attention scales quadratically with sequence length and becomes impractical. To overcome this, we adopt the \textbf{LongT5} model architecture, which is designed for efficient long-context processing.

However, once the sequence length exceeds $\sim 1000$ tokens, local-only attention becomes insufficient, as it restricts token interactions to a fixed neighborhood and fails to capture long-range dependencies common in ARC rules (e.g.\ symmetry relationships, object movement, or counting over entire grids). For this reason, we rely on \textbf{Transient Global Attention} (TGlobal), which provides a sparse but expressive global interaction pattern across tokens.

\subsection{LongT5 Encoder--Decoder Architecture}

LongT5 extends the original T5 architecture to support long input sequences (up to $16\mathrm{k}$ tokens) while maintaining computational efficiency. It retains the \textbf{encoder--decoder} design:
\begin{itemize}
    \item The \textbf{encoder} is adapted to handle long contexts using a combination of local and transient global attention patterns.
    \item The \textbf{decoder} remains unchanged from T5 and performs standard causal self-attention. This choice aligns with our ARC use case, as output sequences typically range from 500 to 900 tokens.
\end{itemize}

\begin{figure}[htbp]
    \centering
    \includegraphics[width=0.7\textwidth]{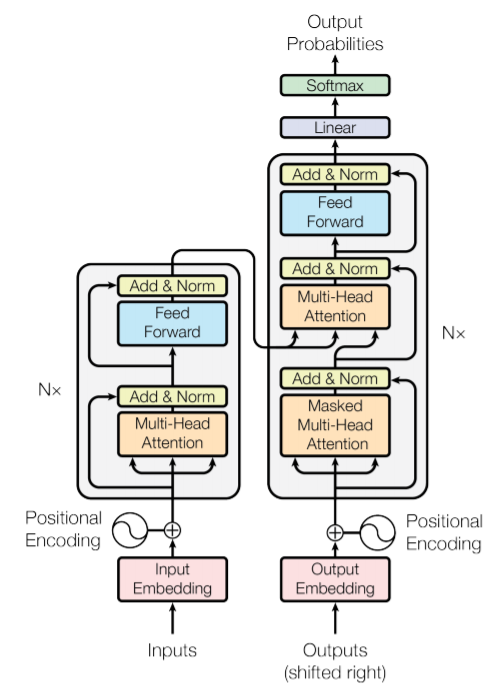} 
    \caption{Transformer architecture. Left part is the encoder, right part is the decoder. T5 architecture is very similar to this one: longT5 however does the positional encoding in a different manner, within the attention mechanism. Taken from \cite{vaswani2017attention}.}
    \label{fig:t5}
\end{figure}

A schematic representation of the encoder--decoder architecture is provided in Figure~\ref{fig:t5}.

ARC tasks are processed as follows:
\begin{itemize}
    \item The \textbf{encoder} receives the input-output training pairs together with the test input.
    \item The \textbf{decoder} generates the unknown test output.
\end{itemize}

During training, a causal mask is applied only in the decoder self-attention. During inference, the test output is generated autoregressively by feeding back previously generated tokens.

\subsubsection{Attention Mechanisms in LongT5}

The main architectural distinction between T5 and LongT5 lies in the encoder attention pattern. LongT5 supports two modes:
\begin{enumerate}
    \item \textbf{Local Attention}: each token attends only to a fixed window of neighbors.
    \item \textbf{Transient Global Attention (TGlobal)}: introduces sparse global tokens that change dynamically across layers.
\end{enumerate}

\begin{figure}[htbp]
    \centering
    \includegraphics[width=0.8\textwidth]{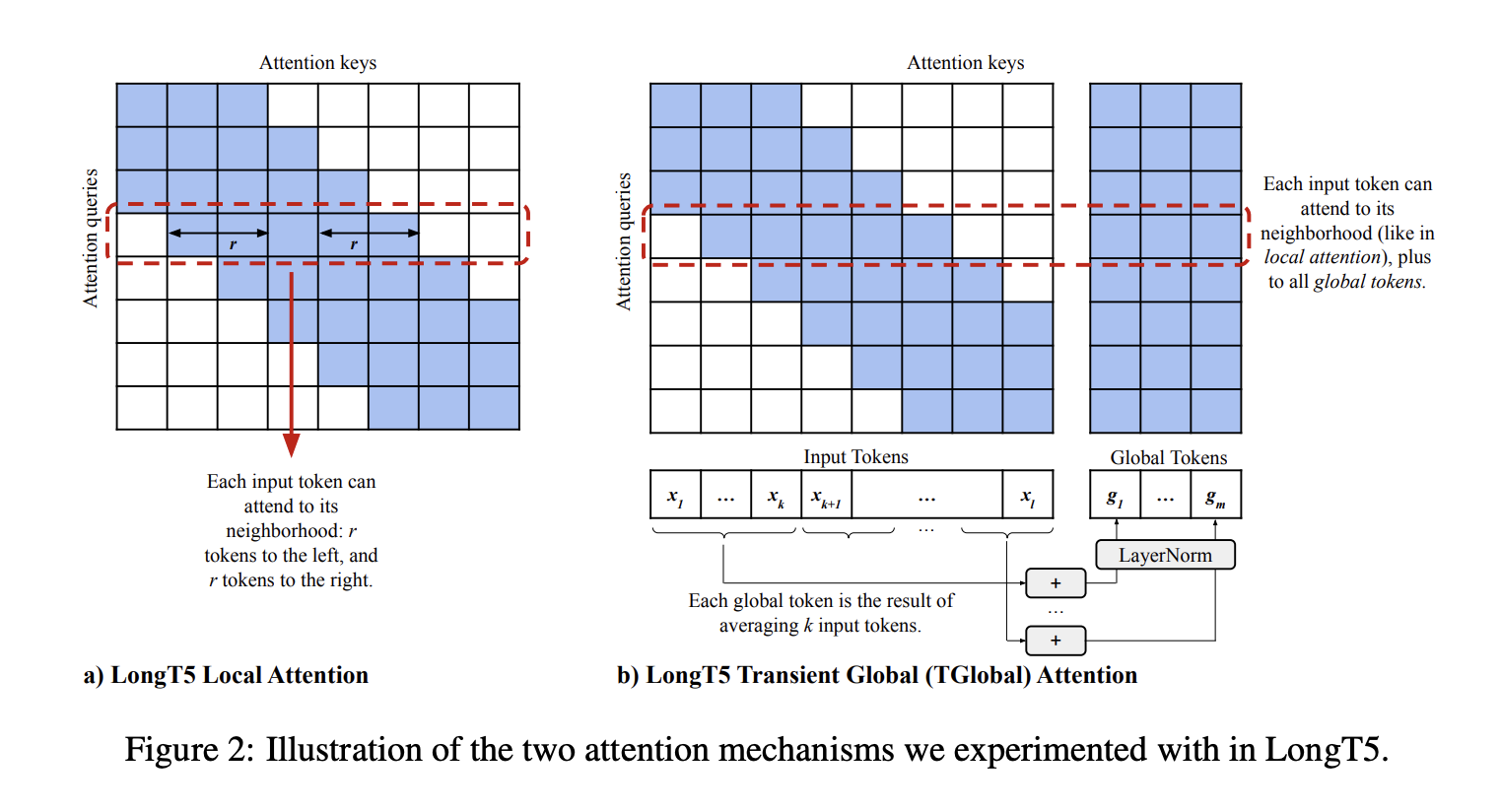} 
    \caption{Differences between the local attention (left) and the transient-global attention (right). Each block is a \textbf{local window}. Taken from \cite{guo2021longt5}.}
    \label{fig:attn}
\end{figure}

Transient global attention is illustrated in Figure~\ref{fig:attn}.

In this mechanism, the input sequence is divided into fixed-size blocks, and each block is summarized into a \emph{temporary global token} obtained by aggregating (e.g., summing and normalizing) the embeddings of the tokens in that block.
During attention, each input token attends both to its local neighborhood (as in Local Attention) and to all global tokens, allowing information to propagate across distant parts of the sequence.
These global tokens are called \emph{transient} because they are dynamically computed at every layer, rather than being fixed once for the whole model (as in other approaches \cite{ravula2020etc}).

Formally, for an input sequence $X \in \mathbb{R}^{n \times d}$, the transient global attention pattern partitions indices into local and global subsets:
\[
    \mathcal{I} = \mathcal{I}_{\mathrm{local}} \cup \mathcal{I}_{\mathrm{global}}, \qquad |\mathcal{I}_{\mathrm{global}}| \ll n.
\]
The attention mask $M$ is then defined as:
\[
    M_{ij} =
    \begin{cases}
        0 & \text{if } j \in \mathrm{LocalWindow}(i) \text{ or } j \in \mathcal{I}_{\mathrm{global}}, \\
        -\infty & \text{otherwise}.
    \end{cases}
\]
This allows each token to attend locally while a rotating subset maintains full connectivity, enabling long-distance reasoning crucial to ARC tasks.

\subsection{Motivation for Implementation Improvements}

We trained LongT5 with transient global attention for our ARC benchmark, achieving strong performance and surpassing our previous state-of-the-art baseline. However, both training and inference were \textbf{extremely slow}, primarily due to the legacy attention implementation in PyTorch, which incurs high memory overhead and poor efficiency for long sequences.

Modern attention mechanisms like \textbf{FlashAttention} dramatically improve speed and memory efficiency through tiling and recomputation strategies. However, FlashAttention does not natively support \textbf{relative positional bias}, which is critical in T5-family models, including LongT5. This makes direct integration non-trivial.

\subsection{FlashAttention Integration}
\label{sec:flash_attn}
To resolve this, we implemented ourselves a version where FlashAttention is extended to support attention bias terms. We adapted this implementation and integrated it into LongT5.

\paragraph{Design Choices:}
\begin{itemize}
    \item FlashAttention was applied only to the \textbf{encoder}, which processes 5k--10k token sequences.
    \item The decoder, which handles shorter sequences ($<1000$ tokens), continues to use standard PyTorch attention.
\end{itemize}

This provided a meaningful trade-off: reduced GPU memory, faster training, and full compatibility with existing attention.

\textbf{Note.} Since FlashAttention does not come natively with dropout, we actually used it only during test-time training only. During offline training we use the standard attention implementation with a dropout of $0.1$, to improve on stability and generalization.

\subsection{Results}

We evaluated the modified LongT5 model on randomly selected 38 ARC tasks. Table~\ref{tab:flash-results} summarizes the results. Online tuning time improved by a factor of:
\[
\frac{1.24}{0.78} = 1.589\times
\]
while maintaining or improving task accuracy.

\begin{table}[htbp]
\centering
\begin{tabular}{lccccccc}
\hline
\textbf{Encoder} & \textbf{LayerNorm} & \textbf{Score} & \textbf{UP} & \textbf{Online-FT} & \textbf{Decoding} & \textbf{Total Time} \\
\hline
Vanilla & Vanilla & 6.8\% & 13.88\% & 1.24 h & 0.71 h & 1.95 h \\
Flash & Vanilla & 7.638\% & 14.72\% & 0.78 h & 0.70 h & \textbf{1.48 h} \\
Flash & Liger\_Triton & \textbf{7.777\%} & \textbf{14.721\%} & \textbf{0.75 h} & \textbf{0.69 h} & \textbf{1.44 h} \\
\hline
\end{tabular}
\caption{Performance comparison between vanilla LongT5 and FlashAttention-enhanced variants.}
\label{tab:flash-results}
\end{table}

Our first submissions to Kaggle used a \textbf{Llama3.1 1B} model: after changing the architecture to \textbf{LongT5} and integrating all the aforementioned optimizations, the score improved significantly, going \textbf{from 5.00\% to 12.08\%}, an absolute \textbf{gain of +7.08\%}.

\subsection{Attempted Speed-Ups via Fused Relative Positional Bias}

Unlike GPT-style models that use absolute positional encodings, LongT5 employs a \textbf{relative position bias} computed dynamically during attention:
\[
\mathrm{Attn}(Q,K,V) = \mathrm{softmax}\left(QK^\top + B_{\mathrm{rel}}\right)V,
\]
where $B_{\mathrm{rel}}$ is computed via bucketed relative distances using:
\[
B_{\mathrm{rel}}(i,j) = W_{\mathrm{bucket}}\big(\mathrm{bucket}(i-j)\big),
\]
with logarithmic bucketing for large distances. To avoid creating tensors in python and passing them to CUDA, we integrated the positional bias directly in the flash-attention: memory impact dropped by a factor of four, but the compute time was slower and ultimately we could not leverage the improvements; hence, they did not reach the final pipeline.

\paragraph{Mechanism of bucketed relative distances.}
In the LongT5 architecture, positional information is encoded using a \emph{relative position bias} instead of absolute embeddings. This means that the model does not directly encode where each token is in the sequence, but rather how far apart any two tokens are. To make this computation efficient and stable for long sequences, relative distances $(i-j)$ between query position $i$ and key position $j$ are not represented explicitly. Instead, they are mapped into a discrete set of \emph{buckets} through a function $\mathrm{bucket}(i-j)$.

Concretely, small distances are treated with high precision -- each offset receives its own bucket -- while larger distances are \emph{grouped logarithmically} to reduce the total number of parameters. The mapping typically follows:
\[
b(i,j) =
\begin{cases}
|i-j|, & \text{if } |i-j| < N_\mathrm{small}, \\[4pt]
N_\mathrm{small} + \left\lfloor \log\left(\frac{|i-j|}{N_\mathrm{small}}\right) \cdot \frac{N_\mathrm{buckets}-N_\mathrm{small}}{\log\left(\frac{L}{N_\mathrm{small}}\right)} \right\rfloor, & \text{otherwise},
\end{cases}
\]
where $N_\mathrm{buckets}$ is the total number of relative position buckets (typically $32$), $N_\mathrm{small}$ defines the range of exact offsets, and $L$ is the maximum sequence length. Each bucket index $b(i,j)$ selects a learnable bias value $B_{b(i,j)}$, which is added directly to the attention logits before the softmax:
\[
\mathrm{Attn}(Q,K,V) = \mathrm{softmax}\!\left(QK^\top + B_{b(i,j)}\right)V.
\]
In this way, the model learns to assign specific preferences to relative positions—nearby tokens receive finely resolved biases, while distant ones are coarsely grouped. This mechanism allows LongT5 to generalize to unseen sequence lengths, maintain translation invariance, and capture both local and long-range dependencies efficiently.

\subsection{LongT5 Transient Global Configuration}

For our experiments, we adopted the \textbf{LongT5 transient global configuration} with the following parameters: 
\texttt{relative\_attention\_max\_distance = 1024}, 
\texttt{local\_radius = 1024}, and 
\texttt{global\_block\_size = 4}.

The \textbf{relative\_attention\_max\_distance} parameter defines the maximum span (in token positions) over which relative positional embeddings are computed in the attention mechanism. A value of 1024 allows the model to capture long-range dependencies across up to 1024 tokens, which is essential for handling long input sequences efficiently.

The \textbf{local\_radius} parameter controls the size of the local attention window around each token. Setting it to 1024 ensures that each token can directly attend to its 1024 neighboring tokens on either side, providing rich local contextualization while maintaining computational efficiency.

The \textbf{global\_block\_size} determines the number of tokens grouped into each global attention block. In our case, a value of 4 indicates that every four tokens share a global attention representation. This design enables sparse yet structured global attention, allowing the model to propagate global information periodically without the full quadratic cost of standard attention.

Together, these settings balance \textit{local detail} and \textit{global context propagation}, making the LongT5 architecture suitable for long-sequence reasoning tasks while keeping computation tractable.

\section{Offline Training}
\label{sec:offline_training}

We pretrained our LongT5 model in a large-scale offline setup using a combination of open-source ARC-style datasets and synthetic transformations. The objective of offline training was to endow the model with a broad understanding of symbolic grid transformations before fine-tuning it with task-specific adaptation.

\subsection{Distributed Training Setup}

Training was performed using the standard HuggingFace \texttt{Trainer} API combined with \textbf{Distributed Data Parallel (DDP)} from PyTorch. The training ran across \textbf{8 compute nodes}, each equipped with $8 \times H100$ GPUs, resulting in a total of 64 GPUs. With a per-GPU batch size of 8, this yielded an \textbf{effective batch size of 512}:
\[
\mathrm{BS_{effective}} = 8 \times 8 \times 8 = 512.
\]

DDP was chosen over Data Parallel (DP) or Fully Sharded Data Parallel (FSDP) due to its stability and simplicity for transformer-based training. Gradients were synchronized across GPUs using \texttt{all\_reduce} at each backward step.

We initialized our model weights from the publicly available \texttt{google/long-t5-tglobal-base} checkpoint hosted on HuggingFace. FlashAttention-enhanced attention kernels, as described in Section~\ref{sec:longt5}, were used to accelerate training.

\subsection{Learning Rate Scheduling}

We trained the model for \textbf{70 epochs} using the AdamW optimizer with weight decay. A \textbf{cosine learning rate scheduler} was applied with a \textbf{linear warm-up phase of 100 steps}. The learning rate $\eta_t$ at step $t$ was defined as:
\[
\eta_t =
\begin{cases}
\eta_{\max} \cdot \frac{t}{T_{\mathrm{warmup}}}, & t \leq T_{\mathrm{warmup}}, \\[6pt]
\frac{1}{2} \eta_{\max} \left(1 + \cos\left(\pi \cdot \frac{t - T_{\mathrm{warmup}}}{T_{\mathrm{total}} - T_{\mathrm{warmup}}}\right)\right), & t > T_{\mathrm{warmup}},
\end{cases}
\]
where $\eta_{\max}$ is the peak learning rate, $T_{\mathrm{warmup}} = 100$, and $T_{\mathrm{total}}$ is the total number of training steps. The warm-up phase helps avoid optimization instability at the beginning of training, while cosine decay promotes smooth convergence.

\subsection{Curriculum Learning Strategy}

To stabilize learning and progressively expose the model to increasingly difficult symbolic transformations, we employed a \textbf{curriculum learning} strategy. Datasets were introduced in stages according to their complexity and variability. Training began with simpler grid mappings and gradually incorporated more complex generalization tasks.

The curriculum progressed through the following stages:

\begin{enumerate}
    \item \textbf{Stage 1: ARC-AGI-1 (Training Set Only).} The initial training phase used the ARC-AGI-1 training split, containing tasks generated from the original ARC training set. This phase allowed the model to learn basic grid semantics such as connected components, color grouping, and local transformations.
    \item \textbf{Stage 2: ARC-AGI-1 (Evaluation Set) + Other Public Datasets.} Next, we introduced ARC-AGI-1 evaluation data, followed by the additional public ARC-like datasets listed in Table~\ref{tab:open_source_data}. These datasets include richer structural transformations, forcing the model to generalize beyond local pixel rules.
    \item \textbf{Stage 3: ARC-AGI-2 Public Data} Finally, we incorporated ARC-AGI-2 open data, which feature tasks that require compositional reasoning. This last phase significantly increased task diversity and helped prepare the model for high-level reasoning.
\end{enumerate}

\subsection{Augmentation Pipeline}
\label{sec:offline:aug}

All datasets in the curriculum were passed through the augmentation pipeline described in Section~\ref{sec:data_aug}, which includes:

\begin{itemize}
    \item \textbf{Symmetry augmentations} (group orbits via rotations, flips, and reflections).
    \item \textbf{Grid traversals} (Section~\ref{sec:traversal}): row-by-row and snake encodings for representational robustness.
    \item \textbf{Automata perturbations}: automata-based augmentations to enforce robustness to minor logic changes.
\end{itemize}

This augmentation strategy ensured that the model learned task rules rather than overfitting to specific grid layouts or token sequences.

\subsection{Training to Solve}
\label{sec:training_to_solve}

The first component of our offline training strategy focuses on \emph{training the model to solve ARC tasks directly}. In this setting, the model receives a serialized prompt containing the input--output demonstrations along with the test input and is trained to \textbf{autoregressively generate the test output grid}. This training mode uses a standard \textbf{causal language modeling} objective based on cross-entropy.

\paragraph{Objective Function.}
Given a target output sequence $\mathbf{y} = (y_1,\dots,y_T)$ and model predictions $\hat{\mathbf{y}}$, we minimize the token-level cross-entropy loss:
\[
\mathcal{L}_{\mathrm{CE}} = - \sum_{t=1}^T \log p_\theta(y_t \mid y_{<t}, \mathbf{x}),
\]
where $\mathbf{x}$ encodes the training examples and test input grid, $\mathbf{y}$ the (true) test output grid, and $p_\theta$ is the model. In case of multiple tests, the task is split into multiple independent tasks, each with only one test.

In some experiments, we explored stability-enhanced variations:
\begin{itemize}
    \item \textbf{Goldfish loss:} a smoothed cross-entropy variant where tokens are randomly masked \cite{hans2024like}. The idea is to avoid "by-heart memorization": however, due to the pixel-level required precision of ARC-AGI, the performance of our models diminished vis-à-vis of standard cross-entropy.
    \item \textbf{Don't go away loss (DGA):} we add a regularization term to keep model weights close to the pretraining initialization $W_0$:
    \[
    \mathcal{L}_{\mathrm{DGA}} = \lambda \| W - W_0\|_2^2.
    \]
    The results were however almost identical to the standard cross-entropy, leading us to prefer simpler version.
\end{itemize}

\paragraph{Optimizer Configuration.}
During offline training, we used the \texttt{adamw\_torch\_fused} optimizer. During test-time training, we employed \texttt{adamw\_8bit}.
Gradient updates follow the same cosine schedule with warm-up described in Section~\ref{sec:offline_training}. The training was run in full precision (\texttt{fp32}).

\paragraph{Regularization.}
We applied only \textbf{Gradient clipping} ($\|\nabla\|_2 \le 1.0$) to stabilize training at large batch sizes.

This stage ensures that the model learns to produce full and pixel-accurate ARC outputs and provides strong decoding behavior.

%-------------------------------------------------------------

\subsection{Training to Understand}
\label{sec:training_to_understand}

Although training the model to solve tasks directly emphasizes output correctness, it does not necessarily lead to deep \emph{task understanding}. To foster deeper reasoning, we introduce a complementary training objective where the model learns to \textbf{reconstruct masked regions of the grid and infer the underlying task logic}. This approach follows the \textbf{Unified Language Learning Paradigm} (UL2)~\cite{tay2022ul2}, which emphasizes \emph{bidirectional reasoning and intermediate inference} through denoising-based pretraining.

\begin{figure}[htbp]
    \centering
    \begin{subfigure}{1.0\textwidth}
    \centering
    \includegraphics[height=3.5cm]{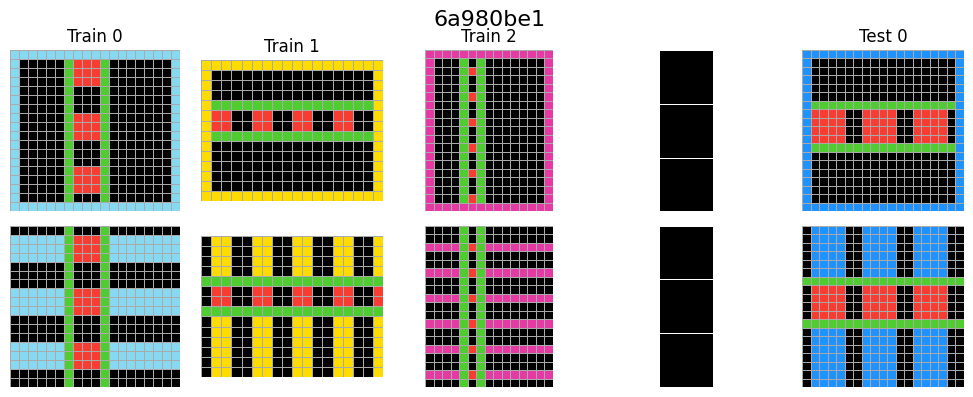}
    \caption{Original ARC transformation task.}
    \end{subfigure}

    \begin{subfigure}{1.0\textwidth}
    \centering
    \includegraphics[height=3.5cm]{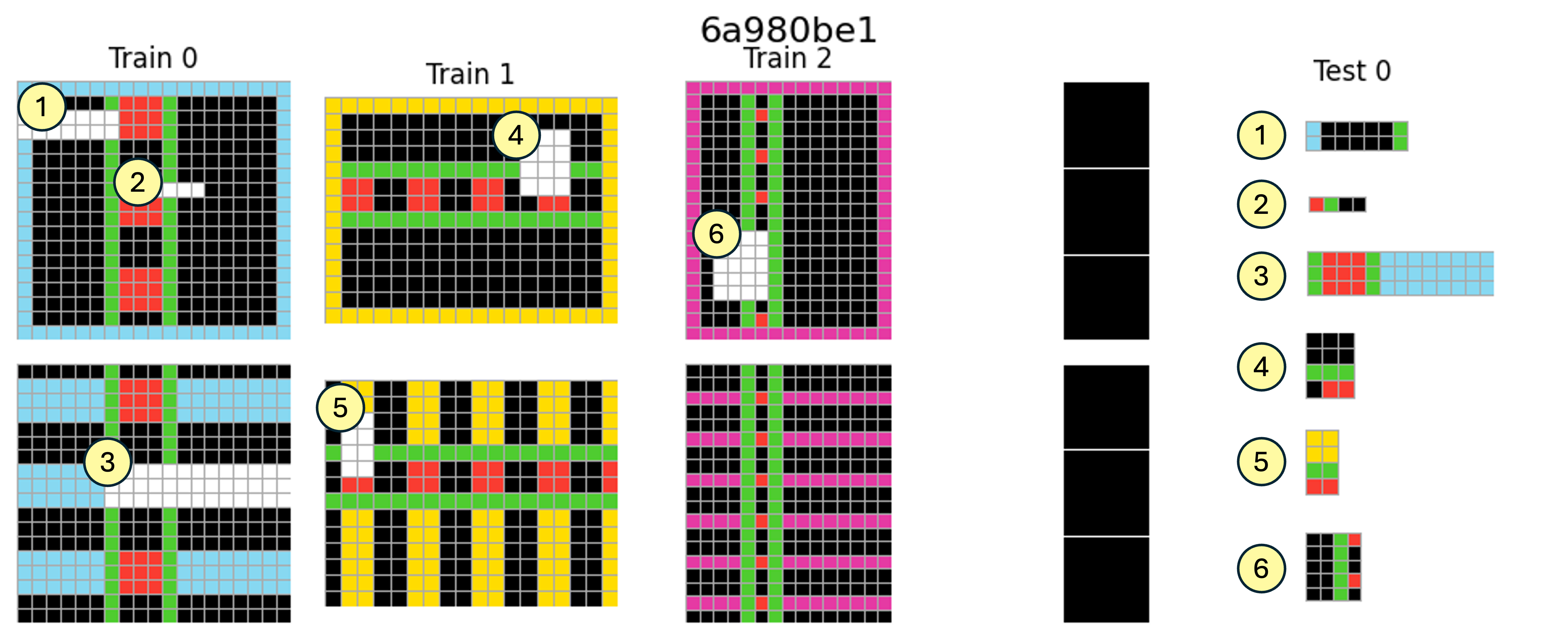}
    \caption{UL2 training.}
    \end{subfigure}
    \caption{Training with UL2-style denoising objectives for ARC tasks: portions of the input grids are masked and become target output sequences to be predicted}
    \label{fig:ul2}
\end{figure}

\paragraph{UL2 Masked Training.}
In UL2, the model is prompted to predict masked tokens within the input sequence rather than being prompted to predict the final test output directly. The framework defines three distinct de-masking modes that vary in noise ratio and masking pattern, but share the same core principle: a subset of grid tokens in the demonstration grids is replaced with special mask tokens (\texttt{<|mask-0|>}, \texttt{<|mask-1|>}, \dots) and the model must reconstruct these missing elements during the decoding phase, token by token. Figure~\ref{fig:ul2} illustrates this process.

We apply this masked reconstruction objective to the training grids, enabling the model to reason over incomplete contexts. This discourages overfitting to autoregressive copying patterns and instead promotes \textbf{latent rule discovery} and \textbf{structural understanding}. 

To further enhance learning stability, we integrated a \textbf{curriculum learning strategy} with UL2. During training, we interleaved standard ARC tasks (where the model predicts the test output) with denoising tasks. The relative proportion of these task types was gradually varied over time: we began with an ARC-heavy phase, progressively increased the proportion of denoising tasks, and finally shifted back toward ARC-heavy batches. Ablation studies on various mixture ratios revealed that the best results were achieved with a 60\% denoising mix—i.e., 60\% of training batches consisting of UL2-style masked reconstruction and 40\% of standard ARC prediction tasks. Interestingly, higher noise regimes yielded the strongest performance, suggesting that noise acts as an effective regularizer that promotes generalization.

In local experiments, this approach yielded consistent improvements of approximately 2 percentage points over the baseline without UL2 training. Overall, these findings indicate that incorporating UL2-style denoising within a curriculum schedule fosters deeper task understanding and leads to more generalizable ARC reasoning models.

\paragraph{Results.}
Training with multitask learning improved robustness to decoding uncertainty and pixel-level precision. Without modifying the model architecture, UL2-style denoising improved performance from \textbf{25.00\% to 27.08\%} on Kaggle, a \textbf{+2.08\% absolute gain} over the baseline.

\subsection{Multi-Task Offline Training}
To combine both capabilities -- solving and understanding -- we train \textbf{with two alternating batch types}:
\begin{itemize}
    \item \textbf{Solve batches:} autoregressive output prediction using cross-entropy loss.
    \item \textbf{Understand batches:} masked-grid reconstruction using a UL2 denoising loss.
\end{itemize}
These are mixed randomly during training to prevent mode collapse.

This mixed objective builds a model that not only generates outputs but also reasons step by step and corrects itself when needed -- an ability that is crucial for ARC reasoning.

\subsection{Grokking}
\label{sec:grokking}

In this final subsection, we briefly discuss \emph{grokking}, which we believe played an important role in improving our model's generalization and, consequently, its final score on Kaggle.
Grokking refers to a phenomenon observed in deep learning where a model initially overfits the training data, achieving low training loss but high validation loss, and then—after continued training, suddenly transitions to a phase of strong generalization without any change in the dataset or model architecture. This delayed generalization phase was first described by \cite{power2022grokking}, and it is typically associated with a shift in the internal representation dynamics and weight-space geometry of the network.

Recent studies suggest that grokking can be quantitatively characterized through the spectral properties of the weight matrices. Tools such as \texttt{WeightWatcher}\footnote{\url{https://weightwatcher.ai/}} analyze the eigenvalue distribution of the correlation matrices associated with each layer's weight matrix:
\[
Y_l = W^{\top}_l W_l,
\]
where \( W_l \) is the weight matrix of a given layer.

By examining the empirical spectral density of all \( Y_l \)'s together and fitting the tail of the eigenvalue distribution to a power law \( P(\lambda) \sim \lambda^{-\alpha} \), one can detect whether the model has entered a grokking regime. Empirically, if the fitted exponent satisfies \( 2 < \alpha < 4 \), the layer is said to exhibit heavy-tailed behavior characteristic of a grokked state, indicative of a model that has transitioned from mere memorization to true generalization \cite{prakash2025grokking}.

In our experiments, we observed spectral exponents within this range for several layers, suggesting that our model had indeed undergone grokking during extended fine-tuning.

\paragraph{Results.}
Grokking significantly improved our model performance on Kaggle, raising our score from \textbf{19.86\% to 25.00\%}, a \textbf{+5.14\% absolute gain} over the non-grokked models.

\section{Inference Pipeline}
\label{sec:inference}
\subsection{Test-Time Training (TTT)}
\label{sec:ttt}

Test-Time Training (TTT) is a crucial component of our pipeline that allows the model to adapt \emph{per task} at inference time. The objective is not to retrain the model but to refine it locally so that it better fits the specific structure of a single unseen ARC task. Since ARC tasks often feature localized logic rules, small model adaptations can significantly improve performance.

\paragraph{Online Fine-Tuning Procedure.}
Given a test task with a set of input -- output demonstrations, we first take one of the training pairs and promote it to test pair: we do this in turn on all the pairs. We have called this step the ``leave-one-out''. Then, we augment each new ``leave-one-out'' pairs using group symmetries (rotations, flips) and color permutations as described in Section~\ref{sec:data_aug}. These augmentations generate multiple training pairs from a single task, forming a small but meaningful adaptation set:
\[
\mathcal{D}_{\text{TTT}} = \{(\{(I_i, O_i)\}_{i=1}^M)_j\}_{j=1}^{N}, \quad M \sim 2 {-} 5, \quad N \sim 20{-}60,
\]
where $(I_i, O_i)$ indicates a pair of grids (and the dataset item -- a.k.a. a task -- is a short sequence of pairs of grids), $M$ corresponds to the number of training examples, and $N$ is the dataset size obtained combining the different augmentations. We then perform a short fine-tuning loop for few gradient steps, updating only a small subset of parameters using LoRA.

\paragraph{LoRA Configuration.}
To prevent overfitting on such few examples and maintain rapid adaptation, we apply \textbf{Low-Rank Adaptation (LoRA)} on the linear layers of the encoder and decoder; Let $W \in \mathbb{R}^{m \times n}$ be a weight matrix of one of these layers, then
\[
W' = W + \alpha \cdot BA,
\]
where $A \in \mathbb{R}^{r \times n}$ and $B \in \mathbb{R}^{m \times r}$ are learned low-rank matrices with rank $r$. We use:
\begin{itemize}
    \item \textbf{Rank:} $r=8$,
    \item \textbf{Scaling:} $\alpha = 16$,
    \item \textbf{Dropout:} $p=0.0$,
    \item \textbf{Target modules:} \texttt{q\_proj}, \texttt{k\_proj}, \texttt{v\_proj}, \texttt{o\_proj}, and the feed-forward layers (\texttt{linear\_1}, \texttt{linear\_2}). The Huggingface convention fo this is to call them``all-linear''.
\end{itemize}
This allows fast adaptation without modifying the base model weights, preserving pretrained knowledge. We also believe that the low rank prevents overfitting and exact memorization.

\paragraph{Memory Management.}
Since TTT is executed many times (once per task), we optimize memory usage via:
\begin{itemize}
    \item Mixed-precision (\texttt{bfloat16}) inference and training,
    \item Dynamic batching. We start with batch $n$ (usually $n=4$) and $grad\_acc = 1$; then, if the operation get out of memory, we half the batch and double the $grad\_acc$.
    \item State reuse between TTT runs. In particular, we mount and dismount the adapters only from the base model, that remains on each GPU and is loaded only once at the start of TTT.
\end{itemize}

\begin{figure}[htbp]
    \centering
    \begin{subfigure}{0.8\textwidth}
        \centering
        \includegraphics[height=3.5cm]{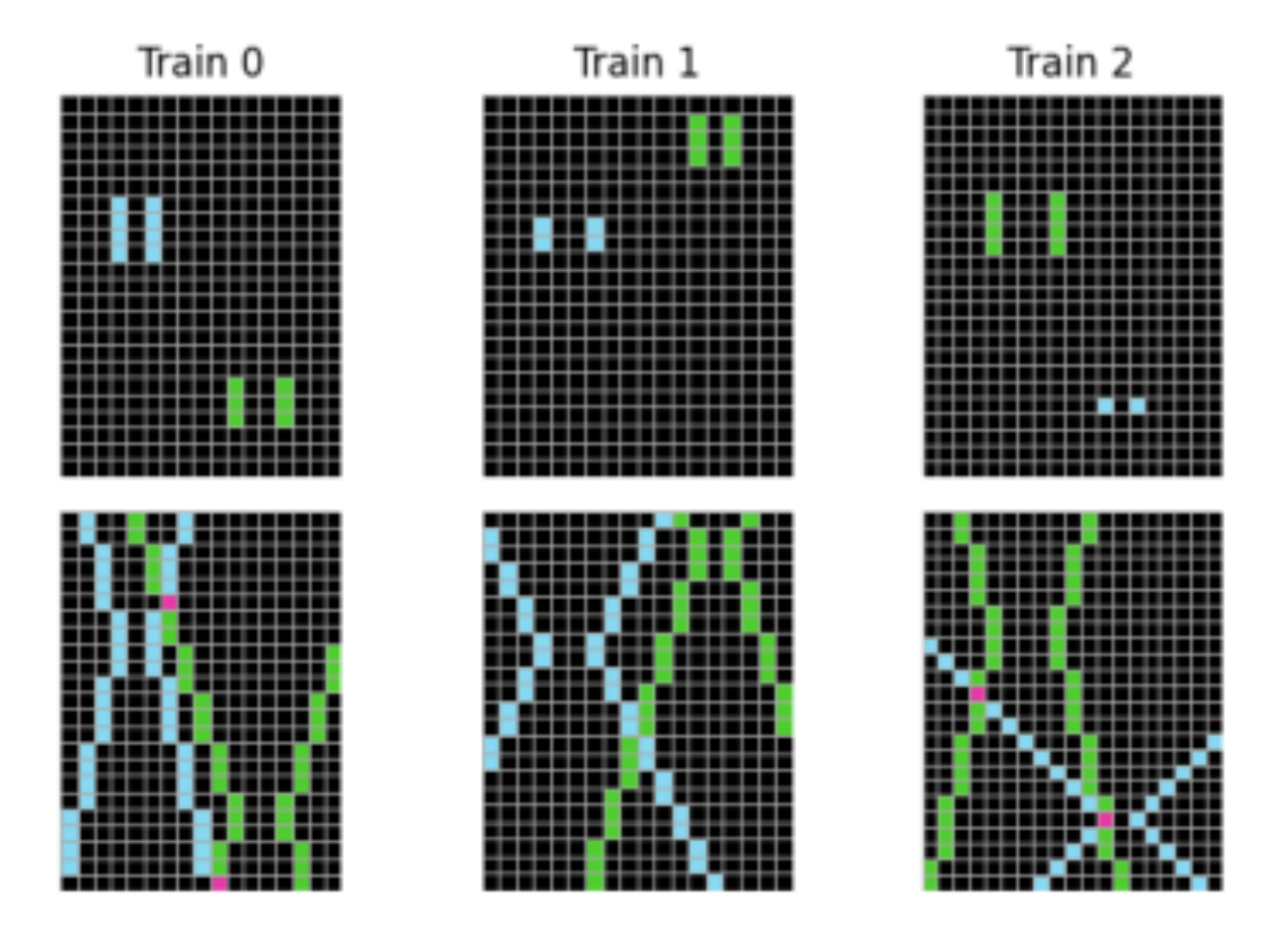}
        \caption{Reference seed task.}
    \end{subfigure}
    \begin{subfigure}{0.8\textwidth}
        \centering
        \includegraphics[height=3.5cm]{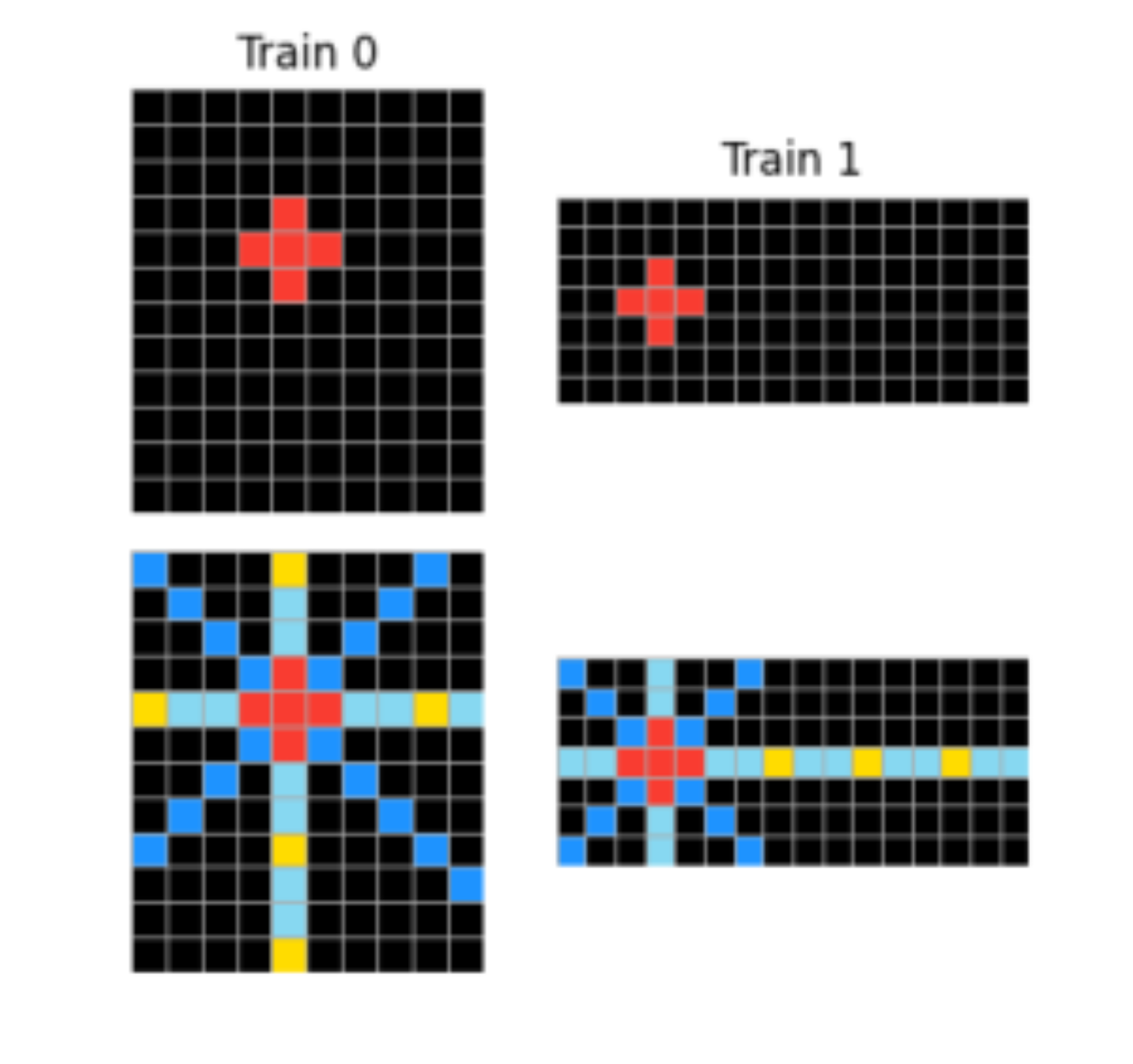}
        \caption{Nearest neighbor task.}
    \end{subfigure}
    \begin{subfigure}{0.8\textwidth}
        \centering
        \includegraphics[height=3.5cm]{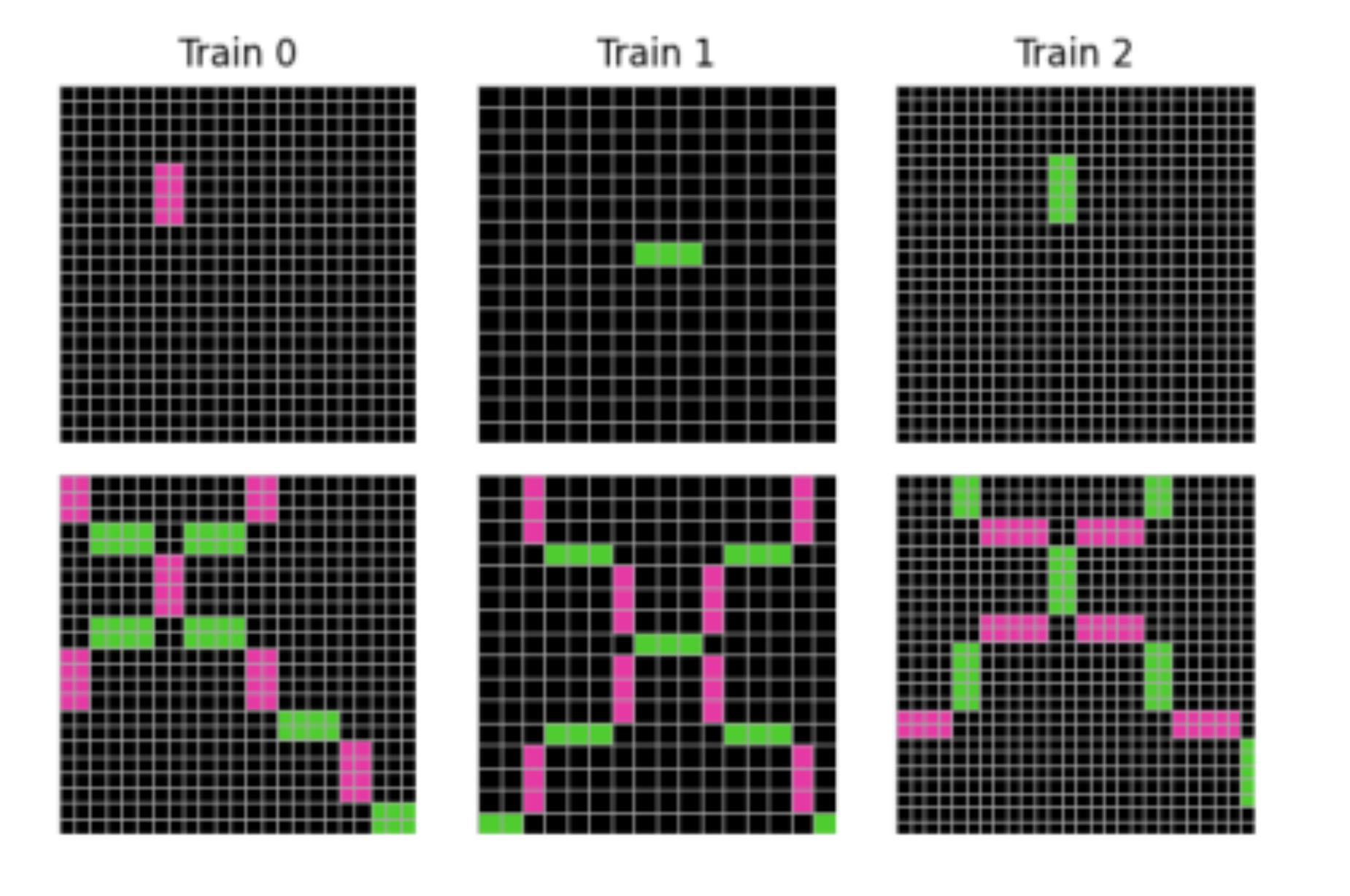}
        \caption{Second-nearest neighbor.}
    \end{subfigure}
    \caption{Nearest neighbor search using cosine similarity on task embeddings.}
    \label{fig:memory-sym}
\end{figure}

\paragraph{External Memory for Cross-Task Retrieval.}
To enhance generalization during TTT, we introduce an \textbf{external memory module} implemented as a vector database\footnote{We used FAISS as vector database (\url{https://github.com/facebookresearch/faiss}).}. Each ARC training task is embedded using the encoder of LongT5 with mean pooling:
\[
\mathbf{e}(X) = \frac{1}{T} \sum_{t=1}^{T} h_t,
\]
where $h_t$ are the encoder hidden states, i.e. the output tensor of the last encoder layer, of dimensions \texttt{(bs, seq\_len, emb\_dim)}. For a given test task, we retrieve nearest neighbors via cosine similarity: an example can be found in Figure~\ref{fig:memory-sym}.
\[
\mathrm{sim}(X_{\text{test}}, X_i) = \frac{\mathbf{e}(X_{\text{test}}) \cdot \mathbf{e}(X_i)}{\|\mathbf{e}(X_{\text{test}})\| \, \|\mathbf{e}(X_i)\|}.
\]

More details on the threshold we used for the consine similarity and the type of augmentations we did is summarized in Table~\ref{tab:sim-ablation}.

\begin{table}[htbp]
\centering
\begin{tabular}{ccccccccc|cc}
\hline
Threshold & Top\_k & Many\_sim & Aug\_0 & Aug\_1 & Aug\_2 & Aug\_3 & Aug\_counts & $\Delta$ Score & $\Delta$ Upper \\
\hline
0.03  & 3 & True  & True  & True  & True  & True  & 5,7,7,7,7    & -0.47 & +0.84 \\
0.03  & 2 & True  & True  & True  & True  & True  & 5,7,7,7,7    & -0.47 & -1.42 \\
0.045 & 2 & True  & True  & True  & True  & True  & 5,7,7,7,7    & -2.54 & -0.95 \\
0.045 & 3 & True  & True  & True  & True  & True  & 5,7,7,7,7    & \textbf{0.00} & +1.03 \\
0.03  & 2 & True  & False & False & False & False & -            & +0.66 & +0.28 \\
0.03  & 3 & True  & False & True  & True  & True  & 7,3,3,7,10   & -2.73 & -1.42 \\
0.03  & 2 & True  & False & True  & True  & True  & 7,3,3,7,10   & +0.94 & +0.56 \\
0.03  & 4 & False & False & False & False & True  & 10,3,3,3,27  & -1.88 & -2.83 \\
0.03  & 2 & True  & True  & False & False & False & 5,15,15,15   & -1.32 & -0.38 \\
0.03  & 2 & True  & False & True  & False & False & 5,15,15,15   & +0.09 & \textbf{0.00} \\
0.03  & 2 & True  & False & False & True  & False & 5,15,15,15   & -1.04 & -0.29 \\
0.03  & 2 & True  & False & False & False & True  & 5,15,15,15   & +0.66 & +1.41 \\
\hline
\end{tabular}
\caption{Performance deltas relative to SOTA baseline (Score = 45.48, Upper bound = 67.33). The Aug\_i corresponds to the different augmentations techniques described at \ref{item:mem}.}
\label{tab:sim-ablation}
\end{table}

Retrieved tasks are incorporated into TTT in three modes:
\begin{enumerate}
    \label{item:mem}
    \item \textbf{Data Expansion:} retrieved tasks are added as extra training samples,
    \item \textbf{Pair Augmentation:} only individual I/O pairs are added, the goal being to enable the model to extract common information from the task and similar ones,
    \item \textbf{Evaluation-Guided TTT:} retrieved pairs are used only for validation to guide early stopping.
\end{enumerate}
There are multiple augmentations:
\begin{enumerate}
    \label{item:mem}
\item \textbf{Many\_sim:} Use the leave-one-out approach to create many tasks from a similar task as in the classic TTT

\item \textbf{Aug\_0:} Take a leave-one-out task created from a similar tas* as the base. Then append to it one input/output (I/O) pair from the online task for training and one I/O pair for testing. The new task will have two test pairs.

\item \textbf{Aug\_1:} Take a leave-one-out task from the online task as the base. Then append to it one I/O pair from a similar task for training and one I/O pair for testing.  
The appended I/O pairs may come from either the training or test set of the similar task.  
This augmentation is not applicable if the similar task has no test pairs (e.g., tasks from the 240 online set).

\item \textbf{Aug\_2:} Same as Aug\_0, but append more I/O pairs (two for training and one for testing).

\item \textbf{Aug\_3:} Same as Aug\_1, but when the similar task is one of the 240 online tasks, use only its training pairs.

\end{enumerate}

\begin{figure}[htbp]
    \centering
    \includegraphics[width=0.6\textwidth]{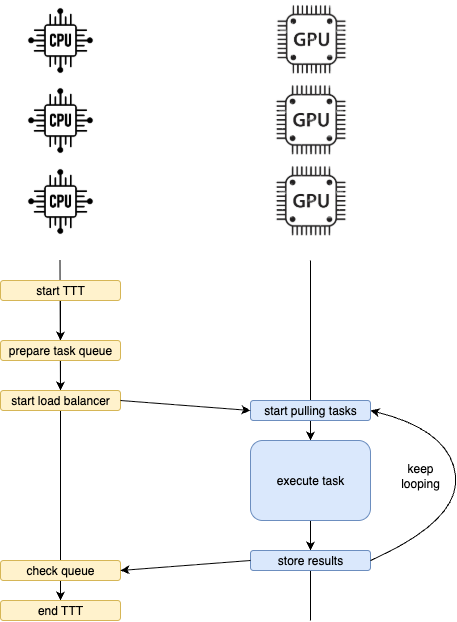}
    \caption{Load balancing for parallel TTT execution across GPUs.}
    \label{fig:ttt}
\end{figure}

\paragraph{Parallelism and Dynamic Load Balancing.}
TTT is embarrassingly parallel since each ARC task is adapted independently. To maximize GPU utilization, we designed a custom load balancer that dispatches TTT jobs dynamically across GPUs. Each GPU is managed by a separate CPU process; once a GPU finishes a task, it is immediately assigned the next job from the queue (Figure~\ref{fig:ttt}).

This strategy ensures near-constant GPU saturation across distributed environments and reduces idle time during fine-tuning.

\subsection{Decoding Strategy}
\label{sec:decoding}

The decoding process converts the model's token-level probability distribution into a complete output grid. Since ARC requires pixel-perfect predictions, decoding plays a crucial role in exploring plausible hypotheses while maintaining computational efficiency. In this section, for ease of reading and to avoid misunderstandings, we provide a unified graph-based formalism for decoding and describe how standard strategies such as greedy decoding, sampling, beam search, depth-first search (DFS), and breadth-first search (BFS) fit within this framework.

During the development of this work, we often changed from sampling to BFS to beam search, as they ultimately gave comparable results: this is the reason behind this review. We will provide more details on the results in the section about ablation studies~\ref{sec:ablations}.

\subsubsection{Graph Formalism for Decoding}
We formalize the decoding process as a \emph{search on a prefix graph}. Let $\mathcal{V}$ denote the token vocabulary and let a decoded sequence $y = (y_1, \ldots, y_T)$ represent the serialized output grid. At decoding time step $t$, the model defines a conditional probability distribution over the next token:
\[
p_\theta(y_t \mid y_{<t}, x),
\]
where $x$ is the serialized ARC prompt.

We define a directed graph $\mathcal{G} = (\mathcal{N}, \mathcal{E})$ where:
\begin{itemize}
    \item Each node $n_t \in \mathcal{N}$ represents a prefix sequence $(y_1, \ldots, y_t)$.
    \item Each directed edge $e_{t \rightarrow t+1}(v) \in \mathcal{E}$ corresponds to appending token $v \in \mathcal{V}$:
    \[
    n_{t+1} = n_t \circ v.
    \]
    \item Each edge is weighted by the log-probability:
    \[
    w(e_{t \rightarrow t+1}(v)) = \log p_\theta(v \mid n_t, x).
    \]
\end{itemize}
A full decoded sequence corresponds to a path from the root node $n_0$ to a terminal node $n_T$ of depth $T$.

\subsubsection{Decoding as Graph Search}
Using this formalism, decoding methods correspond to different ways of exploring the graph $\mathcal{G}$:
\begin{itemize}
    \item \textbf{Greedy Decoding:} At each step, pick the outgoing edge with maximum probability:
    \[
    y_t = \arg\max_{v \in \mathcal{V}} p_\theta(v \mid y_{<t}, x).
    \]
    This corresponds to selecting the best local edge without backtracking.
    
    \item \textbf{Top-$k$ or Nucleus Sampling:} Instead of selecting the most probable token, sampling is done from the truncated distribution:
    \[
    \tilde{p}(v) \propto
    \begin{cases}
        p_\theta(v \mid y_{<t}, x) & \text{if } v \in \text{Top-}k \text{ or } v \in \text{nucleus}(p), \\
        0 & \text{otherwise}.
    \end{cases}
    \]
    This introduces stochastic exploration in graph traversal.
    
    \item \textbf{Beam Search:} Maintains the best $B$ hypotheses at each decoding step by expanding all active prefixes and selecting the top $B$ by cumulative log score:
    \[
    S(y_{1:t}) = \sum_{i=1}^{t} \log p_\theta(y_i \mid y_{<i}, x).
    \]
    This corresponds to a bounded best-first search over $\mathcal{G}$.
    
    \item \textbf{Depth-First Search (DFS):} Traverses each path to depth $T$ before backtracking. Useful when tree depth is small or when we want full candidate enumeration.
    
    \item \textbf{Breadth-First Search (BFS):} Expands all prefixes layer by layer. Equivalent to exploring all hypotheses of length $t$ before considering $t+1$.
\end{itemize}

\subsubsection{Entropy-based branching approach}

The graph-based formalism inspired us a branching approach to generate multiple candidates: the core idea is to estimate the uncertainty of the model’s predictions by computing the entropy of the softmax distribution for each generated token. Given the probability distribution over the vocabulary at a given decoding step,
\[
p_i = \mathrm{softmax}(z_i) = \frac{e^{z_i}}{\sum_j e^{z_j}},
\]
where $z_i$ is the logit value for the $i$-th word in the vocabulary. The entropy is defined as
\[
H(p) = -\sum_i p_i \log p_i,
\]
where $H(p)$ measures the uncertainty of the model’s output distribution.

A low entropy value indicates that the probability mass is concentrated on a few tokens (high model confidence), whereas a high entropy value suggests a more uniform distribution (low confidence). To control branching during generation, we define an entropy threshold $\alpha$. If $H(p) < \alpha$, the model is considered confident, and no branching occurs. Conversely, if $H(p) \geq \alpha$, the model is deemed uncertain, and we allow branching to explore multiple possible continuations. $\alpha$ is a hyperparameter to be calibrated.

Conceptually, this method can be seen as a deterministic analogue of stochastic sampling. While sampling introduces diversity through randomness, entropy-based branching uses a principled measure of uncertainty to decide when to diversify the generation path.

Empirically, this approach produced results comparable (if not slightly superior) to standard sampling-based methods. However, due to the additional computational overhead of per-token entropy computation and dynamic branching decisions, it ran slightly slower. Consequently, we did not include this method in our final submissions.

We also explored similar methods, leveraging the observation that for most grids generated by the model, there are only a small number of high-entropy tokens, typically the first time the model encounters a key object or a key position for the task logic. This enabled us to increase compute significantly for these high-entropy tokens (for example, also using the log-probabilities for a different transform of the task) while limiting overall compute increase to a reasonable amount. However, the conclusion was the same as that for entropy-based branching, the slowdown not being worth the marginal gains.

\subsubsection{Decoding Parameters and Temperature}
The logit distribution can be reshaped using a temperature parameter $\tau > 0$:
\[
p_\theta^\tau(v \mid y_{<t}, x) = \frac{\exp\left(\frac{z_v}{\tau}\right)}{\sum_{v'} \exp\left(\frac{z_{v'}}{\tau}\right)},
\]
where $z_v$ are the unnormalized logits. Lower temperature ($\tau < 1$) makes decoding more deterministic, while higher temperature increases diversity.

We also use output caching across layers to avoid recomputing hidden states during batched beam search, resulting in speedups during inference.

\subsubsection{Grid Traversals During Decoding}
Decoded token sequences are mapped back to $H \times W$ grids. Traversals have already been introduced in Section~\ref{sec:traversal}: in decoding we make use only of the row-by-row one.

\subsubsection{Handling Long Contexts}
Since ARC serialization may exceed 8k tokens, to better handle all pixels in the context we leveraged:
\begin{itemize}
    \item Efficient key-value caching for autoregressive decoding,
    \item Chunked decoding with LongT5’s transient attention (Section~\ref{sec:longt5}),
    \item Memory-augmented decoding using retrieval hints when available (Section~\ref{sec:ttt}).
\end{itemize}

\subsubsection{Speculative decoding}
\label{sec:speculative_decoding}

To accelerate decoding, we explore a lightweight form of \emph{speculative decoding} in which the usual assistant model (or draft model) is replaced by a simple transition matrix that predicts likely next tokens based on local token statistics. This allows us to propose multiple candidate continuations in parallel before validating them with the base model.

\paragraph{Token Transition Matrix.}
We construct a small transition matrix $M \in \mathbb{R}^{144 \times 12}$, where the output vocabulary consists of $12$ symbols corresponding to the $10$ ARC colors plus the two structural markers \texttt{<|start\_row|>} and \texttt{<|end\_row|>}. Each row of $M$ corresponds to one of the $144$ possible ordered token pairs, and each column corresponds to a possible next token. Given a pair of consecutive tokens, we encode it as a one-hot vector $X \in \mathbb{R}^{144}$ indicating its index in the transition matrix. The transition distribution over the next token is then computed as:
\[
Y = M^\top X, \quad \text{with } \sum_{i=1}^{12} Y_i = 1.
\]
Each column of $M$ is normalized to represent a probability distribution, making $Y$ a valid categorical distribution for next-token prediction.

\paragraph{Speculative Expansion.}
Instead of sampling a single next token, we select the top-$k$ tokens from $Y$ (typically $k \in \{2,3,5\}$ depending on available parallelism) to generate $k$ speculative continuations in parallel. The next token pairs are formed by shifting the context window over the most recent two tokens, now including the speculated candidates. Repeating this procedure yields a branching speculative tree:
\[
k \;\rightarrow\; k^2 \;\rightarrow\; k^3 \;\rightarrow \dots
\]
For example, with $k=3$, we produce $3$ candidates after one step, $9$ after two steps, and $27$ after three steps. These candidate prefixes are then verified by performing a batched forward pass through the base model, exactly as in standard speculative decoding. If the transition matrix $M$ is well calibrated, it often predicts the next $2$--$3$ tokens correctly, resulting in substantial decoding speedups. Indeed, the average accuracy on our test data is 84\%.

\paragraph{Learning the Transition Matrix.}
The transition matrix $M$ is built \emph{per task} using a simple frequency-based procedure. We initialize $M$ to zero and iterate once over all tokens from the input and output grids of the current task -- a task may be the original one or one of its augmented variants (rotations, flips, etc.). For each consecutive triplet $(t_{i-1}, t_i, t_{i+1})$, we:
\begin{enumerate}
    \item map $(t_{i-1}, t_i)$ to a row index $r$ of $M$,
    \item map $t_{i+1}$ to a column index $c$,
    \item increment $M_{r,c} \leftarrow M_{r,c} + 1$.
\end{enumerate}
Since each grid contains hundreds of such triplets, and $144 \times 12 = 1728$ possible entries, $M$ is quickly populated with meaningful statistics even at the task level. The matrix is constructed only once per task.

\paragraph{Parallel Speculation Across Beams.}
During beam search, each beam evolves independently and may follow different token trajectories. Therefore, speculative decoding must operate in parallel across all active beams. Efficient batching becomes essential, as the joined candidate set grows with both $k$ (speculative width) and beam size $B$. Thus, practical speedup depends on:
\[
\text{Batch size} = B \times k^d,
\]
where $d$ is the number of speculative steps. This requires careful engineering both as we need to modify a beam search algorithm and to ensure that the batched forward pass remains within hardware limits while maximizing GPU utilization. Ultimately, we got an impressive speed-up of $1.598 \times$.

\paragraph{Summary.}
This approach replaces a learned speculative model with a compact, interpretable transition matrix $M$ derived directly from the current task. It enables fast, batched next-token hypotheses with minimal overhead and integrates smoothly into our beam search framework. When combined with speculative verification by the base model, it offers a promising path toward efficient and accurate decoding for ARC.

\subsubsection{Remarks on BFS}

We experimented with BFS and DFS decoding as in \cite{Franzen2024}, which means exploring the tree of possible candidates, pruning branches that fall below a probability threshold. Using thresholds similar to those reported for ARC-AGI 1, i.e 10-17\%, decoding ran two to three times faster than our beam search configuration, but the performance dropped significantly. To recover good enough scores, we had to lower the threshold to around 1\%, at which point decoding became slightly slower than beam search (with some room for optimization).

Interestingly, even with a 1\% threshold, BFS still scored about 0.5-1\% lower than beam search across multiple evaluation datasets. We found that sometimes beam search produces the true solution with very low probability, sometimes as low as 0.1\% (and only for one rigid transformation of the task). Some of these generated true solutions still appear in top 2 after scoring. For this reason, most of our submissions used beam search in decoding rather than BFS/DFS.

\subsection{Filtering}

Decoding in ARC-AGI yields a large number of candidate grids, many of which are syntactically valid but logically inconsistent with the task. To constrain the hypothesis space and enforce basic task consistency, we apply a set of \textbf{white-box filtering priors} before scoring or selecting solutions. We have already analyzed them in Section~\ref{sec:filtering}: we mention the impact before moving onto the scoring section.

\paragraph{Impact.}
In practice, filtering reduces candidate sets by 35--55\% before scoring (removing only $\sim 1\%$ correct solution across the entirety of our held-out test set), significantly improving the scoring time and results stability. Since these priors are task-agnostic, they generalize well and do not require learned parameters.

% \subsection{Scoring}
% \begin{itemize}
%     \item Scoring strategies overview
%     \item Traversal ranking strategies
%     \item Symmetry-based scoring (mini-arch)

% \begin{figure}[htbp]
%     \centering
%     \includegraphics[width=0.8\textwidth]{figures/symmetry.png} 
%     \caption{Illustration of the idea behind the scoring mechanism. Providing the task to the model in different orientations allows the model to return different scores for the most problematic pixels. In this example, the wrong blue pixels are detected when the task is rotated 180 degrees, while it is not visible for 90 degrees rotations. Picture taken from \cite{Franzen2024}}
%     \label{fig:architects-scoring}
% \end{figure}
    
%     \item Efficiency considerations
% \end{itemize}

\subsection{Scoring}
\label{sec:scoring}

Once a set of candidate solutions has been generated by the decoding pipeline, the final step is to \emph{select} the top two most plausible output grids. Since ARC-AGI scoring is all-or-nothing per task, the ranking strategy used to select candidates is crucial.

\paragraph{Baseline Scoring by Occurrence.}
A naïve and widely used approach is to rank candidates by \emph{occurrence frequency}: the intuition is that the most frequently generated grid across augmented decoding runs is likely to be correct. In case of ties, we break them using the \emph{cumulative log-likelihood} of each grid (computed during beam search). Formally, given a set of candidates $\{y_k\}$, we compute:
\[
\text{rank}(y_k) = \big(\text{occurrence}(y_k),\; \sum_{t}\log p_\theta(y_{k,t} \mid y_{k,<t}, x)\big).
\]
This baseline ranking is efficient and generally effective but does not exploit geometric or structural priors of ARC tasks.

\paragraph{Traversal-Based Scoring Attempts.}
We also experimented with traversal-based scoring, where each candidate grid was scored under different serialization traversals (\texttt{row\_by\_row} and \texttt{snake}). The hypothesis was that encoding the same candidate from different perspectives may expose inconsistencies in ambiguous pixels. For 2-traversals model without UL2 training, changing scoring from vanilla \texttt{row\_by\_row} to using both traversals (\texttt{row\_by\_row} and \texttt{snake}) on all tasks with an additional refinement of the scoring on the top 5 candidates with \texttt{row\_by\_row} only, boosted our Kaggle score of $19.56\%$ by $\sim 2\%$. However, when the UL2 approach was introduced, scoring using \texttt{row\_by\_row} only performed better, leading to our final configuration.

\begin{figure}[htbp]
    \centering
    \includegraphics[width=0.8\textwidth]{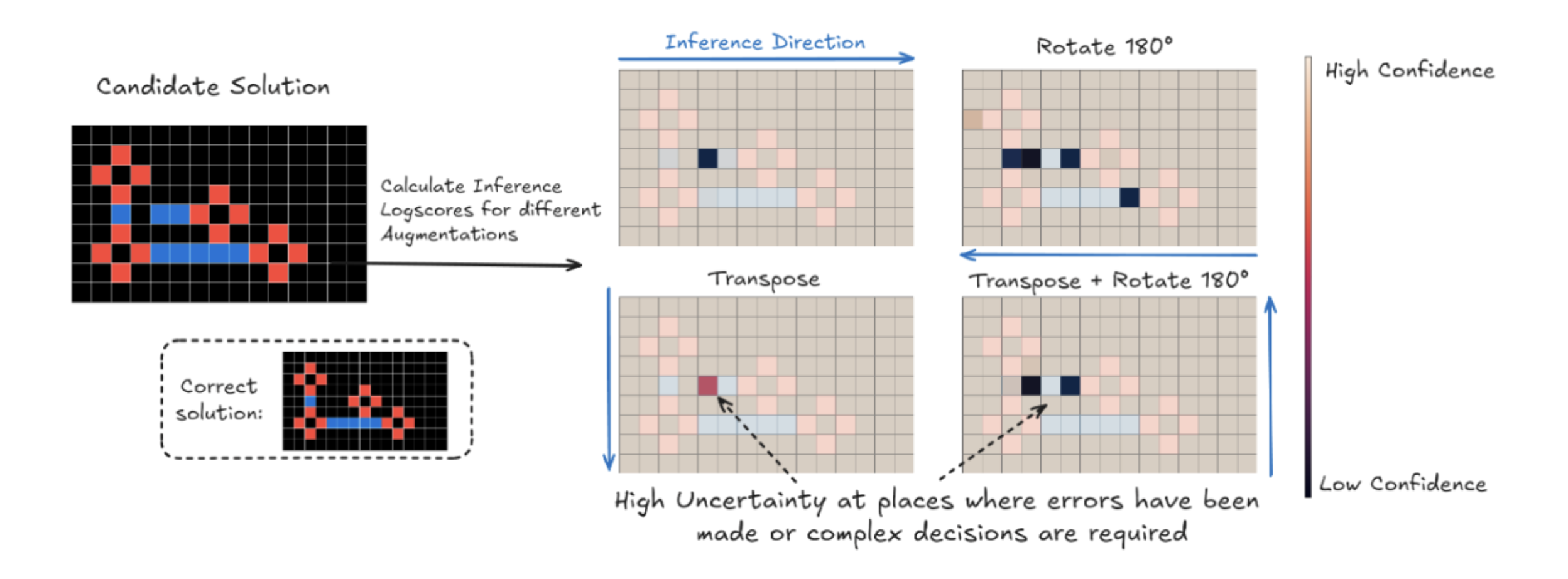}
    \caption{Symmetry-based scoring evaluates each candidate grid under multiple geometric transformations. Certain prediction errors become more evident under specific symmetries. Figure adapted from~\cite{Franzen2024}.}
    \label{fig:architects-scoring}
\end{figure}

\paragraph{Symmetry-Based Scoring (Mini-Arch).}
Inspired by \textit{The ARChitects}~\cite{Franzen2024}, the scoring strategy we adopted is \textbf{symmetry based} that evaluates each candidate across multiple transformed views of the task. The intuition is that a correct solution should remain consistent under symmetry transformations, while incorrect solutions exhibit unstable likelihoods across orientations (Figure~\ref{fig:architects-scoring}).

Let $D_4$ denote the dihedral group of 8 rigid symmetries (rotations and reflections). For a candidate grid $y_k$, we compute an aggregated score:
\[
S(y_k) = \sum_{T \in D_4} \log p_\theta\big(T(y_k) \mid T(x)\big),
\]
where each $(T(y_k),T(x))$ pair is re-encoded and passed through the model to obtain its log-likelihood. The final prediction is:
\[
\hat{y} = \arg\max_{y_k} S(y_k).
\]
This scoring method, which we call \textbf{Mini-Arch} in honor of The ARChitects, dramatically improves stability by enforcing viewpoint consistency.

\paragraph{Two-Stage Selection.}
To keep computation manageable, we perform scoring in two stages:
\begin{enumerate}
    \item \textbf{Preselection:} rank candidates by occurrence, breaking ties using cumulative log-likelihood, and keep only the top 50\% candidates.
    \item \textbf{Symmetry refinement:} apply Mini-Arch scoring to the top-50\% set and return the top-2 final predictions.
\end{enumerate}

\paragraph{Efficiency Considerations.}
Mini-Arch scoring introduces additional forward passes, but these are fully parallelizable and efficient on GPU. With 8 symmetry views, the method adds only $\sim 30$ seconds per task on an H100 while improving selection accuracy by over $25\%$ compared to occurrences-based ranking (more details is Section~\ref{sec:ablations}).

Overall, symmetry-aware scoring is a simple yet powerful mechanism that injects geometric priors into selection and significantly enhances robustness in ARC reasoning.

\section{Monitoring, Logging, and Job Scheduling}
\label{sec:monitoring}

Robust monitoring and logging were essential to manage large-scale distributed training across our compute cluster. This infrastructure ensured full experiment reproducibility, real-time observability, and efficient scheduling across thousands of GPU hours.

\subsection{Infrastructure Overview}

\begin{figure}[htbp]
    \centering
    \includegraphics[width=0.9\textwidth]{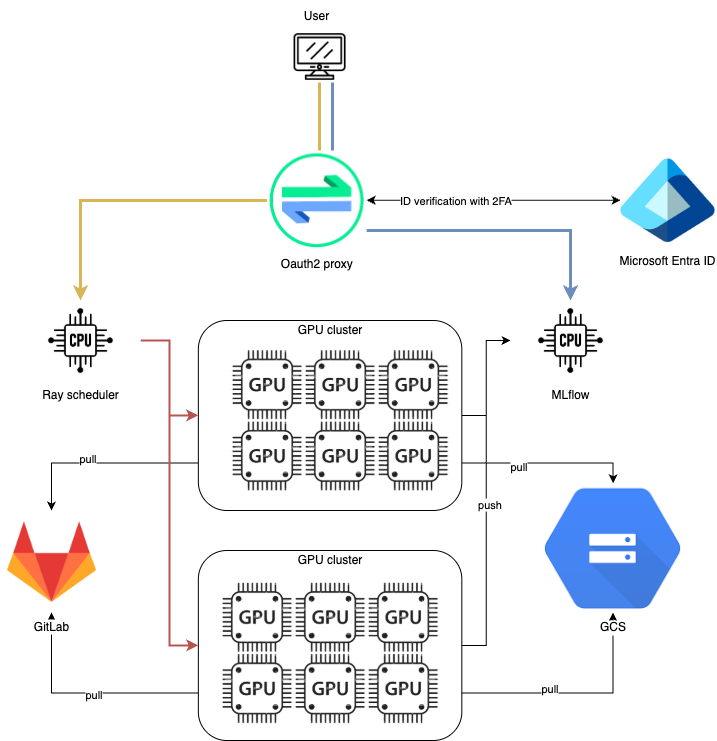} 
    \caption{Basic diagram describing our basic AI infra.}
    \label{fig:infra}
\end{figure}

Our training environment was orchestrated using a \textbf{central Ray cluster}, consisting of:
\begin{itemize}
    \item One \textbf{Ray head node} acting as the control server.
    \item Multiple \textbf{compute nodes}, each equipped with \textbf{8$\times$H100 GPUs}.
\end{itemize}
The Ray cluster handled distributed execution, resource allocation, and job scheduling across offline and online training workloads. Logs and metrics generated by all nodes were collected in a centralized manner. A basic architecture diagram of the infra we used can be found in Figure~\ref{fig:infra}.

\subsection{Experiment Tracking with MLflow}

All experiments were tracked using an internal \textbf{MLflow server}. Each run stored:
\begin{itemize}
    \item Standard HuggingFace metrics: \texttt{train\_loss}, \texttt{eval\_loss}, \texttt{learning\_rate}, \texttt{gradient\_norm}, training epoch, wall-clock runtime, and throughput.
    \item System statistics: GPU utilization, memory usage, and disk I/O, gathered via Ray instrumentation.
    \item \textbf{Artifacts:} Full textual training logs were uploaded as MLflow artifacts for post-mortem debugging when failures occurred.
\end{itemize}

This setup enabled complete experiment reproducibility and easy comparison between model versions, attention kernels, and optimization strategies.

\subsection{Secure Access}

Both Ray and MLflow dashboards were protected using an \textbf{OAuth2 proxy layer} configured with \textbf{Microsoft Entra} authentication. This ensured:
\begin{itemize}
    \item Secure, authenticated access via single sign-on (SSO),
    \item Isolation from external users,
    \item Audit logging for compliance.
\end{itemize}

\subsection{Ray-Based Job Scheduling}

We implemented a custom \textbf{job scheduler on top of Ray} to coordinate experiment execution:
\begin{itemize}
    \item Users submit jobs with resource requests via a queue system.
    \item The scheduler allocates GPU nodes dynamically based on availability.
    \item Priority queues enable rapid iteration during debugging while maximizing cluster utilization.
\end{itemize}
This system was used for both long-running offline training and shorter evaluation workloads.

\subsection{Pixel-based Metrics and Visualizations}

Beyond standard metrics, we introduced ARC-specific diagnostic metrics. In particular, we implemented a \textbf{pixel-wise scoring simulator}:
\begin{itemize}
    \item After each offline training epoch, the current model was evaluated on a held-out ARC test set.
    \item For each predicted grid, we computed the percentage of correctly predicted pixels.
    \item The score distribution was visualized as a histogram showing model consistency across tasks.
\end{itemize}

\begin{figure}[htbp]
    \centering
    \includegraphics[width=0.6\textwidth]{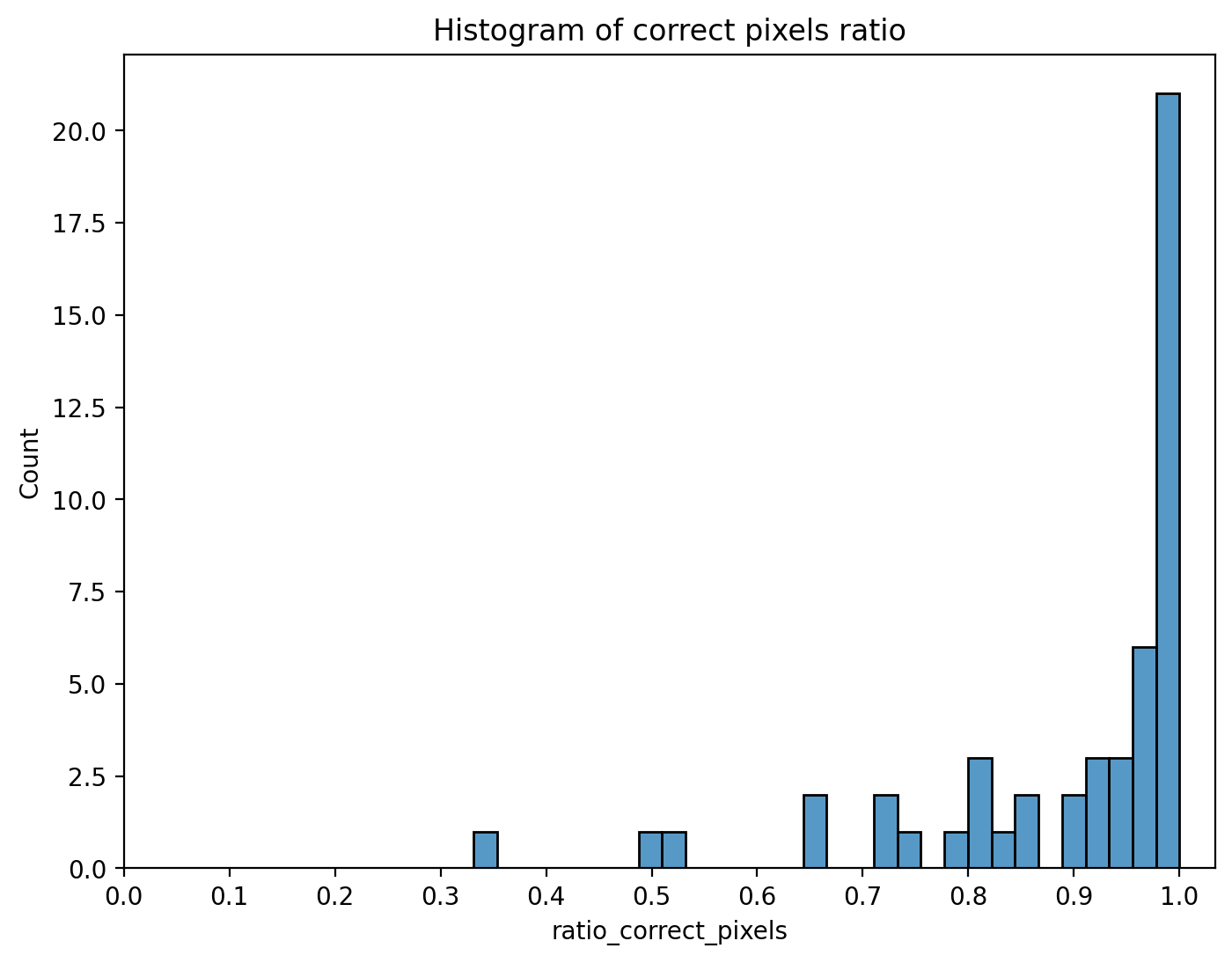} 
    \caption{Histogram of the distribution of correct pixels on our held-out test set of 50 tasks.}
    \label{fig:histo-res}
\end{figure}

The results of a subset of our held-out dataset of 50 tasks are shown in Figure~\ref{fig:histo-res}.

\subsection{Test-Time Training and Decoding Diagnostics: Likelihood-Based Metrics}
\label{sec:ttt_metrics}

During Test-Time Training (TTT), the model adapts its weights on a per-task basis using a small lora rank and a limited number of gradient steps. While this improves task specialization, it also introduces the risk of overfitting or drifting away from good initialization weights. To monitor and study this adaptation process, we introduced a set of \textbf{diagnostic metrics based on log-likelihood distributions}. These metrics were collected across the held-out dataset of 50 tasks and are visualized in Figure~\ref{fig:ttt_metrics}.

In particular, we analyze both:
\begin{itemize}
    \item \textbf{Training loss metrics:} raw cross-entropy loss and length-normalized loss,
    \item \textbf{Log-likelihood metrics:} average and maximum sequence log-likelihood across inference candidates.
\end{itemize}

By comparing these metrics across solved and unsolved tasks, we gain insight into whether probability distributions over tokens encode meaningful task structure that can guide adaptive learning.

\subsection{Metrics Definition}

For each task, we compute:
\begin{itemize}
    \item \textbf{Train loss} $\mathcal{L}_{\text{train}}$: average cross-entropy loss over the task sequence.
    \item \textbf{Normalized loss} $\mathcal{L}_{\text{norm}}$: training loss divided by sequence length to remove length bias.
    \item \textbf{Average cumulative log-likelihood}:
    \[
    \bar{\ell} = \frac{1}{T} \sum_{t=1}^{T} \log p(y_t \mid y_{<t}, x),
    \]
    \item \textbf{Maximum cumulative log-likelihood}: highest log-likelihood candidate among beam search generations.
\end{itemize}

\subsection{Observations}

\begin{figure*}[htbp]
    \centering
    \includegraphics[width=\textwidth]{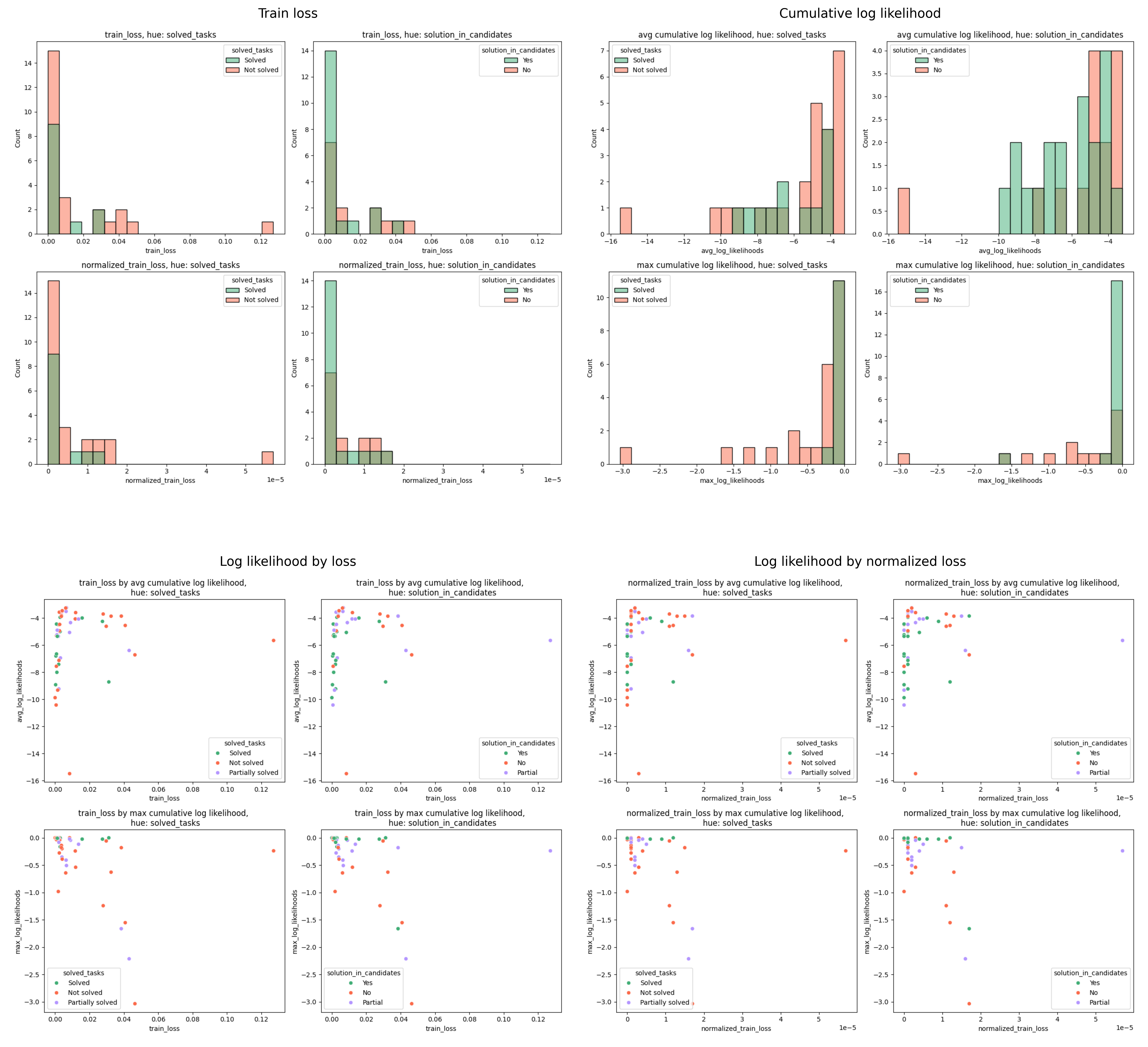}
    \caption{Distribution of train loss and log-likelihood metrics across ARC tasks during TTT and decoding. Green indicates solved tasks, red indicates unsolved tasks. Metrics are grouped by both solution correctness and presence of the true output among beam candidates.}
    \label{fig:ttt_metrics}
\end{figure*}

Figure~\ref{fig:ttt_metrics} shows the distribution of these metrics across three categories:
\begin{enumerate}
    \item Tasks solved by the model,
    \item Tasks where the correct output appeared among the candidate generations but was not ranked first,
    \item Tasks for which no correct hypothesis was generated.
\end{enumerate}

We observe that:
\begin{itemize}
    \item Solved tasks tend to cluster at lower training loss and higher log-likelihood values.
    \item Tasks where the correct output appears among candidate solutions are often distinguishable by high \emph{max log-likelihood}, even if train loss is suboptimal.
    \item The normalized loss $\mathcal{L}_{\text{norm}}$ shows stronger correlation with task success than raw train loss.
    \item Scatter plots reveal that log-likelihood is a \textbf{better predictor of solvability} than loss alone, especially when using beam search.
\end{itemize}

\subsection{Implications for Test-Time Training}

These findings indicate that:
\begin{itemize}
    \item Log-likelihood is a useful confidence signal and can guide adaptive decoding.
    \item Max log-likelihood can identify tasks where the model knows the correct answer but fails to select it—these tasks are ideal candidates for targeted refinement.
    \item Normalized loss is a more stable criterion for dynamic learning rate adjustment during TTT.
\end{itemize}

This helped detect overfitting and measure early task-level reasoning ability, even when exact solutions had not yet been achieved.

\subsection{Reproducibility}
\label{sec:reproducibility}

Reproducibility is a critical requirement for fair comparison and scientific rigor in ARC-AGI research. Due to the stochastic nature of deep learning pipelines -- especially when involving GPU computation and distributed training -- it is essential to control randomness across all components of the system.

To ensure reproducible experiments, we explicitly set seeds for all relevant libraries and enforce deterministic execution where possible. The function below was used at the beginning of every training, fine-tuning, and evaluation script:

\begin{verbatim}
def set_seed(seed: int = 42) -> None:
    random.seed(seed)  # Python random module
    np.random.seed(seed)  # Numpy module
    os.environ["PYTHONHASHSEED"] = str(seed)  # Python environment
    transformers_set_seed(seed)  # Transformers library
    torch.manual_seed(seed)  # PyTorch
    torch.cuda.manual_seed(seed)  # CUDA RNG
    torch.cuda.manual_seed_all(seed)  # Multi-GPU
    torch.backends.cudnn.deterministic = True  # Deterministic ops
    torch.backends.cudnn.benchmark = False  # Disable heuristics
\end{verbatim}

Despite using best practices for reproducibility, we observed \textbf{small fluctuations across runs}, typically within \textbf{$\pm 1\%$ on key metrics}. We attribute this variability to:
\begin{itemize}
    \item Non-deterministic CUDA kernels used by PyTorch in certain matrix operations,
    \item Minor differences in kernel scheduling on multi-GPU hardware,
    \item Floating-point accumulation order during distributed computation.
\end{itemize}

These variations do not affect the qualitative conclusions of this work, and model rankings remained stable across repeated runs. All experiments were executed under fixed seeds, and hyperparameter configurations are made available to ensure reproducible results (see Appendix~\ref{app:yaml}).

\section{Results}
\label{sec:results}

We evaluated our full ARC pipeline on a private human-curated evaluation set of \textbf{177 ARC-style tasks} (as mentioned in \ref{sec:data:human}). This dataset is fully disjoint from ARC-AGI-1 and ARC-AGI-2 data and was constructed to test generalization under previously unseen abstractions. In this section we report overall system performance using \emph{pass@$k$} metrics, along with statistical analysis of candidate quality and model behavior.

\subsection{Overall Performance (pass@\texorpdfstring{$k$}{k})}

Table~\ref{tab:passatk} summarizes our pass@$k$ results, where a task is considered solved if any of the top-$k$ submitted candidates exactly matches the ground truth output.

\begin{table}[htbp]
\centering
\begin{tabular}{c|c}
\hline
\textbf{$k$ (number of attempts)} & \textbf{pass@$k$ (\%)} \\
\hline
1 & 35.12 \\
\bf{2} & \bf{45.99} \\
3 & 51.79 \\
4 & 52.54 \\
5 & 55.93 \\
\hline
\end{tabular}
\caption{pass@$k$ results computed over 177 evaluation tasks. Internal best performance.}
\label{tab:passatk}
\end{table}

These results show strong improvements as $k$ increases, indicating that our decoding and scoring pipeline generates diverse candidate outputs that progressively cover the search space.

\subsection{Candidate Quality Analysis}

\begin{figure}[htbp]
    \centering
    \includegraphics[width=0.75\textwidth]{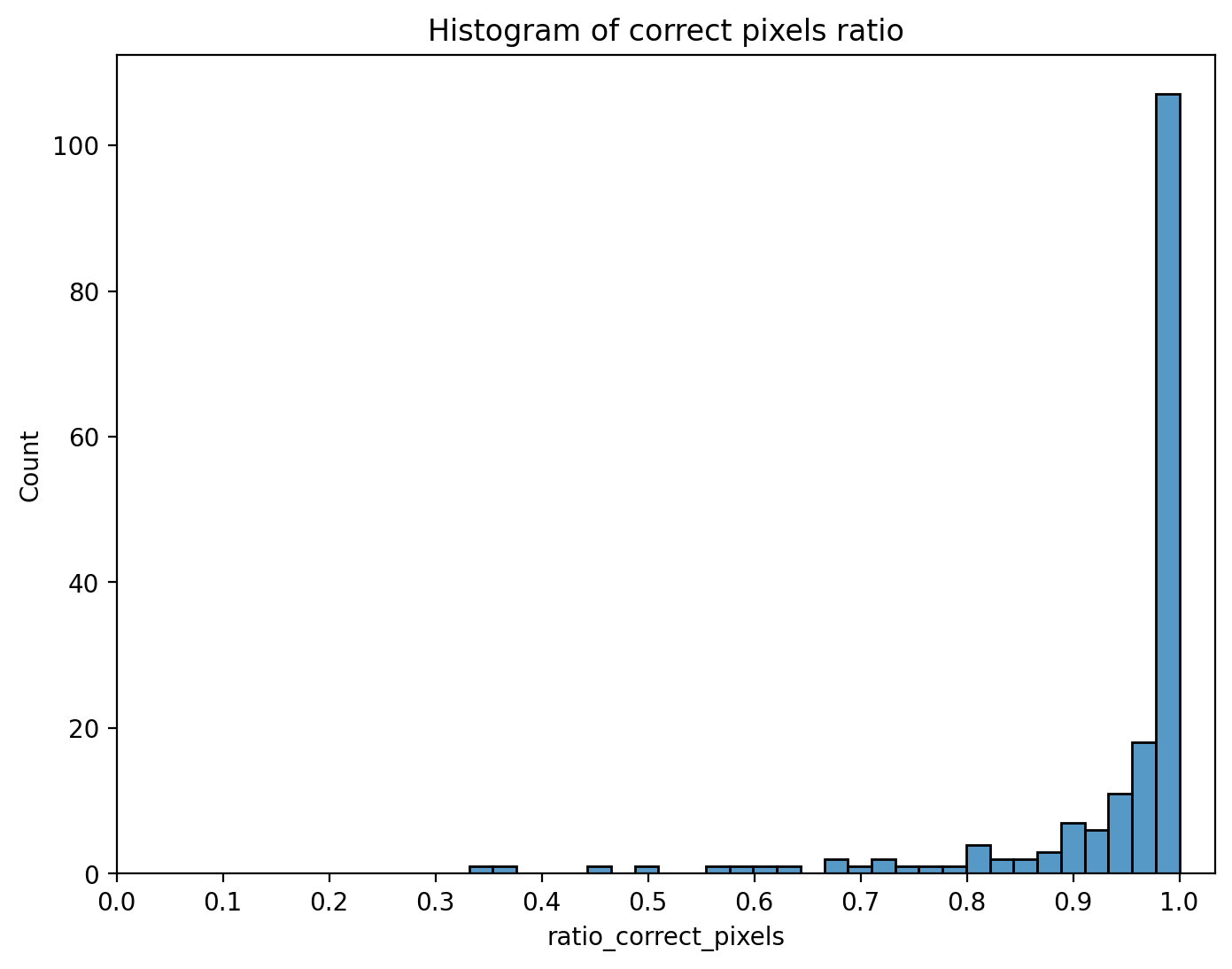}
    \caption{Histogram of pixel-wise accuracy for best candidate before final scoring.}
    \label{fig:pixel-accuracy-hist}
\end{figure}

To better understand model behavior beyond binary task success, we analyze the pixel-level accuracy of the best candidate generated per task before filtering and scoring. Figure~\ref{fig:pixel-accuracy-hist} shows the distribution of the \textit{ratio of correctly predicted pixels} across all 177 tasks.

The distribution is sharply skewed towards values close to $1.0$, indicating that \textbf{most incorrect attempts are very close to the correct solution}. This validates our strategy of generating and refining multiple hypotheses: the model often learns most of the underlying transformation, but small errors -- typically localized -- prevent task resolution. This reinforces the importance of candidate selection via structured scoring (Section~\ref{sec:scoring}).

\subsection{Likelihood and Loss Behavior in Test-Time Training}

\begin{figure}[htbp]
    \centering
    \includegraphics[width=\textwidth]{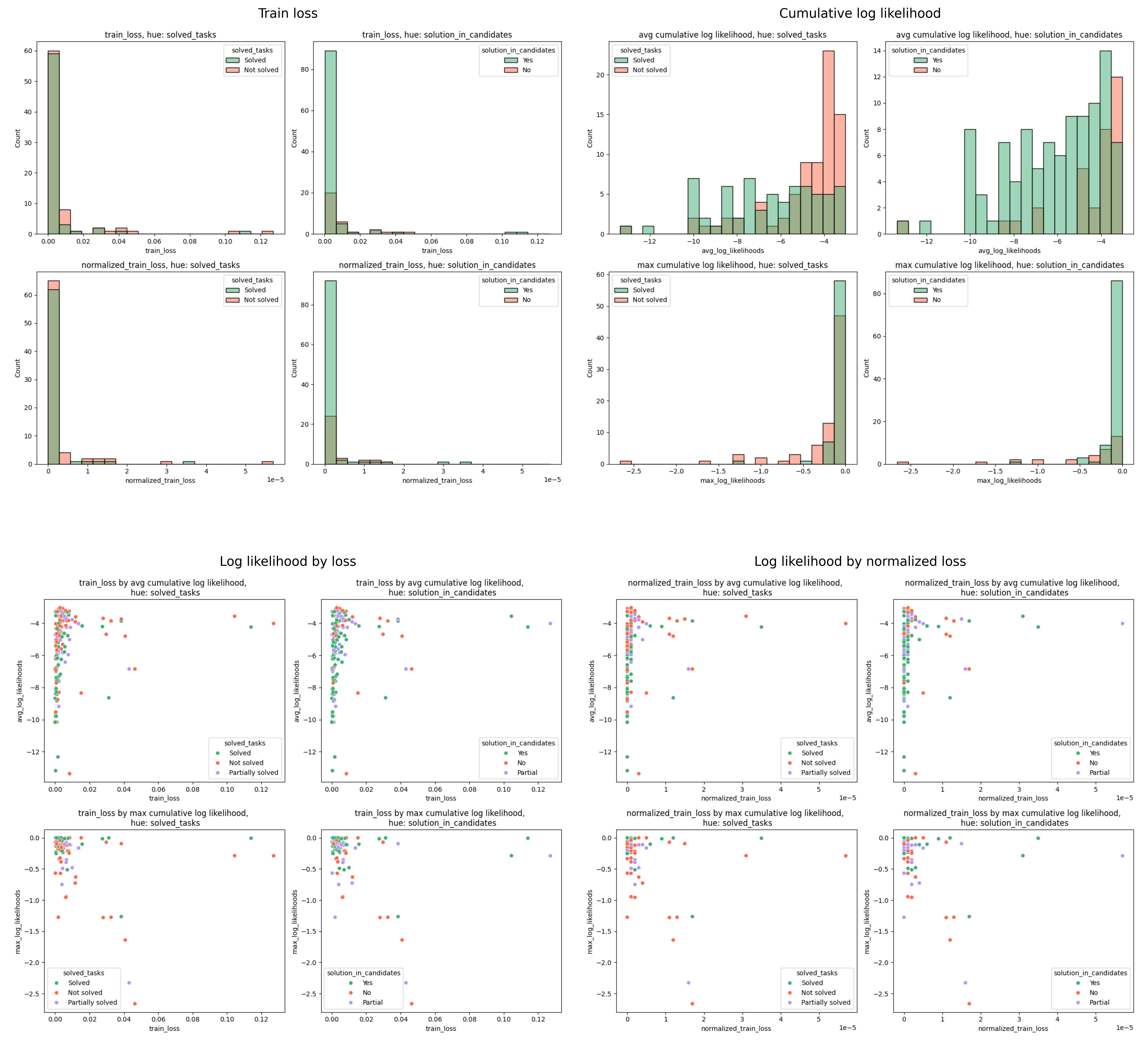}
    \caption{Distribution of TTT loss and candidate log-likelihood metrics across solved and unsolved tasks.}
    \label{fig:ttt-metrics}
\end{figure}

Figure~\ref{fig:ttt-metrics} reports a deeper analysis of per-task likelihood and training loss after Test-Time Training (TTT). Tasks are grouped by \emph{solved} vs. \emph{unsolved}, and by whether the correct grid appeared in the candidate set. Two key trends emerge:
\begin{itemize}
    \item \textbf{Solved tasks exhibit consistently higher log-likelihoods} for top candidates compared to unsolved ones.
    \item \textbf{Normalized loss is a weak predictor of task success}, suggesting that per-task TTT convergence is not sufficient—candidate consistency across symmetries is needed.
\end{itemize}

These findings support the use of log-likelihood-based ranking during scoring and motivate our symmetry-based aggregation introduced later.

\subsection{Upper Bound and Final System Accuracy}

Before scoring, we estimated an upper bound on achievable accuracy assuming perfect candidate selection. Considering all generated candidates before filtering, the upper bound was \textbf{69.02\%}. After removing invalid candidates using filtering rules (Section~\ref{sec:filtering}), the upper bound was reduced to \textbf{68.17\%}. 

Our final system -- after Test-Time Training, filtering, and Mini-Arch scoring -- achieved a \textbf{final pass@2 score of 46.99\%} on the 177-task benchmark, demonstrating that:
\begin{itemize}
    \item The decoding pipeline consistently generates correct hypotheses (\(\sim\)68\% coverage).
    \item Scoring and selection remain the main bottleneck for bridging the gap to upper bound performance.
\end{itemize}

\noindent
A run on $8 \times H100$ has these stats:
\begin{itemize}
    \item TTT time: 1.52 hours
    \item Decoding time: 0.66 hours
    \item Scoring time: 0.22 hours
    \item \textbf{Total runtime: 2.42 hours for 177 tasks (49.3 seconds per task)}
\end{itemize}

As per kaggle results, the same computations on $4 \times L4$ take around $11.7$ hours.

Overall, these results demonstrate that our system achieves strong performance while remaining computationally practical through efficient TTT and symmetry-based scoring.

In the next section we are going to detail the contribution of each step to the pipeline.

%% on the ablations, make sure to add the traversals/no traversals and maybe something about augmentations
\section{Ablation Studies}
\label{sec:ablations}

To understand the contribution of each component in our pipeline, we conducted a series of ablation experiments on our private evaluation benchmark: we used all 177 tasks. We systematically disabled or varied, one at a time, key components of the system and measured three metrics:
\begin{itemize}
    \item \textbf{Upper Bound (UB)} before filtering – measures decoding diversity; it is the theoretical coverage of the tasks.
    \item \textbf{Filtered Upper Bound} – UB after white-box filtering.
    \item \textbf{Final Score} – pass@2 after decoding, filtering, and scoring.
\end{itemize}

Each experiment setting was run five times with different seeds for statistical robustness. The default configuration uses: \textbf{beam search decoding}, \textbf{test-time training (TTT)}, \textbf{filtering enabled}, and \textbf{Mini-Arch symmetry-based scoring}. The ablated settings modify one component at a time.

\begin{figure}[hbpt]
    \centering
    \includegraphics[width=\textwidth]{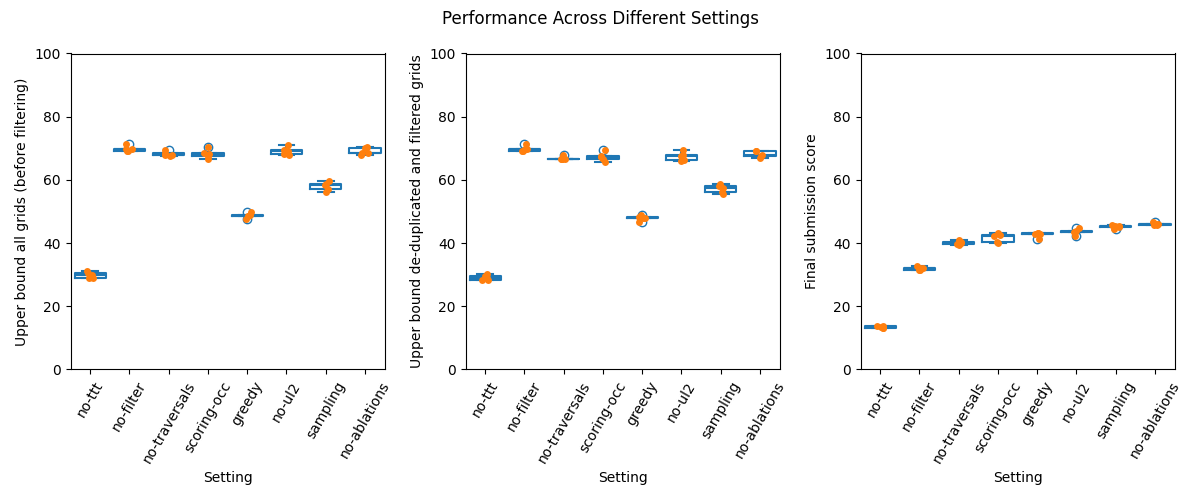}
    \caption{Ablation results across decoding, scoring, filtering, and TTT configurations and well as offline trainings with/without ul2 and traversals. Orange markers denote each performance across five seeds.}
    \label{fig:ablation-boxplot}
\end{figure}

Figure~\ref{fig:ablation-boxplot} summarizes the results.

\subsection{Effect of Test-Time Training (TTT)}
We compared full runs with TTT against runs without online finetuning (\texttt{no-ttt}). Removing TTT reduces the model’s ability to adapt to each task and significantly reduces the upper bound (diversity of high-quality hypotheses). Final score drops by around \textbf{33 percentage points}, confirming that \textit{TTT is essential for per-task generalization}.

\subsection{Effect of Filtering}
Removing symbolic filtering rules (\texttt{no-filter}) slightly improves the theoretical upper bound but causes a \textbf{catastrophic drop in final accuracy of around 14 percentage points}. This is because invalid candidate grids (wrong shape, colors, or containment violations) dominate decoding. Filtering is therefore \textit{crucial for pruning low-quality hypotheses before scoring}.

\subsection{Effect of Scoring Strategy}
We compared Mini-Arch scoring to a simple baseline that ranks candidates by occurrence frequency (\texttt{scoring-occ}). Performance decreased by \textbf{around 4 percentage points}, showing that \emph{symmetry-aware scoring is effective at selecting correct outputs when decoding produces many near-correct variants}. This reinforces the hypothesis that candidate consistency across viewpoints is a useful signal for reasoning.

\subsection{Effect of Decoding Strategy}
We compared three decoding methods:
\begin{itemize}
    \item \texttt{beamsearch} (default): $B=10$
    \item \texttt{sampling}: temperature $T=1.0$
    \item \texttt{greedy}: deterministic
\end{itemize}
Sampling produced a comparable upper bound with beam search but failed to improve the final score, confirming that \textbf{decoding diversity alone does not improve task accuracy without strong scoring}. Greedy decoding was the most brittle, dropping the Final Score by \textbf{around 2 percentage points}. Beam search provided the best balance between diversity and quality, supporting its use in ARC-style structured generation.

\subsection{Effect of Removing UL2 Training}
The configuration \texttt{no-ul2} (UL2 objective and denoise batches disabled during offline training) achieves one of the highest upper bounds (\(69.15\%\)) and remains competitive in final pass@2 accuracy (\(43.56\%\)), only \(\sim 2.5\%\) below the default full system. This indicates that while UL2-based masked modeling improves robustness and slightly stabilizes convergence, it is a welcomed upgrade but not a critical component for performance.

\subsection{Effect of Removing Traversal Representations}
In contrast, disabling grid traversal augmentation (\texttt{no-traversals}) during offline training causes a substantial drop in final performance (\(\sim 6\%\) compared to the default setting). Although the upper bound remains reasonably high (\(\sim 68\%\)), many promising candidates are incorrectly ranked during selection, leading to degraded pass@2 accuracy. This highlights the importance of providing the model with \textbf{multiple input perspectives}: different traversals (row-by-row vs.\ snake) encourage the model to focus on transformation rules rather than memorizing specific spatial layouts. These results support our broader claim that \emph{representation matters}, and that \textbf{variation in how data is presented improves abstraction and generalization}.

\subsection{Ablation Summary (Averaged over Seeds)}
\label{sec:ablation-avg}
For each configuration, we ran five seeds and report the mean $\pm$ standard deviation for: (i) upper bound before filtering, (ii) filtered upper bound, (iii) final pass@2 score, and (iv) total runtime on a single node with $8 \times H100$ GPUs. Results are reported in Table~\ref{tab:ablation-avg}.

\begin{table}[htbp]
\centering
\begin{tabular}{lcccc}
\hline
\bf Setting & \bf UB before filtering (\%) & \bf UB filtered (\%) & \bf Final score (pass@2, \%) & \bf Total time (h) \\
\hline
\bf no-ablations & 69.10 $\pm$ 1.09 & 68.14 $\pm$ 0.94 & \bf 46.08 $\pm$ 0.42 & 1.440 $\pm$ 0.010 \\
greedy & 48.74 $\pm$ 0.81 & 47.89 $\pm$ 0.81 & 42.75 $\pm$ 0.82 & 0.910 $\pm$ 0.014 \\
no-traversals & 68.29 $\pm$ 0.79 & 66.87 $\pm$ 0.58 & 40.04 $\pm$ 0.65 & 1.436 $\pm$ 0.021 \\
no-filter & \bf 69.83 $\pm$ 0.87 & \bf 69.83 $\pm$ 0.87 & 31.98 $\pm$ 0.61 & 1.562 $\pm$ 0.015 \\
no-ttt & 29.94 $\pm$ 0.87 & 29.10 $\pm$ 0.87 & 13.56 $\pm$ 0.28 & \bf 0.734 $\pm$ 0.005 \\
no-ul2 & 69.15 $\pm$ 1.35 & 67.46 $\pm$ 1.35 & 43.56 $\pm$ 0.81 & 1.386 $\pm$ 0.005 \\
sampling & 58.06 $\pm$ 1.35 & 57.21 $\pm$ 1.35 & 45.12 $\pm$ 0.43 & 1.010 $\pm$ 0.020 \\
scoring-occ & 68.19 $\pm$ 1.41 & 67.34 $\pm$ 1.41 & 41.67 $\pm$ 1.42 & 1.276 $\pm$ 0.021 \\
\hline
\end{tabular}
\caption{Ablation summary over 5 seeds per setting. Reported as mean $\pm$ std.}
\label{tab:ablation-avg}
\end{table}

\paragraph{Summary of Findings}

\begin{itemize}
    \item TTT is the single most impactful component for increasing candidate quality.
    \item Filtering is mandatory to prevent combinatorial error cascading.
    \item Symmetry-based scoring (Mini-Arch) consistently improves candidate selection.
    \item Beam search remains a strong default decoding strategy.
    \item Traversal augmentation contributes to robustness: \texttt{no-traversals} leads to a $\sim 6\%$ drop in accuracy.
    \item UL2 contributes modestly: \texttt{noul2} remains competitive, being $\sim 2.5\%$ worse than full training.
    
\end{itemize}

These ablations validate the structure of our pipeline: \emph{TTT improves hypotheses, filtering removes invalid ones, and symmetry scoring selects the best one.} Future work will explore mixed decoding strategies, traversal-aware scoring, and adaptive beam allocation.

\section{Conclusion and Generalization}
\label{sec:conclusion}

\subsection{Insights and Discussion}
In this work, we demonstrated that solving ARC-AGI benefits from combining \emph{representation priors}, \emph{test-time learning}, and \emph{symmetry consistency}. Perhaps the most important lesson is that \textbf{how information is presented to the model profoundly influences its reasoning ability}. Providing different \emph{perspectives} of the same task—through data augmentations, traversals, and geometric symmetries—improves robustness and reduces overfitting to a particular representation.

Related findings have emerged in other domains: multilingual tokenization studies show that the same information encoded under different symbol systems unlocks generalization \cite{alephalpha2023tfree}, while dynamic multi-view encoding has been shown to improve model robustness \cite{zhang2024reverse}. Our results support this observation in the visual-schematic domain of ARC, and suggest that \textbf{perspective-aware modeling is a key ingredient toward abstraction}.

\subsection{Strengths and Weaknesses of the Approach}
Our pipeline exhibits several strengths:
\begin{itemize}
    \item \textbf{Systematic generalization via symmetry}: By enforcing $D_4$ invariance in scoring, the model learns transformation rules rather than blatant memorization.
    \item \textbf{Localized adaptation through TTT}: Test-Time Training allows per-task specialization, enabling the system to infer new task rules dynamically.
    \item \textbf{Interpretable augmentation design}: Symmetries, automata, and traversals act as \emph{knowledge priors}, making the system more sampling efficient and prone to better generalization.
    \item \textbf{External memory integration}: Retrieval of similar tasks enhances adaptation and suggests long-term potential for fusion of retrieval and gradient updates.
\end{itemize}

However, limitations remain:
\begin{itemize}
    \item \textbf{The decoding step does not natively incorporate symmetries} and instead relies on post-hoc white-box filtering.
    \item \textbf{Scoring is still a bottleneck}: Candidate generation often contains the correct hypothesis, but ranking it requires costly per-task evaluation.
    \item \textbf{Pipeline complexity}: The multi-stage process introduces engineering challenges for latency-sensitive applications.
    \item \textbf{Heuristic elements remain}: Some rules (e.g.\ inclusion priors) require careful tuning and may not generalize uniformly.
\end{itemize}

\subsection{Generalization Beyond ARC}
The principles developed in this work extend naturally to several real-world domains:

\paragraph{Next-gen RAG.}
Our use of TTT with external memory retrieval aligns with recent work on \emph{test-time training and nearest-neighbour retrieval} \cite{hardt2023test}, where models learn useful behavior from nearest neighbors at inference time. The same mechanism may also apply to code generation, theorem proving, and formal verification.

\paragraph{Natural language processing.}
Symmetry-based scoring generalizes to text by evaluating semantic consistency across paraphrases, languages, or styles. For instance, a hypothesis should retain its meaning under translation or synonym replacement -- mirroring our geometric invariance in ARC. Automata-based augmentation can extend to structured synonym graphs or pattern-preserving rewrites in sequences.

\paragraph{Traversal representation.}
Recent work has shown that \emph{the order in which information is presented to a model can fundamentally affect its reasoning ability}. For example, \cite{zhang2024reverse} demonstrated that simply reversing the order of digits in arithmetic expressions significantly improves numerical accuracy in LLMs. This result provides an early indication of the importance of \emph{alternative traversals} -- processing the same content under different sequential layouts.

We argue that this idea generalizes far beyond arithmetic. Many real-world data structures are inherently two-dimensional or graph-like, yet are currently linearized in a naïve way for language models. Applying traversal-based representations to domains such as mathematical notation, chemical formulas, spreadsheets, or scientific tables may enable models to uncover latent structural relationships that are otherwise obscured by standard token orderings.

\paragraph{Computer vision and robotics.}
Symmetry augmentation is already state-of-the-art in image processing via equivariant networks. Our findings suggest that \emph{symmetry-consistent scoring} could improve scene understanding and planning in robotics by evaluating candidate trajectories under rotated or mirrored viewpoints.

\paragraph{Self-driving and temporal heteroscedastic systems.}
In safety-critical sequential tasks like self-driving decision making or financial forecasting, where the underlying distribution of the data may shift abruptly, it is essential to be able to adapt models quickly. TTT and data augmentation could help models learn new paradigm effectively, similar to what humans do \cite{wang2025test}.

\paragraph{Test-Time Training as an Alternative to Long Contexts.}
Our findings suggest that \textbf{TTT can be used as a substitute for long context windows}. Instead of requiring the model to store large amounts of context, small LoRA updates allow the model to \emph{internalize new information} with a few gradient steps. This paradigm has implications for resource-constrained deployments: smaller models can retrieve missing knowledge via \emph{micro-adaptation} rather than maintaining huge context buffers.

\subsection{Future Directions}
There are clear opportunities for expansion:
\begin{itemize}
    \item \textbf{Additional reward head}. We have observed how multitask learning improves the accuracy of the model: thus, we would like to push this further by creating an evaluation head, that is trained to evaluate the probability of the configuration to be correct while generating it (ideas similar to \cite{silver2017mastering}).
    \item \textbf{Self-supervised TTT}. Following \cite{wang2025test}, we plan to add auxiliary denoising objectives during TTT using multi-head architectures for continuous adaptation.
    \item \textbf{Streaming adaptation}. Combining LoRA-based TTT with streaming data seem to enable continual learning without catastrophic forgetting. We plan to extend the study to thoroughly assess the claim.
    \item \textbf{Autopilot orchestration}. Our pipeline already supports non-linear execution paths, where decisions to refine TTT, decoding, or scoring are made dynamically. A future “reasoning autopilot” may learn this control policy, leading us closer to controller models: this could be the new approach to recursion and routing.
\end{itemize}

\paragraph{Closing remark.}
ARC challenges us to move beyond pattern recognition toward genuine rule discovery. Our work shows that progress comes not just from scale, but from \textbf{structured variation, representation, and perspective}. We hope these insights contribute to more general forms of machine reasoning.

\appendix
\section{Appendix}

\subsection{Algorithms for data generation with Cellular Automata}

\label{app:datagen_automata_algos}
\begin{algorithm}
\caption{Cellular Automata data generation: schema 1}
\label{alg:CA_datagen_schema_1}
\KwIn{$(I)$ = task input grids, automata sampling bounds, N = no. tasks to generate}
\KwOut{$\bm{T}_{new}$ = list of N new tasks}
\SetKwFunction{CheckGeneratedTaskQuality}{CheckGeneratedTaskQuality}
Initialize $\bm{T}_{new}$ as an empty list\;
\While{\(len(\bm{T}_{new} <= N)\)}{
    sample new automata $a$\;
    \(O'\gets a(I)\)\;
    \If{\( O' \neq I \)}{
        \( T' \gets (I, O')\)\;
        \If{\(\CheckGeneratedTaskQuality(T')\)}{
            append $T'$ to $\bm{T}_{new}$
        }
    }
}
\Return $\bm{T}_{new}$
\end{algorithm}

\begin{algorithm}
\caption{Cellular Automata data generation: schema 2}
\label{alg:CA_datagen_schema_2}
\KwIn{$(I,O)$ = task input and output grids, automata sampling bounds, N = no. tasks to generate}
\KwOut{$\bm{T}_{new}$ = list of N new tasks}
\SetKwFunction{CheckGeneratedTaskQuality}{CheckGeneratedTaskQuality}
Initialize $\bm{T}_{new}$ as an empty list\;
\While{\(len(\bm{T}_{new} <= N)\)}{
    sample new automata $a$\;
    \(O'\gets a(O)\)\;
    \If{\( O' \neq O \)}{
        \( T' \gets (I, O')\)\;
        \If{\(\CheckGeneratedTaskQuality(T')\)}{
            append $T'$ to $\bm{T}_{new}$
        }
    }
}
\Return $\bm{T}_{new}$
\end{algorithm}

\begin{algorithm}
\caption{Cellular Automata data generation: schema 3}
\label{alg:CA_datagen_schema_3}
\KwIn{$(I,O)$ = task input and output grids, automata sampling bounds, N = no. tasks to generate}
\KwOut{$\bm{T}_{new}$ = list of N new tasks}
\SetKwFunction{CheckGeneratedTaskQuality}{CheckGeneratedTaskQuality}
\SetKwFunction{CheckAutomataInvertibility}{CheckAutomataInvertibility}
Initialize $\bm{T}_{new}$ as an empty list\;
\While{\(len(\bm{T}_{new} <= N)\)}{
    sample new automata $a$\;
    \(I'\gets a(I)\)\;
    \If{\( I' \neq I \)}{
        \If{CheckAutomataInvertibility(T = (I', I))}{
            \( T' \gets (I', O)\)\;
            \If{\(\CheckGeneratedTaskQuality(T')\)}{
                append $T'$ to $\bm{T}_{new}$
            }
        }
    }
}
\Return $\bm{T}_{new}$
\end{algorithm}

\begin{algorithm}
\caption{Cellular Automata data generation: schema 4}
\label{alg:CA_datagen_schema_4}
\KwIn{$(I,O)$ = task input and output grids, automata sampling bounds, N = no. tasks to generate}
\KwOut{$\bm{T}_{new}$ = list of N new tasks}
\SetKwFunction{CheckGeneratedTaskQuality}{CheckGeneratedTaskQuality}
\SetKwFunction{CheckAutomataInvertibility}{CheckAutomataInvertibility}
Initialize $\bm{T}_{new}$ as an empty list\;
\While{\(len(\bm{T}_{new} <= N)\)}{
    sample new automata $a$\;
    \(I'\gets a(I)\)\;
    \If{\( I' \neq I \)}{
        \If{CheckAutomataInvertibility(T = (I', I))}{
            \( O' \gets a(O)\)\;
            \If{\(O' \neq O\)}{
                \( T' \gets (I', O')\)\;
                \If{\(\CheckGeneratedTaskQuality(T')\)}{
                    append $T'$ to $\bm{T}_{new}$
                }
            }
        }
    }
}
\Return $\bm{T}_{new}$
\end{algorithm}

\clearpage

\subsection{The conjugation trick: locally invertible automata}
\label{app:conjugation}

The so-called ``conjugation trick" seeks to extend the logic of a task while maintaining its flavor. To achieve this, it is sufficient to find ``small perturbations" of the grids that are locally invertible. This explains our fourth data generation scheme from \ref{automata_data_gen}; the third scheme there is similar.

We first provide an example. Assume that the task involves taking all objects on the grid and translating them to the right until they hit the border. Suppose that we have an example where, in the input, all objects are filled. Let the perturbation involve removing all the interiors of the objects in the grid (note that this is invertible in this context: the inverse operation is filling the interiors). When we apply this perturbation to both the input and the output, we obtain an example where objects with empty interiors are translated to the right until they hit the border.

Formally, if $f$ is a function that solves the task $(I_k, O_k)$ (i.e. $f(I_k) = O_k$) and $g$ is a map on grids that is locally invertible on the inputs i.e., there exists a function $g^{-1}$ in the same category as $g$ (e.g. in the category of automata) such that $g^{-1}(gI_k) = I_k$, then we obtain a new task $(gI_k, gO_k)$ with the logic $gfg^{-1}$, which naturally ``transports" the logic of the original task to the new task. We note that if $f$ and $g$ commute then we obtain in fact new demonstrations of the original task.

Under mild heuristics, we generate a "small perturbation" automaton $g$ (also possibly using features). If $g$ turns out to be invertible like above in the automata category, we create a new task. To verify that $g$ is locally invertible we use a discrete search algorithm.   

Finally, we emphasize that no assumptions are made about $f$ in this method.

\[
\begin{tikzcd}
I_k \arrow[r, "f"] \arrow[d, "g"'] & O_k \arrow[d, "g"] \\
gI_k \arrow[r, "gfg^{-1}", dashed] \arrow[u, ""', bend left] & gO_k
\end{tikzcd}
\]

\subsection{Cellular Automata and pixel features}
\label{app:automata_features}

One of the challenges of using cellular automata (CA) is that the transformations applied to the grids can be overly complex and lack human-like qualities. To address this, we extended the automata approach to operate not only on single pixels but also on more general concepts such as symmetries, holes, objects, and other forms of prior knowledge. This is achieved through the use of so-called Pixel Features (PF).

Given a grid, a PF is a mask with the same dimensions as the grid. It can contain either boolean or integer values and is used to highlight specific properties of the grid. Figure~\ref{fig:example_automata_features.png}-top shows three examples of boolean PFs (blue = true, black = false): the internal region of the objects, a shadow projected onto the weight of each pixel, and the maximum rectangular region enclosing the objects. Using PFs, we can embed additional information, including human priors, directly into the grids.

Let the grid \( G \) be a two-dimensional array defined as:
\[
G = \{g_{ij} \mid g_{ij} \in \{0, 1, 2, \ldots, 9\}, \, 1 \leq i \leq M, \, 1 \leq j \leq N \}
\]
where \( g_{ij} \) represents the value at the \( i \)-th row and \( j \)-th column, \( M \) is the number of rows, and \( N \) is the number of columns. We formally define a pixel feature as a function \( F \) such that:
\[
F: \{0, 1, \ldots, 9\}^{M \times N} \to \mathbb{Z}^{M \times N}
\]
In case of boolean features, \((\text{false}, \text{true})\) are converted to \((0,1)\) so the definition still holds. We can compute \( n \) pixel features starting from the original grid and stack them as additional channels of \( G \), giving the \emph{augmented grid} \( G' \):
\[
G' = (G, F_0(G), ..., F_n(G))
\]
\( G' \) has in total \( n+1 \) channels.

We can extend the CA approach defined earlier to be applied to the augmented grid \( G' \) instead of the original grid. The neighbor \( l \) now is defined as:
\[
l = (i,j,k,c)
\]
where \( k \) is the channel index: if \( k=0 \) we have the standard automata checking the value of the target pixel on the original grid \( G \), whereas with \( k \neq 0 \) the target pixel is checked on the channel \( k \) related to the pixel feature \( k-1 \). An example of CA used with pixel features is reported in fig. \ref{fig:example_automata_features.png}-bottom.

\subsection{YAML Configuration}
\label{app:yaml}
\begin{verbatim}
experiment_name: "longt5_sota_torch_280_4xL4"
kaggle_mode: false
full_determinism: false
seed: 42
monitoring:
  mlflow:
    experiment_name: ${...experiment_name}
    # uri: "https://mlflow.giotto.ai"
  logging:
    logger_name: ${...experiment_name}
setup:
  data:
    offline:
      - name: "arc-public"
        version: "v1"
        variant: "mini"
        local_folder: "data"
    online:
      name: "arc-augmented"
      version: "v9"
      variant: "unseen120"
      local_folder: "data"
  model:
    name: "our_best_model"
    local_folder: "model"
inference:
  online_fine_tuning:
    time_limit_parallel_in_seconds: 3_000_000
    max_num_gpus_to_use: 8

    run_id_mlflow: empty
    log_level: "info"
    logger_name: "online_ft"
    enable_log_to_file: true

    dataset_dir: "./kaggle/input"
    n_dataloader_workers: 0
    dataloader_persistent_workers: false

    # dataset_category: "${...setup.data.online.version}-${...setup.data.online.variant}"
    dataset_category: newunseenred1
    start_index_tasks: 0
    end_index_tasks: 177
    sort_tasks_by: "total_processed_token"
    sort_tasks_order: "desc"
    traversals: ["row_by_row"]

    model_id: "model/${...setup.model.name}"
    use_flash_attention_encoder: true
    use_triton_gated_mlp: false
    wrapper: "CausalSeq2Seq"
    quantization: "no"

    output_dir: "online_ft_adapters"

    fixed_effective_batch_size: true
    gradient_accumulation_steps: 1
    per_device_batch_size: 4
    apply_all_rigids: true
    num_train_epochs: 35
    learning_rate: 0.0008

    prompt_type: prompt_solve_short
    transform_background_color: true
    compress_colors: false

    neftune_noise_alpha: 0.0
    weight_decay: 0.01 # original: 0.01
    max_grad_norm: 0.3 # original: 0.3
    optimizer_name: adamw_8bit

    lora_target_modules: ["all-linear"]
    lora_dropout: 0.00
    lora_alpha: 16
    lora_r: 8

    loss_method: "ce"
    low_memory: false

    early_stopping_patience: 7
    use_early_stopping: false
    eval_steps: 1000
    save_total_limit: 1
    logging_steps: -1
    do_not_use_eval_dataset: true

    untie_word_embeddings: false
    gradient_checkpointing: false
    disable_input_mask: false
    dtype: bfloat16
    kaggle_mode: true # Note: check this one

    padding_side: null # Note: can be ["right", "left", "null"]
    max_length: 20000

    training_set_size: 1 # Not used, it is here because of inheritance
    evaluation_set_size: 1 # Not used, it is here because of inheritance
    eval_split_from_train: false # Note: should not be used in online finetuning

    from_pretrained_config: {}
    use_cpu: false

    input_tokens_limit: 10000

    full_determinism: ${...full_determinism}
    seed: ${...seed}
    load_existing_adapters: false

  decoding_strategy:
    time_limit_parallel_in_seconds: 172800 # 2 days
    max_num_gpus_to_use: ${..online_fine_tuning.max_num_gpus_to_use}

    dataset_dir: ${..online_fine_tuning.dataset_dir}
    dataset_category: ${..online_fine_tuning.dataset_category}
    start_index_tasks: ${..online_fine_tuning.start_index_tasks}
    end_index_tasks: ${..online_fine_tuning.end_index_tasks}
    sort_tasks_by: ${..online_fine_tuning.sort_tasks_by}
    sort_tasks_order: ${..online_fine_tuning.sort_tasks_order}

    wrapper: "${..online_fine_tuning.wrapper}"
    model_id: "model/${...setup.model.name}"
    use_flash_attention_encoder: false
    use_triton_gated_mlp: false
    quantization: "no"
    adapters_location: ${..online_fine_tuning.output_dir}

    traversals: ["row_by_row"]
    dtype: bfloat16
    batch_size: 16
    n_attempts: 2
    apply_all_rigids: false
    always_run_greedy_first: false
    n_transforms: 18
    temperature: 0.0
    do_sample: false
    use_cache: true
    num_beams: 10
    num_return_sequences: 10
    max_new_tokens: 970

    # dynamics_steps: 3
    # dynamics_cum_log_likelihood: -1.0
    # dynamics_occurrence: 0.5
    # dynamics_variance: 5
    # dynamics_operator_1: "or"
    # dynamics_operator_2: "and"
    dynamics_bs_application: full

    monitor_interval: 1.0
    random_seed: 42

    cpu_only: false

    bfs_sampling: false
    bfs_batch_size: 16
    bfs_threshold: 0.1

    save_generation_metadata: false
    input_tokens_limit: 10000
    n_dataloader_workers: 1

    output_dir: "decoding_attempts"

    from_pretrained_config: {}
    use_cpu: false

    full_determinism: ${...full_determinism}
    seed: ${...seed}

  filtering:
    dataset_dir: ${..online_fine_tuning.dataset_dir}
    dataset_category: ${..online_fine_tuning.dataset_category}
    start_index_tasks: ${..online_fine_tuning.start_index_tasks}
    end_index_tasks: ${..online_fine_tuning.end_index_tasks}
    sort_tasks_by: ${..online_fine_tuning.sort_tasks_by}
    sort_tasks_order: ${..online_fine_tuning.sort_tasks_order}
    attempts_location: ${..decoding_strategy.output_dir}
    output_dir: filtered_attempts

    full_determinism: ${...full_determinism}
    seed: ${...seed}
    use_color_features: false

  scoring:
    wrapper: "${..online_fine_tuning.wrapper}"
    time_limit_parallel_in_seconds: 172800 # 2 days
    max_num_gpus_to_use: ${..online_fine_tuning.max_num_gpus_to_use}

    dataset_dir: ${..online_fine_tuning.dataset_dir}
    dataset_category: ${..online_fine_tuning.dataset_category}
    start_index_tasks: ${..online_fine_tuning.start_index_tasks}
    end_index_tasks: ${..online_fine_tuning.end_index_tasks}
    sort_tasks_by: ${..online_fine_tuning.sort_tasks_by}
    sort_tasks_order: ${..online_fine_tuning.sort_tasks_order}

    model_id: "model/${...setup.model.name}"
    use_flash_attention_encoder: true
    use_triton_gated_mlp: false
    adapters_location: ${..online_fine_tuning.output_dir}
    quantization: "no" # "4bit-nf4" # Note: trying to emulate situation in master

    attempts_location: ${..filtering.output_dir}
    output_dir: scored_attempts
    n_attempts_to_keep: 2 # only use to decide if to skip computation

    save_traversals_architects_scores: False
    traversal_ranking_strategy: [['row_by_row', 10], ['row_by_row']]
    dtype: bfloat16
    n_dataloader_workers: 1
    scoring_method: "scoring_with_augmentations"
    mini_arch_top_k: 80
    scoring_with_augmentation_batch_size: 12
    scoring_with_augmentation_aggregation_to_produce_score: sum

    from_pretrained_config: {}
    use_cpu: false

    full_determinism: ${...full_determinism}
    seed: ${...seed}

    input_tokens_limit: 10000

\end{verbatim}

\bibliographystyle{alphaurl}
\bibliography{bibliography}

\end{document}